\documentclass{article}

\usepackage{times}
\usepackage{graphicx} % more modern
\usepackage{subfigure} 

 \usepackage{color}

\usepackage{algorithm}
\usepackage{algorithmic}

\usepackage{amssymb,amsfonts,amsmath,amsthm,amscd,dsfont,mathrsfs}
\usepackage{graphicx,float,psfrag,epsfig}
\usepackage{epstopdf}

\newtheorem{propo}{Proposition}[section]
\newtheorem{lemma}[propo]{Lemma}

\newtheorem{theorem}[propo]{Theorem}

\def\I{\mathbb{I}}

\def\E{{\mathbb{E}}}
\def\P{{\mathbb{P}}}
\def\reals{{\mathbb{R}}}
\def\htk{{\hat{t}^{(k)}}}
\def\i{{\mathbf{I}}}
\def\G{{\mathbf{G}}}
\def\x{{\mathbf{x}}}
\def\y{{\mathbf{y}}}
\def\z{{\mathbf{z}}}
\def\p{p}
\def\q{q}
\def\A{A}
\def\N{{\mathcal{N}}}
\def\F{{{\cal P}}}

\def\l{{\hat{\ell}}}
\def\tl{{\tilde{\ell}}}
\def\r{{\hat{r}}}
\def\t{{\hat{t}}}
\def\I{{\mathbb{I}}}
\def\H{{\mathbb{H}}}
\def\cL{{\cal L}}
\def\cP{{\mathscr{F}}}
\def\cQ{{\mathscr{G}}}

\def\sp{{\sigma^2}}
\def\sq{{\rho^2}}
\def\mup{{\mu}}
\def\stk{{\tilde{\sigma}_k^2}}
\def\pb{{\bar{\mathbf{p}}}}
\def\qb{{\bar{\mathbf{q}}}}
\def\tlambda{{\tilde{\lambda}}}
\def\tdelta{{\tilde{\delta}}}
\def\T{{\tilde{T}}}
\newcommand{\Tau}{\mathscr{T}}

\def\bp{{\bar{p}}}
\def\bq{{\bar{q}}}
\def\lq{\widehat}
\def\lt{\widetilde}

\newcommand{\ceil}[1]{\left \lceil{#1} \right \rceil}

\begin{document} 
\date{}
\author{
Ashish Khetan\thanks{Department of Industrial and Enterprise Systems Engineering, University of Illinois at Urbana-Champaign, email: \texttt{khetan2@illinois.edu}} \;\;
and \;\;
Sewoong Oh\thanks{Department of Industrial and Enterprise Systems Engineering, University of Illinois at Urbana-Champaign, email: \texttt{swoh@illinois.edu}}
}
%\twocolumn[
%\icml\
\title{ Achieving Budget-optimality with Adaptive Schemes in Crowdsourcing }

\maketitle

\begin{abstract}
Crowdsourcing platforms provide marketplaces where task requesters can pay to get labels on their data. 
Such markets have emerged recently as popular venues for collecting annotations that are crucial in training machine learning models in various applications. 
However, as jobs are tedious and payments are low, 
errors are common in such crowdsourced labels. 
A common strategy to overcome such noise in the answers is to add redundancy by 
getting multiple answers for each task and aggregating them using some  methods such as majority voting. 
For such a system, there is a fundamental question of interest: 
how can we maximize the accuracy given a fixed budget on 
how many responses we can collect on 
the crowdsourcing system.
We characterize this 
fundamental trade-off between the budget (how many answers the requester can collect in total) and 
the accuracy in the estimated labels. 
In particular, we ask 
whether adaptive task assignment schemes 
lead to a more efficient trade-off between the accuracy and the budget. 

Adaptive schemes, where tasks are assigned adaptively based on the data collected thus far, 
are widely used in practical crowdsourcing systems to efficiently use a given fixed budget.   
However, existing theoretical analyses of crowdsourcing systems suggest that 
the gain of adaptive task assignments is minimal. 
To bridge this gap, we investigate this question under a strictly more general probabilistic model, which has been recently introduced to model practical crowdsourced annotations. 
Under this generalized Dawid-Skene model, 
we characterize the fundamental trade-off between budget and accuracy. 
We introduce a novel adaptive scheme that matches this fundamental limit. 
We further quantify the fundamental gap between adaptive and non-adaptive schemes, 
by comparing the trade-off with the one for non-adaptive schemes. 
Our analyses confirm that the gap is significant.

\end{abstract}

\section{Introduction}

Crowdsourcing platforms provide labor markets in which pieces of micro-tasks are electronically
distributed to any workers who are willing to complete them for a small fee. 
 In typical crowdsourcing scenarios, such as those on Amazon's
Mechanical Turk, a requester first posts a collection of tasks, for  example a set of images to be labelled.  
Then, from a pool of workers, whoever is willing can pick up a subset of those tasks and provide her 
 labels for a small amount of payment. 
Typically, a fixed amount of payment per task is predetermined and 
agreed upon between the requester and the workers, 
and hence the worker is paid the amount proportional to the number of tasks she answers. 
Further, as the verification of the correctness of the answers is difficult, and also as  the requesters are  
afraid of losing reputation among the crowd, 
requesters typically choose to pay for every label she gets regardless of the 
correctness of the provided labels. 
Hence, the  budget of the total payments the requester makes to the workers is proportional to the total number of labels she collects. 
% task, requester, worker, budget 

One of the major issues in such crowdsourcing platforms is label quality assurance. %\textcolor{blue}{Mostly have used {\em{quality}} for workers parameter.}
Some workers are spammers trying to make easy money, 
and even  those who are willing to work frequently make 
mistakes as  the reward is small and tasks are tedious.  
To correct for these errors, a common approach is to introduce redundancy by collecting answers from multiple workers on the 
same task and aggregating these responses using some schemes such as majority voting. 
A fundamental problem of interest in such a system  is how to maximize the accuracy of thus aggregated answers, 
while minimizing the cost. 
Collecting multiple labels per task can improve the accuracy of our estimates, 
but increases the budget proportionally.
Given a fixed number of tasks to be labelled, 
a requester hopes to 
 achieve the best trade-off between the accuracy and the budget, 
 i.e.~the total number of responses the requester collects on the crowdsourcing platform.
There are two design choices the requester has 
in achieving this goal: {\em task assignment} and {\em inference algorithm}. 
% budget accuracy tradeoff, task assignment, inference algorithm 

In typical crowdsourcing platforms, tasks are assigned as follows. 
Since the workers are fleeting, the requester has no control 
over who will be the next arriving worker. 
Workers arrive in an online fashion, complete the tasks that they are given, and leave. 
Each arriving worker is completely new and you may never get her back. 
Nevertheless, it might be possible to improve accuracy under the same budget, 
by designing better task assignments. 
The requester has the following control over the {\em task assignment}.  
At each point in time, we have the control over which tasks to assign to the next arriving worker.  
The requester is free to use all the information collected thus far, 
including all the task assignments to previous workers and 
the answers collected on those assigned tasks. 
By adaptively identifying tasks that are more difficult and 
assigning more (future) workers on those tasks, 
one hopes to be more efficient in the budget-accuracy trade-off. 
This paper makes this intuition precise, by 
studying a canonical crowdsourcing model 
and comparing the fundamental trade-offs between adaptive schemes and non-adaptive schemes. 
Unlike adaptive schemes, a non-adaptive scheme fixes all the task assignments before any labels are collected and does not allow  future assignments to 
adapt to the labels collected thus far for each arriving worker. Precise definitions of adaptive and 
non-adaptive task assignments are provided in Section \ref{sec:model}. 

While adaptive task assignments handle the
 heterogeneity in the task difficulties by 
assigning more workers to difficult tasks, 
inferring such unknown difficulty of the tasks (as well as inferring unknown heterogeneity of the worker reliabilities) 
requires {\em inference}: 
estimating the latent parameters and the ground truth labels from 
crowdsourced responses thus far. 
Some workers are more reliable than the others, but we do not know their latent reliabilities. 
Some tasks are more difficult than the others, but we do not know their latent difficulty levels. 
We only get to observe the answers provided by those workers on their assigned tasks. 
Nevertheless, by comparing responses from multiple workers, we can estimate the true labels and the difficulties of  the tasks,
and use them in subsequent steps in our inference algorithm to learn the reliability of the workers.
We perform such inferences at several points in time 
over the course of collecting all the labels we have budgeted for.   
%During a single inference execution, 
%given the data collected thus far, 
%our beliefs on the true answers of the tasks as well as the difficulty of the tasks and the reliability of the workers are iteratively refined.
%and one can potentially choose to assign more workers to the more difficult tasks.  
The inference algorithm outputs the current estimates for the labels and 
difficulty levels of the tasks, which 
are used in subsequent time to assign tasks.  
%We would like to understand such intricate interplay of task assignment and inference. 

\subsection{Model and problem formulation}
\label{sec:model}

We assume that 
the requester has $m$ binary classification tasks to be labelled 
by querying a crowdsourcing platform 
multiple times. 
For example, those might be image classification tasks, where 
the requester wants to classify $m$ images as either 
suitable for children ($+1$) or not ($-1$). 
The requester has a budget $\Gamma$ on how many responses 
she can collect on the crowdsourcing platform, assuming one unit of payment 
is made for each response collected.  
We use $\Gamma$ interchangeably to refer to both a target budget 
and also the budget used by a particular task assignment scheme 
(as defined in \eqref{eq:defgamma}), and it should be clear from the context which one we mean. 
%using a crowdsourcing platform. 
%all tasks are binary classification tasks. 
%We use  $m$ to denote the number of tasks we have to be labelled. 
We want to find the true label  by querying noisy workers who are arriving in an 
online fashion, one at a time. 

\bigskip
\noindent
{\bf Task assignment and inference.} 
Typical crowdsourcing systems are modeled as a discrete time systems where 
at each time we have a new arriving worker. 
At time $j$, the requester 
chooses an action $T_j \subseteq [m]$, 
which is a subset of tasks  to be assigned to the $j$-th  arriving worker. 
Then, the $j$-th arriving worker provides her answer  
$A_{ij} \in \{+1,-1\}$ for each task $i\in T_j$. 
We use the index $j$ to denote both the $j$-th time step 
in this discrete time system as well as the $j$-th arriving worker. 
At this point (at the end of $j$-th time step), 
all previous responses are stored in a sparse matrix $A \in \{0,+1,-1\}^{m\times j}$, and this data matrix is increasing by one column at each time. 
We let $A_{ij}=0$ if task $i$ is not assigned to worker $j$, i.e.~$i\notin T_j$, 
and otherwise we let $A_{ij}\in\{+1,-1\}$ be the previous worker $j$'s response on task $i$.  
At the next time $j+1$, the next task assignment $T_{j+1}$ is chosen, and this process is repeated.  
At time $j$, the action (or the task assignment) 
can depend on all previously collected responses up to 
the current time step stored in 
a sparse (growing) matrix $A \in \{0,+1,-1\}^{m\times (j-1)}$. 
This process is repeated until 
the task assignment scheme decides to stop, typically when 
the total number of collected responses 
(the number of nonzero entires in $A$) meet a certain budget constraint 
or when a certain target accuracy is estimated to be met.  

%In summary, this is a discrete time process  
%where at time $j$ 
%the requester's action is $T_j \subseteq [m]$ (the set of tasks to be assigned to the worker arriving at time $j$). 
%This $T_j$ can be chosen as a function of all previous responses encoded in a sparse (growing) matrix $A \in \{0,+1,-1\}^{m\times (j-1)}$. 
%Once the responses are collected and stored in the $j$-th column of $A$, this process is repeated. 
We consider both a {\em non-adaptive scenario} and 
an {\em adaptive scenario}. 
In a non-adaptive scenario,  
a fixed number $n$ of workers to be recruited are 
pre-determined (and hence the termination time is set to be $n$) 
and also fixed task assignments $T_j$'s for all $j\in[n]$ are pre-determined, before any response is collected. 
In an {\em adaptive scenario}, the requester 
chooses $T_j$'s in an online fashion 
based on all the previous answers collected thus far. 
%However, we emphasize that the workers are fleeting and cannot be reused based on the their response; in particular, we cannot explore reliable workers and exploit them, using techniques similar to multi-armed bandits.  
For both adaptive and non-adaptive scenarios, 
when we have determined that we have collected all the data we need, 
an inference algorithm is applied on the collected data $A\in\{0,+1,-1\}^{m\times n}$ 
to output an estimate 
$\hat{t}_i\in\{+1,-1\}$ for the ground truth label $t_i\in\{+1,-1\}$ for the $i$-th task for each $i\in[m]$. 
Note that we use $n$ to denote the total number of 
workers recruited, which is a random variable under  the adaptive scenario. 
Also, note that the estimated labels for all the tasks do not have to be simultaneously output in the end, and 
we can choose to output estimated labels on some of the tasks in the middle of the  process before termination.   
The average accuracy of our estimates is measured by the average probability of error 
$P_{\rm error} = (1/m) \sum_{i=1}^m \P[t_i\neq \hat{t}_i]$ under a probabilistic model to be defined later in this section in Eq.~\eqref{eq:defA}. 

The total {\em budget} used in one instance of such a process  
is measured by the total number of responses collected, which is equal to 
the number of non-zero entries in $A$. 
This inherently assumes that there is a prefixed fee of one unit
 for each response that is agreed upon, and the requester pays this
constant fee for every label that is collected. 
The expected budget used by a particular task assignment scheme will be denoted by 
\begin{eqnarray}
	\Gamma \;\;\equiv\;\;  \E \Big[ \sum_{j=1}^n |T_j| \Big] \;, 
	\label{eq:defgamma}
\end{eqnarray}
where the expectation is over all the randomness in the model 
(the problem parameters
representing the quality of the tasks and  
the quality of the workers, 
and the noisy responses from workers) 
and any randomness used in the task assignment. 
We are interested in designing task assignment schemes and inference algorithms that achieve the best accuracy within a target expected budget,  
under the following canonical model of how workers respond to tasks. 

\bigskip
\noindent
{\bf Worker responses.} 
We assume that when a task is assigned to a worker, 
the response follows a probabilistic model introduced by  
\cite{ZLPCS15}, which is a recent generalization of 
the Dawid-Skene model originally introduced by \cite{DS79}. 
%to model the responses, which captures the heterogeneity in the tasks as well as the workers. 
Precisely, each new arriving worker is 
parametrized by a latent worker quality parameter $p_j\in[0,1]$ (for the $j$-th arriving worker). 
Each task is parametrized  by a latent task quality parameter $q_i\in[0,1]$ (for the $i$-th task). 
When a worker $j$ is assigned a task $i$, 
the {\em generalized Dawid-Skene model} assumes that    
the response $A_{ij}\in\{+1,-1\}$ is a random variable  distributed as 
\begin{eqnarray}
	A_{ij} \;\; = \;\;  \left\{ 
	\begin{array}{rl}
	 +1, &\;\;\;\; \text{w.p.} \;\;\; q_ip_j + \bar{q}_i \bar{p}_j \;, \\
	  -1, &\;\;\;\; \text{w.p.} \;\;\;  q_i \bar{p}_j + \bar{q}_i p_j \;,
	 \end{array}
	\right.\;,
	\label{eq:defA}
\end{eqnarray}
conditioned on the parameters $q_i$ and $p_j$, 
where $\bar{q}_i=1-q_i$ and $\bar{p}_j = 1-p_j$.
 The task parameter $q_i$ represents the probability that a task is perceived as a positive task to a worker, and the worker parameter $p_j$ represents the probability the worker makes a mistake in labelling the task.
Concretely, when a task $i$ is presented to any worker, the task is perceived as a positive task  with a probability $q_i$ or a negative task otherwise,  independent of any other events. 
Let $\tilde{t}_{ij}$ denote this perceived label of task $i$ as seen by worker $j$. 
Conditioned on this perceived label of the task,  
a worker $j$ with parameter $p_j$ 
makes a mistake with probability $1-p_j$. 
She provides a `correct' label $\tilde{t}_{ij}$ 
as she perceives it with probability $p_j$, 
or provides an `incorrect' label $-\tilde{t}_{ij}$ with probability $\bar{p}_j$.  
Hence, the response $A_{ij}$ follows the distribution in \eqref{eq:defA}.
The response $A_{ij}$ is, for example, a positive label 
if the task is perceived as a positive task and the worker does not make a mistake (which happens with a probability $q_ip_j$), 
or if the task is perceived as a negative task and the worker does not make a mistake (which happens with a probability $\bar{q}_i\bar{p}_j$). 
\textcolor{black}{Alternately, the task parameter $q_i$ represents the probability that a task is labeled as a positive task by a perfect worker, a worker with parameter $p_j = 1$. That is $q_i$ represents inherent ambiguity of the task being labeled positive.}
The strengths and weaknesses of this model are discussed in 
comparisons to  related work in Section \ref{sec:related}.

\bigskip
\noindent
\textcolor{black}{{\bf Prior distribution on worker reliability.}}
We assume that worker parameters $p_j$'s are 
 i.i.d.~according to some prior distribution $\cP$. 
 For example, each arriving worker might be sampled with replacement from 
 a pool of workers, and $\cP$ denotes the discrete distribution of the 
quality parameters of the pool. 
The individual reliabilities  $p_j$'s are hidden from us, 
and the prior distribution $\cP$ is also unknown. 
\textcolor{black}{We assume we only know some statistics of the 
prior distribution $\cP$,} namely 
\begin{eqnarray}
	\label{eq:defmu}
	\mu \; \equiv\; \E_\cP[2p_j-1] \; \text{, and } \;\;\;\;\;
	\sigma^2 \;\equiv \; \E_\cP[(2p_j-1)^2] \;, 
\end{eqnarray}
where $p_j$ is a random variable distributed as $\cP$, and 
$\mu\in[-1,1]$ is the (shifted and scaled) average reliability of the crowd  
and $\sigma^2\in[0,1]$ is the key quantity of $\cP$ capturing the 
collective quality of the crowd as a whole. 
Intuitively, when all workers are truthful and have $p_j$ close to a one,
 then the collective reliability $\sigma^2$ will be close to its maximum value of one. On the other hand, if most of the workers are giving completely random answers with $p_j$'s close to a half, then $\sigma^2$ will be close to its minimum value of a zero. 
The fundamental trade-off between the accuracy and the budget will primarily depend on the distribution of the crowd $\cP$ via $\sigma^2$. We do not impose any conditions on the distribution $\cP$. 

\bigskip
\noindent
\textcolor{black}{{\bf Prior distribution on task quality.}}
We assume that the 
task parameters $q_i$'s are  
drawn i.i.d.\ according to some prior distribution $\cQ$. 
The individual difficulty of a task with a quality parameter $q_i$ is naturally captured by 
\begin{eqnarray}
	\lambda_i \;\;\equiv\;\; (2q_i-1)^2 \;,
	\label{eq:deflambdai}
\end{eqnarray}
as 
tasks with  $q_i$ close to a half  are    
confusing and ambiguous tasks and hence difficult to correctly label ($\lambda_i$ close to zero), 
whereas tasks with $q_i$ close to zero or one are unambiguous tasks and easy to correctly label ($\lambda_i$ close to one). 
\textcolor{black}{The average difficulty} and the collective difficulty of 
tasks drawn from a prior distribution $\cQ$
are captured by the quantities $\rho \in [0,1]$ and $\lambda \in [0,1]$, defined as  
\begin{eqnarray}
	\textcolor{black}{\rho \;\;\equiv\;\;  \E_\cQ\left[ {(2q_i-1)^2}\right] }\;,\quad \lambda \;\;\equiv\;\;  \left(\E_\cQ\left[ \frac{1}{(2q_i-1)^2}\right]\right)^{-1} \;,
	\label{eq:deflambda}
\end{eqnarray}
where $q_i$ is distributed as $\cQ$. 
The fundamental budget-accuracy trade-off depends  on $\cQ$ 
primarily via this $\lambda$. 
Another quantities that will show up in our main results is the worst-case difficulty in the given set of $m$ tasks (conditioned on all the $q_i$'s) defined as  
\begin{eqnarray}
	\lambda_{\min}  \;\; \equiv\;\;  \min_{i \in [m]} (2q_i-1)^2\,, \qquad \text{and} \qquad \lambda_{\max}  \;\; \equiv\;\;  \max_{i \in [m]} (2q_i-1)^2 \;.  
	\label{eq:deflambdamin}
\end{eqnarray}
When we refer to a similar quantities from 
the population distributed as $\cQ$, we abuse the notation and 
denote $\lambda_{\rm min} = \min_{q_i \in {\rm supp}(\cQ)} (2q_i-1)^2$ and  $\lambda_{\rm max} = \max_{q_i \in {\rm supp}(\cQ)} (2q_i-1)^2$. 
The individual task parameters $q_i$'s are hidden from us.  
We do not have access to the prior distribution $\cQ$ on the task qualities  $q_i$'s, 
but we assume we  know  the statistics $\rho$, $\lambda$, $\lambda_{\min}$, and $\lambda_{\max}$, 
and we assume we also know a quantized version of the prior distribution on the task difficulties $\lambda_i$'s, which we explain below. 

\bigskip\noindent
\textcolor{black}{{\bf Quantized prior distribution on task difficulty.}}
Given a distribution $\cQ$  on $q_i$'s, let $\lt\cQ$ be the induced distribution on $\lambda_i$'s. 
For example, if $\cQ(q_i)= (1/10)\mathbb{I}_{(q_i=0.9)} + (3/10)\mathbb{I}_{(q_i=0.1)} + (1/10)\mathbb{I}_{(q_i=0.8)} + (3/10)\mathbb{I}_{(q_i=0.2)} + (2/10)\mathbb{I}_{(q_i=0.6)} $, 
then the induced distribution on $\lambda_i$ is $\lt\cQ(\lambda_i) = (4/10)\mathbb{I}_{(\lambda_i=0.64)} + (4/10)\mathbb{I}_{(\lambda_i=0.36)}  + (2/10)\mathbb{I}_{(\lambda_i=0.04)}$.
Our approach requires only the knowledge of 
 a quantized version of the distribution $\lt\cQ$, namely $\lq\cQ$. 
This quantized distribution 
has support at $\T$ discrete values $\{\lambda_{\rm max}, \lambda_{\rm max}/2, \ldots, \lambda_{\rm max}/2^{(\T-1)} \}$, where 
%$\lq\cQ$ is supported at at most at $\T$ points, 
\begin{eqnarray}
	\T \;\; \equiv \;\; 1 +  \Big\lceil \log_2 \Big(\frac{\lambda_{\max}}{\lambda_{\min}}\Big)  \Big\rceil \;,
\end{eqnarray}
 such that 
$\lambda_{\max} 2^{-(\T-1)} \leq \lambda_{\min} \leq \lambda_{\max} 2^{-(\T-2)}$. 
We denote these values by $\{\tlambda_a\}_{a \in [\T]}$ such that  
% be the set of possible support points of $\lq\cQ$. We define support points to be such that they are geometrically decreasing. In particular, 
 $\tlambda_{a} = \lambda_{\rm max}2^{-(a-1)}$ for each $a \in [\T]$. 
%The choice of 2 for the ratio of $\tlambda_a$'s is arbitrary and can be further optimized for a given distribution of $\lambda_i$'s.
Then the quantized distribution is 
$\sum_{a=1}^\T \tdelta_a {\mathbb I}_{(\lambda_i=\tlambda_a)}$, where 
the probability mass $\tdelta_a$ for the $a$-th partition is 
%Let $\tdelta_a$ be the total fraction of tasks whose difficulty $\lambda_i$ is smaller than $\tlambda_1 2^{-(a-2)}$ and larger than $\tlambda_1 2^{-(a-1)}$. 
%Precisely,   
%$\tdelta_a = (1/n) \sum_{i \in [n]}  \I\big\{ \tlambda_1/2^{(a-1)} \leq \lambda_{i} < \tlambda_1/2^{(a-2)} \big\}$\,, for $a \in [T]\,$.  
%Precisely, let $h_{\cQ_{\lambda}}$ be the CDF of the distribution $\cQ_{\lambda}$. Then 
$$\tdelta_a = \lt\cQ( \, ( {\lambda_{\rm max}/2^{a}},{\lambda_{\rm max}/2^{(a-1)}}] \,) \,, \qquad \text{for}\;\; a \in [\T]\,,$$
which is the fraction of tasks whose difficulty $\lambda_i$ is in $(\tlambda_{a+1},\tlambda_a] $. %smaller than $\tlambda_1 2^{-(a-2)}$ and larger than $\tlambda_1 2^{-(a-1)}$. 
We use the closed interval $[(1/2)\tlambda_\T,\tlambda_\T]$ for the last partition. 
In the above example, we have 
$\T=5$, 
$\{\tlambda_a\}_{a\in\T}=\{0.64, 0.32, 0.16, 0.08, 0.04\}$, 
and $\{\tdelta_a\}_{a\in\T} = \{0.8,0,0,0,0.2\}$.
For notational convenience, we eliminate those partitions with zero probability mass, and 
 re-index the quantization $\{\tlambda_a, \tdelta_a\}_{a \in [\T]}$ to get $\{\lambda_a, \delta_a\}_{a \in [T]}$, for $T \leq \T$, such that $\delta_a \neq 0$ for all $a \in T$. 
 We define $\lq\cQ$ to be the re-indexed quantized distribution $\{\lambda_a, \delta_a\}_{a \in [T]}$.
 In the above example, we finally have $\lq\cQ(\lambda_i)= 0.8{\mathbb I}_{(\lambda_i=0.64)} + 0.2{\mathbb I}_{(\lambda_i=0.04)}$. 
 
We denote the maximum and minimum probability mass 
in $\lq\cQ$ as 
\begin{eqnarray}
	\delta_{\max} \;\; \equiv \;\; \max_{a \in [T]} \delta_a
	\;,\;\;\text{ and }\;\;\;\;\;
	\delta_{\min} \;\;\equiv\;\; \min_{a \in [T]} \delta_a\;. 
	\label{eq:defdeltamin}
\end{eqnarray}
Similar to the collective quality $\lambda$ defined for the distribution $\cQ$ in \eqref{eq:deflambda}, we define $\lq\lambda$, collective quality for the quantized distribution $\lq\cQ$, 
which is used in our algorithm. $\lq\lambda \equiv (\sum_{a \in [T]} (\delta_a/\lambda_a))^{-1}$.

\bigskip\noindent
\textcolor{black}{{\bf Ground truth.}}
The ground truth label of a task is also naturally defined as 
what the majority of the crowd would agree on if we ask all the workers to 
label that task, i.e.~$t_i \,\equiv\, {\rm sign}(\E[A_{ij}|q_i]) = {\rm sign}(2q_i-1){\rm sign}(\mu)$, 
where the expectation is with respect to the prior distribution of 
$p_j\sim \cP$ and the randomness in the response as per the generalized Dawid-Skene model in \eqref{eq:defA}.  
Without loss of generality, we assume that the average reliability of the worker is positive, i.e.~${\rm sign}(\mu)=+1$ and take  
${\rm sign}(2q_i-1)$ as the ground truth label $t_i$ of task $i$ conditioned on its difficulty parameter $q_i$:
\begin{eqnarray}
	t_i \;\; =\;\;  {\rm sign}(2q_i-1) \,.
	\label{eq:deft}
\end{eqnarray} 
The latent parameters $\{q_i\}_{i\in[m]}$, $\{p_j\}_{j\in[n]}$, and $\{t_i\}_{i\in[m]}$ are unknown, 
and we want to infer the true labels $t_i$'s from only $A_{ij}$'s. 

\bigskip
\noindent
\textcolor{black}{{\bf Performance measure.}}
The accuracy of the final estimate is measured by 
the average probability of error: 
\begin{eqnarray}
	\label{eq:deferror}
	P_{\rm error} \;\;= \;\; \frac1m\sum_{i=1}^m  \P[t_i\neq \hat{t}_i] \;.
\end{eqnarray}
We investigate the fundamental trade-off between 
budget and error rate by identifying the 
sufficient and necessary conditions on the expected budget $\Gamma$ 
for achieving a desired level of accuracy $P_{\rm error}\leq \varepsilon$.
Note that we are interested in achieving the best trade-off, which in turn can give 
the best  approach for both scenarios: 
when we have a fixed budget constraint and want to minimize the error rate, 
and when we have a target error rate and want to minimize the cost.

\subsection{Related work} 
\label{sec:related}

The generalized Dawid-Skene model studied in this paper 
allows the tasks to be heterogeneous (having different difficulties) 
and the workers to be heterogeneous (having different reliabilities). 
%Note that we focus on only binary tasks with two types of classes, and also the workers are assumed to be symmetric, i.e. the error probability 
%is the same whether the task is perceived as a positive or a negative task. 
The original  Dawid-Skene (DS) model introduced in \cite{DS79} and analyzed in \cite{KOS14OR} is a special case,  
when only workers are allowed to be heterogeneous. 
All tasks have the same difficulty with $\lambda_i=0$ for all $i\in[m]$ 
and $q_i$ can be either zero or one depending on the true label. 
 Most of existing work on the DS model assumes that tasks are randomly assigned and focuses only on the inference problem of 
 finding the true labels. 
Several inference algorithms have been proposed \cite{DS79,smyth95,JG03,SPI08,GKM11,KOSnips11,Liu12,ZPBM12,LY14,ZCZJ14,DDKR13,KOS13SIGMETRICS,OOSY16,BC16,BC16_adapt,MOS17}. 

A most relevant work is by 
 \cite{KOS14OR}. It is shown that 
 in order to achieve a probability of error less than a small positive constant 
 $\varepsilon>0$, 
 it is necessary to have an expected budget scaling as 
 $\Gamma = O((m/\sigma^2)\log(1/\varepsilon))$, 
 even for the best possible inference algorithm together with the best possible 
 task assignment scheme, including all possible adaptive task assignment schemes.  
Further, a simple randomized non-adaptive task assignment 
is proven to achieve this optimal trade-off with a novel spectral inference algorithm. Namely,  an efficient task assignment and an inference algorithm are proposed that together guarantees to achieve $p_{\rm error}\leq \varepsilon$ with budget scaling as $\Gamma=O((m/\sigma^2)\log(1/\varepsilon))$. 
 It is expected that this necessary and sufficient  budget constraint 
 scales linearly in $m$, the number of 
 tasks to be labelled. 
 The technical innovation of  \cite{KOS14OR} is in $(i)$ designing a new spectral algorithm that achieves a logarithmic dependence in the target error rate $\varepsilon$; and 
 $(ii)$ 
 identifying $\sigma^2$ defined in \eqref{eq:defmu} as the fundamental 
 statistics of $\cP$ that captures the collective quality of the crowd.  
 The budget-accuracy trade-off mainly depends on the prior distribution of the crowd $\cP$ via a single parameter $\sigma^2$. 
When we have a reliable crowd with many workers having 
$p_j$'s close to one, the collective quality $\sigma^2$ is close to one and the required budget $\Gamma$ is small. 
When we have an unreliable crowd with many workers having $p_j$'s close to 
a half, then the collective quality is close to zero and the required budget is large. 
However, perhaps one of the most surprising result of 
 \cite{KOS14OR} is that the optimal trade-off is matched by a non-adaptive task assignment scheme. In other words, there is only a marginal gain in using adaptive task assignment schemes. 

%
%Perhaps surprisingly, for the standard DS model, a {\em non-adaptive task assignment} scheme achieves the fundamental limit.  
%Namely, given $m$ tasks and a total budget for $m\ell$ responses, 
%the requester first constructs  a bipartite task-assignment graph with $m$ task nodes, $n=m\ell/r$ worker nodes, and 
%edges drawn uniformly at random with degree $\ell$ for the task nodes and $r$ for the worker nodes. 
%Then, $j$-th  arriving worker is assigned a batch of $r$ tasks that are adjacent to the $j$-th worker node.  
% Together with an inference algorithm explained in detail in Section \ref{sec:main}, this achieves a near-optimal performance. 
%Namely, to achieve an average probability of error $\varepsilon$, it is sufficient to have total budget $O((m/ \sigma^2) \log(1/\varepsilon))$, 
%where $\sigma^2=\E_\cP[(2p_j-1)^2]$ is the quality of the workers defined in \eqref{eq:defbeta1}. 
%Perhaps surprisingly, no adaptive assignment can improve upon it. 
%Even the best adaptive scheme and the best inference algorithm still requires 
%$\Omega((m /\sigma^2) \log(1/\varepsilon))$ total budget. 
%Hence, there is no gain in adaptivity. 

This negative result relies crucially on the fact that, under the standard DS model, all tasks are inherently equally difficult. 
As all tasks have $q_i$'s either zero or one, 
the individual difficulty of a task is $\lambda_i\equiv (2q_i-1)^2=1$, and  
a worker's probability of making an error on one task is 
the same as any other tasks. 
Hence, adaptively assigning more workers to relatively more ambiguous tasks has only a marginal gain. 
However, simple adaptive schemes are widely used in practice, where significant gains are achieved.   
In real-world systems, tasks are widely heterogeneous.
Some images are much more difficult to classify (and find the true label) compared to other images. 
To capture such varying difficulties in the tasks, 
generalizations of the DS model were proposed in \cite{whitehill09,Welinder10,ZLPCS15,SBW16} and 
significant improvements have been reported  on  real datasets. 

The generalized DS model serves as the missing piece in 
bridging the gap between practical gains of adaptivity and 
theoretical limitations of adaptivity (under the standard DS model). 
We investigate the fundamental question of ``do adaptive task assignments improve accuracy?'' under this generalized Dawid-Skene model of Eq.  \eqref{eq:defA}.

%, in particular in the case of spammer-hammer model as it gives same weight to all the responses. 
%In a general scenario, when the crowd is of diverse quality, 
%we can iteratively infer reliability of each worker and update our beliefs on the tasks.  
%This idea was first introduced by Dawid and Skene \cite{DS79} who proposed an Expectation Maximization (EM) approach. 
% Their iterative algorithm is based on expectation maximization (EM) which does the following: it starts with an initial guess of task labels and estimates worker reliabilities based on their answers; and then based on the estimated worker reliabilities it estimates task labels given workers responses; and repeats. 
%Following this work, several approaches have been proposed to solve the inference problem under the DS model 
%\cite{smyth95,SPI08,KOSnips11,Liu12,ZPBM12,LY14,ZCZJ14,DDKR13}. 

On the theoretical understanding of the original DS model, the dense regime has been studied first, 
where all workers are assigned all tasks. 
A spectral method for finding the true labels was first analyzed in \cite{GKM11} and an EM approach followed by spectral initial step is 
analyzed in \cite{ZCZJ14} to achieve a near-optimal performance. The minimax error rate of this problem was identified in \cite{GZ13} by analyzing the MAP estimator, which is computationally intractable. 
%However, in this dense regime, all tasks are assigned a growing number of workers as the problem size increases and eventually all tasks are labelled correctly with high probability. 

In this paper, we are interested in a more challenging setting 
where each task is assigned only a small number of workers 
of $O(\log m)$. 
% where $n$ is the total number of workers used by the taskmaster. 
For a non-adaptive task assignment, a novel spectral algorithm based on the non-backtracking operator of the matrix $A$ 
has been analyzed under the original DS model %in \cite{DDKR13}, and 
by \cite{KOSnips11}, which showed that the proposed spectral approach is  near-optimal. 
Further, \cite{KOS14OR} showed that any non-adaptive task assignment scheme will have only marginal improvement  in the error rate under the original DS model. 
Hence, there is no significant  gain in adaptivity. 
%Although the new spectral method provably achieves near-optimal performance, 
%there is still a small but non-zero gap of a constant factor in the upper and lower bound on the required budget. 
%This gap was closed in a recent work by \cite{OOSY16}, 
%where belief propagation for this model is analyzed and proven to be exactly optimal under certain conditions. 
%Precisely, it was proven that there is no algorithm that can achieve a smaller error probability than belief propagation 
%for crowdsourcing under the original DS model.  
%A weighted majority voting algorithm has been analyzed in \cite{LYZ13}, showing the dependence on the choice of the weights. 

One of the main weaknesses of the DS model is that it does not capture how some tasks are more difficult than the others.
To capture such heterogeneity in the tasks, 
several practical models have been proposed recently \cite{JG03,whitehill09,Welinder10,ZLPCS15,HJV13}. 
Although such models with more parameters can potentially better describe  real-world datasets,   
 there is no analysis on their performance under adaptive or non-adaptive task assignments. 
We do not have the analytical tools to understand the fundamental trade-offs involved in those models yet. 
In this work, we close this gap by providing a theoretical analysis of one of the generalizations of the DS model, 
namely the one proposed in \cite{ZLPCS15}.
It captures the heterogeneous difficulties in the tasks, 
while remaining simple enough for theoretical analyses. 
%There are more general models in crowdsourcing literature, such as those in 
%\cite{whitehill09,Welinder10,HJV13}, but 

\subsection{Contributions} 

To investigate the gain of adaptivity, 
we first characterize the fundamental lower bound on the budget required 
to achieve a target accuracy. 
To match this fundamental limit, 
we introduce a novel {\em adaptive} task assignment scheme.  
The proposed adaptive task assignment is simple to apply in practice, and numerical simulations confirm the superiority compared to state-of-the-art non-adaptive schemes. 
Under certain assumptions on the choice of parameters in the algorithm, which requires a moderate access to an oracle, 
we can prove that the performance of the proposed adaptive scheme matches that of the fundamental limit up to a constant factor. 
% the theoretical findings. 
Finally, we quantify the gain of adaptivity by proving a strictly larger lower bound on the budget required for any non-adaptive schemes to achieve 
a desired error rate of $\varepsilon$ for some small positive $\varepsilon$. 

Precisely, we show that the minimax rate on the budget required to achieve 
a target average error rate of $\varepsilon$ scales as 
$\Theta((m/ \lambda \sigma^2 ) \log(1/\varepsilon))$. 
The dependence on the prior $\cP$ and $\cQ$ are solely captured in $\sigma^2$ (the quality of the crowd as a whole) and  
$\lambda$ (the quality of the tasks as a whole). 
%We provide a novel adaptive task assignment scheme that achieves this fundamental tradeoff. 
We show that the fundamental trade-off for {\em non-adaptive} schemes is 
$\Theta((m/ \lambda_{\rm min} \sigma^2 ) \log(1/\varepsilon))$, requiring 
a factor of $\lambda/\lambda_{\rm min}$ larger budget for non-adaptive schemes. 
This  factor of $\lambda/\lambda_{\rm min}$ is always at least one and 
quantifies precisely how much we gain by adaptivity. 
%, and this gain can be made arbitrarily large for a worst case distribution $\cQ$. 

\subsection{Outline and notations} 

We present a list of notations and their definitions 
in Table \ref{tbl:notation}.  
In Section \ref{sec:main}, we present the fundamental lower bound on 
the necessary budget to achieve a target average error rate of $\varepsilon$. 
We present a novel adaptive approach which achieves the fundamental lower bound up to a constant. 
In comparison, we provide the fundamental lower bound on the necessary budget for non-adaptive approaches in Section \ref{sec:nonad_ub}, 
and 
we present a non-adaptive approach that achieves this fundamental limit. 
In Section \ref{sec:spectral}, we give a spectral interpretation of our approach 
justifying the proposed inference algorithm, leading to a parameter estimation 
algorithm  that serves as a building block in the main approach of Algorithm \ref{algo:adapt}. 
As our proposed sub-routine using Algorithm \ref{algo:mp} suffers when the budget is critically limited (known as spectral barrier in Section \ref{sec:spectral}), we present another algorithm that can substitute 
Algorithm \ref{algo:mp} in Section \ref{sec:algo} and compare their performances. 
The proofs of the main results are provided in Section \ref{sec:proof}. 
We present a conclusion with future research directions in Section \ref{sec:discussion}.

%
%
%We consider all randomized task assignment schemes, whose expected number of assignment per task is $\ell$, and all inference algorithms. 
%We study the minimax rate when the nature chooses the worst case priors $\cP$ and $\cQ$ 
%(from a family of priors parametrized by average worker reliability $\sigma^2$ and average task difficulty $\lambda$ defined in \eqref{eq:defbeta1}), and we choose the best possible adaptive task assignment together with the best possible inference algorithm.  
%We further propose a novel adaptive approach that achieves this minimax rate up to a constant factor. 
%Our approach is different from existing adaptive schemes studied by  
%\cite{HJV13}, and a detailed comparison is provided in Section \ref{sec:discussion}. 
%where there are multiple types of tasks and 
%the main source of uncertainty is which type the next arriving worker is expert on. 
%Golden tasks with known answers are used to explore expertise  
%and tasks are assigned accordingly. 

\begin{table}[h!]
\begin{center}
  \begin{tabular}{ | c | c | c|  }
    \hline
    notation& data type & definition    \\ \hline    
    \hline
    $m$ & $\mathbb{Z}_+$ & the number of tasks    \\ \hline
    $n$ &  $\mathbb{Z}_+$ & total number of workers recruited  \\ \hline
    $A=[A_{ij}]$   & $\{0,+1,-1\}^{m\times n}$ & labels collected from the workers \\\hline
    $\Gamma$ &$\reals_+$ & budget used in collecting $A$ is the number of nonzero entries in $A$ \\\hline
        $\ell$ &$\mathbb{Z}_+$  & average budget per task $:\Gamma/m$\\\hline
    $\Gamma_\varepsilon$ & $\reals_+$ & the  budget  required to achieve error at most $\varepsilon$ \\\hline
$\Tau_{\Gamma}$ &  & set of task assignment schemes 
using  at most $\Gamma$ queries in expectation \\\hline
    $i$ & [m]& index for tasks \\\hline
    $j $  & [n] & index for workers  \\\hline
    $W_i$ & subset of $[n]$ & a set of workers assigned to task $i$ \\\hline
    $T_j$ &  subset of $[m]$ & a set of tasks assigned to worker $j$ \\\hline
    $q_i$ &  $[0,1]$ & quality parameter of task $i$  \\\hline
    $t_i$ &   $\{-1,+1\}$ & ground truths label of task $i$  \\\hline
    $\hat{t}_i$ &   $\{-1,+1\}$ & estimated label of  task $i$  \\\hline
    $p_j$ &  $[0,1]$ & quality parameter of worker $j$ \\\hline
    $\cP$ &  $[0,1] \to \reals$ & prior distribution of $p_j$   \\\hline
    $\cQ$ &  $[0,1]\to\reals$  & prior distribution of $q_i$ \\\hline
    $\lt\cQ$ &  $[0,1]\to\reals$  & \textcolor{black}{prior distribution of $\lambda_i$  induced from $\cQ$}\\\hline
    $\lq\cQ$ &  $[0,1]\to\reals$  & \textcolor{black}{quantized version of the distribution $\lt\cQ$ }\\\hline    
    $\mu$ &  $[-1,1]$& average reliability of the crowd as per $\cP$: $\E_{\cP}[2p_j-1]$  \\\hline
    $\sigma^2$ & $[0,1]$& collective reliability of the crowd as per $\cP$: $\E_{\cP}[(2p_j-1)^2]$   \\\hline
    $\lambda_i$ & $[0,1]$& individual difficulty level of task $i$: $(2q_i-1)^2$   \\\hline
    $\lambda_{\rm min}$ &  $[0,1]$& worst-case difficulty as per $\cQ$: $\min_{q_i \in {\rm supp}(\cQ)} (2q_i-1)^2$  \\\hline
    $\lambda_{\rm max}$ &  $[0,1]$& \textcolor{black}{best-case difficulty as per $\cQ$: $\max_{q_i \in {\rm supp}(\cQ)} (2q_i-1)^2$}  \\\hline    
    $\lambda$ & $[0,1]$& collective difficulty level of the tasks as per $\cQ$: $\E_{\cQ}[(2q_i-1)^{-2}]^{-1}$   \\\hline
    $\lq\lambda$ & $[0,1]$& collective difficulty level of the tasks as per $\lq\cQ$: $(\sum_{a \in [T]} {\delta_a}/{\lambda_a})^{-1}$   \\\hline    
     $\rho^2$ & $[0,1]$ & average difficulty of tasks as per $\cQ:$ $\E_\cQ[(2q_i-1)^2]$  \\\hline       
%    $K$ &   $\mathbb{Z}_+$ & support size of discrete distribution $\cQ$ \\\hline
    $a$ &  $[T]$ & \textcolor{black}{index for support points of quantized distribution $\lq\cQ$}  \\\hline
%    $q_a$ &  $[0,1]$ & \textcolor{blue}{quality parameter of index $a$} \\\hline
    $\lambda_a$ &  $[0,1]$ & difficulty level of $a$-th support point of $\lq\cQ$  \\\hline
    $\delta_a$ &  $[0,1]$ & \textcolor{black}{probability mass at $\lambda_a$ in $\lq\cQ$} \\\hline
    $\delta_{\rm min}$ & $[0,1]$ & minimum probability mass in $\lq\cQ$: $\min_{a\in[T]} \delta_a$ \\\hline
    $\delta_{\rm max}$ & $[0,1]$ & maximum probability mass in $\lq\cQ$: $\max_{a\in[T]} \delta_a$   \\\hline
    $T$ & $\mathbb{Z}_+$& number of rounds in Algorithm \ref{algo:adapt} \\\hline
	$t$ & $\mathbb{Z}_+$ & index for a round in Algorithm \ref{algo:adapt} \\\hline
	$s_t$ &$\mathbb{Z}_+$ & number of sub-rounds in round $t$ of Algorithm \ref{algo:adapt} \\\hline
	$u$&$\mathbb{Z}_+$ & index for a sub-round   of Algorithm \ref{algo:adapt} \\\hline
  \end{tabular}
\end{center}
\caption{Notations}
\label{tbl:notation}
\end{table}

\section{Main Results under the Adaptive Scenario}
\label{sec:main}

\textcolor{black}{In this section, we present our main results under the adaptive task assignment scenario.}

\subsection{Fundamental limit under the adaptive scenario}

With a slight abuse of notations, we let $\hat{t}(A)$ be a 
mapping from $A\in\{0,+1,-1\}^{m\times n}$ to $\hat{t}(A) \in \{+1,-1\}^m$ 
representing an inference algorithm outputting the estimates of the true labels. We drop $A$ and write only $\hat{t}$ whenever it is clear from the context. 
We let $\F_{\sp}$ be the set of all the prior  distributions on $p_j$ such that the collective  worker quality is $\sp$, i.e. 
\begin{eqnarray}
\F_{\sp} \equiv \left\{\F \, | \, \E_{\F}[(2p_j-1)^2] = \sp\right\}\,.
\end{eqnarray}
We let $\cQ_{\lambda}$ be the set of all the prior  
distributions on $q_i$ such that the collective 
task difficulty  is $\lambda$, i.e. 
\begin{eqnarray}
\cQ_{\lambda} \equiv \left\{\cQ \, \Big| \, \Big( \E_{\cQ}\left[\frac{1}{(2q_i-1)^2}\right]\Big)^{-1} = \lambda  \right\}\,.
\end{eqnarray}
%We are interested in the minimax tradeoff in this set of prior distributions $\cP$'s in $\cP_{\sigma^2}$ and $\cQ$'s in $\cQ_\lambda$.
%achieving the same collective quality parametrized by $\sigma^2$ 
 %achieving the same collective difficulty parametrized by $\lambda$. 
We  consider all task assignment schemes in $\Tau_{\Gamma}$, the set of all task assignment schemes 
that make at most $\Gamma$ queries to the crowd in expectation. 
We prove a lower bound on the standard minimax error rate: the error that is achieved by the best inference algorithm $\t$ using the best adaptive task assignment scheme $\tau \in \Tau_\Gamma$ under a worst-case worker parameter distribution $\F \in \F_{\sigma^2}$ and the worst-case 
task parameter distribution $\cQ \in \cQ_{\lambda}$. 
%We first prove an individual lower bound on a task $i$ 
%conditioned on its  difficulty level $\lambda_i$, and build upon this fundamental bound to prove a lower bound on the average error rate. 
A proof of this theorem is provided in Section \ref{sec:adapt_lb}. 

\begin{theorem} \label{thm:lb_adapt} 
For $\sigma^2<1$,  there exists a positive constant $C'$ such that 
the average probability of error is lower bounded by  
\begin{eqnarray}
\min_{\tau \in \Tau_\Gamma, \t}\;\; \;\;\max_{\cQ \in \cQ_{\lambda}, \F \in \F_{\sp} } \;\;\;\;\frac{1}{m} \sum_{i=1}^m \P[t_i \neq \t_i] 
\;\; \;\;& \geq & \;\; \;\; \frac{1}{4} \,e^{-C' \frac{\Gamma \lambda \sp}{m} } \;,
\label{eq:lb_adapt}
\end{eqnarray}
where $m$ is the number of tasks, 
$\Gamma$ is the expected budget allowed  in $\Tau_\Gamma$, 
$\lambda$ is the collective difficulty of the tasks from a prior distribution $\cQ$ 
defined in \eqref{eq:deflambda}, 
and $\sigma^2$ is the collective reliability of the crowd from 
a prior distribution $\cP$ defined in \eqref{eq:defmu}. 
\end{theorem}
In the proof, 
we provide a proof of a slightly stronger statement in Lemma \ref{lem:lb_adapt_pre}, 
where a similar lower bound holds for not only the worst-case $\cQ$ but  
for all $\cQ\in\cQ_{\lambda}$. 
One caveat is that there is now an extra additive term in the error exponent 
in the RHS of the lower bound that depends on $\cQ$, which is subsumed in the constant term $(1/4)$ for the worst-case $\cQ$ in the RHS of \eqref{eq:lb_adapt}. 
We are assigning $\Gamma/m$ queries per task on average, 
and it is intuitive that the error decays exponentially 
in $\Gamma/m$. 
The novelty in the above analysis is that it characterizes 
how the error exponent 
depends on the $\cP$, which determines the quality of the crowd 
you have in your crowdsourcing platform, and 
$\cQ$, which determines the quality of the tasks you have in your hand. 
If we have easier tasks and reliable workers, the error rate should be smaller. 
Eq.~\eqref{eq:lb_adapt} shows that this is captured by 
the error exponent scaling linearly in $\lambda\sigma^2$.
This gives a lower bound (i.e.~a necessary condition) on  the  budget  required to achieve error at most 
 $\varepsilon$; there exists a constant $C''$ such that if  the total budget is 
\begin{eqnarray}
	\Gamma_\varepsilon \;\; \leq \;\; C''  \frac{m}{\lambda \sigma^2 } \log\left(\frac{1}{\varepsilon}\right)   \;, 
	\label{eq:budget_lb}
\end{eqnarray}
	then no  task assignment scheme (adaptive or not) with any inference algorithm can achieve error less than $\epsilon$.
This recovers the known fundamental limit for standard DS model where all tasks have $\lambda_i=1$ and hence $\lambda=1$ in \cite{KOS14OR}. For this standard DS model, it is known that there exists a constant $C'''$ such that if the 
total budget is less than 
$$	\Gamma_\varepsilon \;\; \leq \;\; C'''  \frac{m}{ \sigma^2} \log\left(\frac{1}{\epsilon}\right)\;, $$
then no task assignment with any inference algorithm can achieve 
error rate less than $\varepsilon$.
For example, consider two 
types of prior distributions
 where in one we have the original DS  tasks with 
$\cQ(q_i=0)= \cQ(q_i=1) = 1/2$ and in the other we have 
$\cQ'(q_i=0)=\cQ'(q_i=1)=\cQ'(q_i=3/4)=\cQ'(q_i=1/4)=1/4$.
We have $\lambda=1$ under $\cQ$ and $\lambda'=2/5$ under $\cQ'$.
Our analysis, together with the matching upper bound in the following section,  shows that one needs $5/2$ times more budget to achieve the same accuracy under the tasks from $\cQ'$. 

\subsection{Upper bound on the achievable error rate}
We present an adaptive task assignment scheme and an iterative inference algorithm that asymptotically achieve an error rate of $C_1e^{-(C_{\delta}/4) (\Gamma/m)\lambda \sigma^2}$,  when the number of tasks 
$m$ grows large and 
the expected budget is increasing as $\Gamma = \Theta(m\log m )$ where 
$C_1 = \log_2(2\delta_{\max}/\delta_{\min})\log_2(2\lambda_{\max}/\lambda_{\min})$ 
and $C_\delta$ is a constant that only depends on $\{\delta_a\}_{a\in[T]}$. This matches the lower bound in \eqref{eq:lb_adapt} when $C_1$ and $C_\delta$ are $O(1)$. 
Comparing it to a fundamental lower bound in Theorem \ref{thm:lb_adapt} establishes the near-optimality of our approach, and 
the sufficient condition  to achieve average error $\varepsilon$
is for the average total budget to be larger than, 
 \begin{eqnarray}
 	\Gamma_\varepsilon \;\; \geq \;\;  \frac{4}{C_\delta} \frac{m}{\lambda \sigma^2 }\log \Big(\frac{C_1}{\varepsilon} \Big)\;. 
	\label{eq:budget_ad_ub}
 \end{eqnarray}

% ----------------------------------------------------------------------------------------------------
Our proposed adaptive approach in Algorithm \ref{algo:adapt} takes as input 
the number of tasks $m$, a target budget $\Gamma$, 
hyper parameter $C_\delta$ to be determined 
by our theoretical analyses in Theorem \ref{thm:ub_adapt}, 
the quantized prior distribution $\lq\cQ$, the statistics $\mu$ and $\sigma^2$ on the worker prior $\cP$. 
The proposed scheme makes at most $\Gamma$ queries in expectation 
to the crowd  and 
outputs  the estimated labels $\hat{t}_i$'s for all the tasks $i\in[m]$.

% ----------------------------------------------------------------------------------------------------
\begin{algorithm}[ht]
\caption{Adaptive Task Assignment and Inference Algorithm}
\label{algo:adapt}
  \begin{algorithmic}[1]
    \REQUIRE number of tasks $m$, 
    allowed budget $\Gamma$, 
    hyper parameter $C_{\delta}$, 
    quantized prior distribution $\{\lambda_a, \delta_a\}_{a \in [T]}$, 
	%$\rho^2$, 
	collective quality of the workers $ \sigma^2$, 
	average reliability $\mu$
    \ENSURE Estimated labels $\{\t_i\}_{i \in [m]}$
    \STATE $M \leftarrow\{1,2,\cdots,m\}$ 
    \STATE $\lq\lambda \leftarrow \Big(\sum_{a \in [T]} (\delta_a/\lambda_a)\Big)^{-1}$
	\FORALL {$t = 1,2, \cdots,T$} 
%	\IF{$M \neq \varnothing$}    
    \STATE $\ell_t \leftarrow (C_{\delta} \lq\lambda\, \,\Gamma) / (m\,\lambda_t)$ , $r_t \leftarrow \ell_t$   \label{linegamma} 
    \STATE $s_t \leftarrow \max\Big\{0, \ceil{\log\Big(\frac{2\delta_t}{\delta_{t+1}}\Big)}\Big\}\I\{t < T\}+  1\, \I\{t = T\}$\label{a0}
	\FORALL {$u = 1,2,\cdots,s_t$}
	\IF{$M \neq \varnothing$} 
	\STATE  $n \leftarrow |M| $ \;,\;\; $k \leftarrow \sqrt{\log|M|}$ 
	 \STATE Draw $E \in \{0,1\}^{|M|\times n} \sim (\ell_t,r_t)\text{-regular random graph}$ \label{a1}
	 \STATE Collect answers $\{A_{i,j} \in \{1,-1\}\}_{(i,j) \in E}$ \label{a2}
%    \STATE $\mathcal{X} \leftarrow \mathcal{X}\I\{t < T\}+  0\, \I\{t = T\}$, $s_t \leftarrow s_t\I\{t < T\}+  1\,\I\{t = T\}$	
	\STATE $\{x_i\}_{i \in M} \leftarrow $ Algorithm \ref{algo:mp} $\big[E, \{A_{i,j}\}_{(i,j) \in E}, \textcolor{black}{k} \big]$ \label{a3}
   	\STATE $\rho^2_{t,u} \leftarrow\text{Algorithm \ref{algo:est} }[E,\{A_{i,j}\}_{(i,j) \in E},\ell_t,r_t]$ \label{a4}
    	\STATE $\mathcal{X}_{t,u} \leftarrow \sqrt{\lambda_t} \mu \ell_t \big((\ell_t-1)(r_t-1)\textcolor{black}{\rho^2_{t,u}\sigma^2 }\big)^{k-1}\I\{t < T\}+  0\, \I\{t = T\}$ \label{a5}
%	\STATE  $\big\{\hat{t}_i = \I\{x_i > \mathcal{X}_{t,u} \} -\I\{x_i < -\mathcal{X}_{t,u} \}\big\}_{i \in M} $ 
	\FOR{$i \in M$}
	\IF{$x_i > \mathcal{X}_{t,u}$} 
	\STATE $\hat{t}_i \leftarrow +1$
	\ELSIF{$x_i <- \mathcal{X}_{t,u}$} 
	\STATE $\hat{t}_i \leftarrow -1$
	\ENDIF
	\ENDFOR
	\STATE $M \leftarrow \{i \in M : |x_i| \leq \mathcal{X}_{t,u} \}$ \label{a6}
	\ENDIF	
	\ENDFOR
%	\ENDIF
	\ENDFOR 
  \end{algorithmic}
\end{algorithm}

\subsubsection{The proposed adaptive approach: overview.}

At a high level, our approach 
works in $T$ {\em rounds} indexed by $t\in[T]$, the support size of the quantized distribution $\lq\cQ$, 
and $s_t$ {\em sub-rounds} at each round $t$, 
where  $s_t$  is chosen by the algorithm in line \ref{a0}. 
In each sub-round, we perform both task assignment and inference, sequentially.  
\textcolor{black}{Guided by the inference algorithm, we permanently label a subset of the tasks and carry over the remaining ones to subsequent sub-rounds.}
{\em Inference} is done in line \ref{a3} to get a 
confidence score $x_{i}$'s on the tasks $i\in M$, 
where $M\subseteq [m]$ is the set of tasks 
that are remaining to be labelled at the current sub-round.  
The {\em adaptive task assignment} of our approach is entirely managed by the choice of this set $M$ in line \ref{a6},  
as only those tasks in $M$ will be assigned new workers in the next sub-round 
in lines \ref{a1} and \ref{a2}. 

At each round, we choose how many responses to collect for each task present in that round as prescribed by our theoretical analysis. 
Given this choice of $\ell_t$, the number of responses collected for each task at round t, 
we repeat the key  inner-loop 
 in  line \ref{a1}-\ref{a6} of Algorithm \ref{algo:adapt}.
 In round $t$ the sub-round is repeated $s_t$ times
 to ensure that sufficient number of `easy' tasks are classified.  
Given a set $M$ of remaining tasks to be labelled, 
the sub-round collects $\ell_t$ response per task on those tasks in $M$ 
and runs an inference algorithm (Algorithm \ref{algo:mp}) to 
give confidence scores $x_i$'s to all $i\in M$. 
Our theoretical analysis prescribes a choice of a threshold $\mathcal{X}_{t,u}$ 
to be used in round $t\in[T]$ sub-round $u\in[s_t]$. 
All tasks in $M$ with confidence score larger than $\mathcal{X}_{t,u}$ are 
permanently labelled as positive tasks, 
and those with confidence score less than  $-\mathcal{X}_{t,u}$ are 
permanently labelled as negative tasks. 
Those permanently labelled tasks are referred to as `classified' and removed from the set $M$. 
The remaining tasks with confidence scores between $\mathcal{X}_{t,u}$ and 
$-\mathcal{X}_{t,u}$ are carried over to the next sub-round. 
The confidence scores are designed such that the sign of $x_i$ provides the estimated true label, and  
we are more confident about this estimated label if
 the absolute value of the score $x_i$ is larger. 
The art is in choosing the appropriate number of responses to be collected for each task $\ell_t$ 
and the threshold $\mathcal{X}_{t,u}$, and our theoretical analyses, 
together with the provided statistics of the prior distribution $\cP$, and the prior quantized distribution $\lq\cQ$  
allow us to choose the ones that achieve a near optimal performance. 

Note that we are mixing inference steps and task assignment steps.
Within each sub-round, we are performing both task assignment and inference. 
Further,  the inner-loop within itself 
uses a non-adaptive task assignment, 
and hence our approach is a series of non-adaptive task assignments with inference in each sub-round. 
However, Algorithm \ref{algo:adapt} is an adaptive scheme, 
where the adaptivity is fully controlled by the set of remaining unclassified 
tasks $M$. 
We are adaptively choosing which tasks to carry over in the set $M$ based on 
all the responses we have collected thus far, 
and we are assigning more workers to only those tasks in $M$. 

Since difficulty levels are  varying across the tasks, it is intuitive to assign fewer workers to easy tasks and more workers to hard tasks. 
Supposing that  we know the difficulty levels $\lambda_i$'s, 
we could choose to assign the ideal number of workers to each task 
according to $\lambda_i$'s. 
%optimizing the lower bound \eqref{eq:lb_ad1} over $\tl_i$'s, it suggests to assign 
%$\tl_i  \simeq \ell ({\lambda}/{\lambda_i})$ 
%workers to the task $i$ with difficulty $\lambda_i$, 
%when given a fixed budget of $\ell$ workers per task on  average. 
%Ignoring the second term that does not depend on $\ell$, we target to assign $\ell_i = (\ell\lambda)/\lambda_i$ workers to a task $i$ having difficulty level $\lambda_i$. 
%In the proof, we will show that assigning $\ell_i$ workers to each task $i$ gives the desired accuracy .  
However, the difficulty levels are not known. 
%Non-adaptive schemes can be arbitrarily worse (see Theorem \ref{thm:thm_lb}). 
%We propose a novel approach that works in multiple rounds. 
The proposed approach starts with a smaller budget in the first round 
classifying easier tasks, and carries over the more difficult tasks to the later rounds where more budget per task will be assigned. 

\subsubsection{The proposed adaptive approach: precise.}
More precisely, 
given a budget $\Gamma$ and the statistics of $\cP$, and the known quantized distribution $\lq\cQ$ 
we know what target probability of error to aim for, say $\varepsilon$, from Theorem 
\ref{thm:ub_adapt}. 
The main idea behind our approach is 
to allocate the given budget $\Gamma$ over multiple rounds appropriately, 
and at each round get an estimate of the labels of the 
remaining tasks in $M$ and also the confidence scores, 
such that with an appropriate choice of the threshold $\mathcal{X}_{t,u}$ 
those tasks we choose to classify  in the current round 
achieve the desired target error rate of $\P[t_i\neq \hat{t}_i \,\big|\,  |x_i|>\mathcal{X}_{t,u}]\leq\varepsilon$. 
As long as this guarantee holds at each round for all classified  tasks, 
then the average error rate will also be bounded by $(1/m)\sum_{i=1}^m \P[t_i\neq \hat{t}_i ]\leq\varepsilon$ when the process terminates eventually. 
The only remaining issue  is 
how many queries are made in total
when this process terminates. We guarantee that in expectation at most $\Gamma$ queries are made under our proposed choices of 
$\ell_t$'s and $\mathcal{X}_{t,u}$'s in the algorithm.

At round zero, we initially put all the tasks in $M = [m]$.  
A fraction of tasks are permanently labelled 
 in each round and the un-labelled ones are 
 taken to the next round. 
At round $t \in \{1,\ldots,T\}$, our goal is to classify a sufficient fraction of those tasks in the $t$-th difficulty  group $\{i\in M\,|\, \lambda_i\in [(1/2)\lambda_t,\lambda_t]\}$ 
 with the desired level of accuracy. 
The art is in choosing the 
right number of responses to be collected per task $\ell_t$ for that round 
and also the right threshold $\mathcal{X}_{t,u}$ on the confidence score, 
to be used in the 
inner-loop in line \ref{a1}-\ref{a6} of Algorithm \ref{algo:adapt}. 
If $\ell_t$ is too low and/or threshold $\mathcal{X}_{t,u}$ too small, 
then misclassification rate will be too large. 
If $\ell_t$ is too large and/or $\mathcal{X}_{t,u}$ is too large, we are wasting our budget 
and achieving unnecessarily high accuracy on those tasks classified in the current round, and not enough tasks will be classified in that round. 
We choose $\ell_t$ 
and $\mathcal{X}_{t,u}$ appropriately to ensure that the misclassification probability is at most $C_1e^{-(C_{\delta}/4) (\Gamma/m)\lambda \sigma^2}$ 
based on our analysis (see \eqref{eq:adapt6}) of the inner-loop.  
We run the identical sub-rounds  $s_t=\max\{0,\lceil \log_2(2\delta_t / \delta_{t+1})\rceil\}$ times to ensure that enough fraction of tasks with difficulty $\lambda_i \in [(1/2)\lambda_t,\lambda_t]$ are classified. 
Precisely, the choice of $s_t$ insures that 
the expected number of tasks with difficulty $\lambda_i \in [(1/2)\lambda_t,\lambda_t]$ remaining unclassified after $t$-th round is at most equal to the number of tasks in the next group, i.e.,  difficulty level $ \lambda_i \in [(1/2)\lambda_{t+1},\lambda_{t+1}]$.

Note that statistically, the fraction of the $t$-th group (i.e.~tasks with difficulty $[\lambda_{t+1},\lambda_t]$) 
that get classified before the $t$-th round %until $t' \in s_t$ set of rounds 
is very small as the threshold set in these rounds is more than their absolute mean message. Most tasks in the $t$-th group  will get classified in round $t$. 
Further, the proposed pre-processing step of binning the tasks 
%the binning of the original given distribution to get $\{\lambda_a, \delta_a\}$ %_{a \in [T]}$ 
ensures that $\ell_{t+1}\geq 2 \ell_t$. %'s are at least $(1/2)$ apart. 
This ensures that the total extraneous budget spent on the $t$-th group of tasks %in $\sum_{t' = 1}^{t-1} s_t$ rounds 
 is not more than a constant times the allocated budget on those tasks. 

The main algorithmic component is the inner-loop  in 
line \ref{a1}-\ref{a6} of Algorithm \ref{algo:adapt}. 
For a choice of the (per task) budget $\ell_t$, 
we collect responses according to a $(\ell_t,r_t=\ell_t)$-regular random graph 
on $|M|$ tasks  and $|M|$ workers. 
The leading eigen-vector of the non-backtracking operator on this bipartite graph, weighted 
by the $\pm1$ responses reveals a noisy observation of the true class and the difficulty levels of the  tasks. 
Let $x \in\reals^{|M|}$ denote this top left eigenvector, computed as per the message-passing algorithm of Algorithm \ref{algo:mp}. 
Then the $i$-th entry $x_i$ asymptotically converges in the large number of tasks $m$ limit to a 
Gaussian random variable with mean proportional to the difficulty level $(2q_i-1)$, 
with mean and variance specified in Lemma \ref{lem:gaussian}. 
This non-backtracking operator approach to crowdsourcing was first introduced in  \cite{KOSnips11} for the standard DS model. 
 We generalize their analysis to this generalized DS model in 
Theorem \ref{thm:main_thm} for finite sample regime, and further give a sharper characterization based on central limit theorem in the asymptotic regime (Lemma \ref{lem:gaussian}).
For a detailed explanation of Algorithm \ref{algo:mp} and its analyses, we refer to 
Section \ref{sec:nonad_ub}. 

\subsubsection{Justification of the choice of $\ell_t$ and $\mathcal{X}_{t,u}$.} 
The main idea behind our approach is to allocate a target budget to each $i$-th task according to its quantized difficulty $\lambda_t$ where $t$ is such that $\lambda_i \in [(1/2)\lambda_t,\lambda_t]$. Given a total budget $\Gamma$ and the quantized  distribution $\lq\cQ$ which gives the  collective difficulty of tasks $\lambda$ (line $2$, Algorithm \ref{algo:adapt}), we target to assign $(\lq\lambda/\lambda_t)(\Gamma/m)$ workers to a task of quantized difficulty $\lambda_t$. This choice of the target budget is motivated from the proof of the lower bound Theorem \ref{thm:lb_adapt}. If we had identified the tasks with respect to their difficulty then the near-optimal choice of the budget that achieves the lower bound is given in \eqref{eq:opt_ell}. Our target budget is a simplified form of the near-optimal choice and ignores the constant part that does not depend upon the total budget. 
This choice of the budget would give 
the equal probability of misclassification 
for the tasks of varying difficulties.  We refer to this error rate as the desired probability of misclassification.
As we do not know  which tasks belong to 
which quantized difficulty group $\lambda_t$, a factor of $1/C_\delta$  is needed to compensate for the  extra budget needed to 
infer those difficulty levels. 
 This justifies our choice of budget in line $4$ of the Algorithm \ref{algo:adapt}.
 
%However, we overcome this challenge with a constant loss in the effective total budget. 
%Since we do not know difficulties of the tasks and our algorithm in various rounds attempts 
%to identify them, it needs $1/C_\delta$ times of total budget to assign the target budget to 
%each task. This justifies our choice of budget in line $4$ of the Algorithm \ref{algo:adapt}.

From our theoretical analysis of the inner loop, 
we know the probability of misclassification for a task that belongs to difficulty group $\lambda_t$ as a function of the classification threshold $\mathcal{X}$ and the budget that is assigned to it. Therefore, in each round we set the classification threshold $\mathcal{X}_{t,u}$ such that even the possibly most difficult task 
 achieves the desired probability of misclassification. This choice of $\mathcal{X}_{t,u}$ is provided in line $13$ of Algorithm \ref{algo:adapt}.

%Therefore, in the first round, $t=1$, we assign each task the target budget of the least difficult task, the ones that have difficulty $\lambda_{1}$, and set the classification threshold $\mathcal{X}$ such that the most difficult tasks, the ones that have difficulty $\lambda_{T}$, achieve the desired probability of error. These choices ensure that in expectation at least half of the tasks that have difficulty $\lambda_{1}$ get classified and also any task that gets classified achieves the desired probability of error. We repeat the task assignment and classification with these choices of the budget and the threshold $\mathcal{X}$ for $s_1$ sub-rounds to ensure that the expected fraction of the tasks with difficulty $\lambda_{1}$ remaining un-classified is no more than the original fraction of the tasks with difficulty $\lambda_2$, that is $\delta_2$. 

\subsubsection{Numerical experiments.} 
\begin{figure}[h]
 \begin{center}
	\includegraphics[width=.35\textwidth]{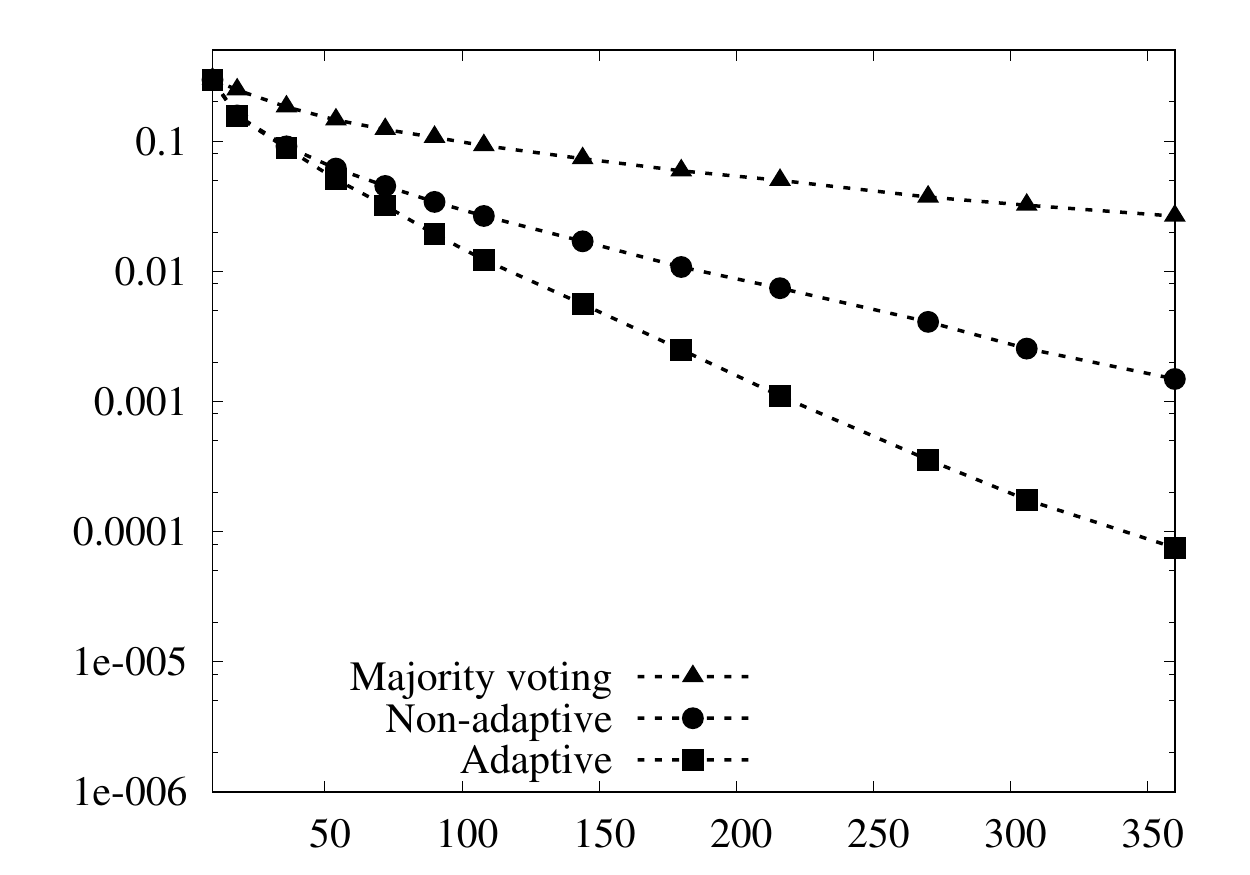}
	\put(-145,114){{\small{probability of error}}}	
	\put(-141,-9 ){\small{number of queries per task $\Gamma/m$}} 
	\includegraphics[width=.35\textwidth]{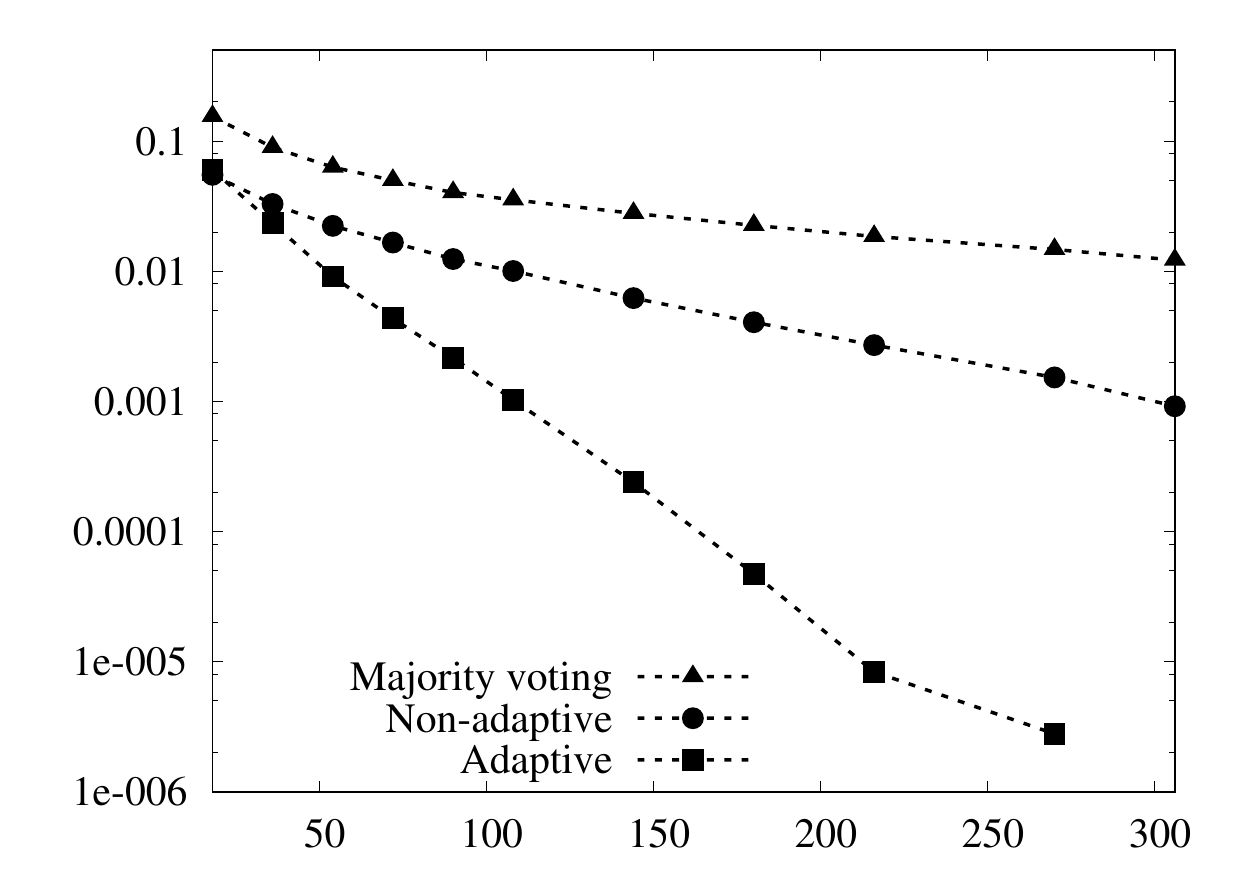}
	\put(-145,114){{\small{probability of error}}}	
	\put(-141,-9 ){\small{number of queries per task $\Gamma/m$}} 
\end{center}
\caption{ Algorithm \ref{algo:adapt} improves significantly over its non-adaptive version and majority voting with a non-adaptive task assignment for 
tasks with $\lambda=1/7$ (left) and $\lambda=4/13$ (right).}
\label{fig:fig1}
\end{figure}

In Figure \ref{fig:fig1}, we compare the performance of our algorithm with majority voting and also a non-adaptive version of our Algorithm \ref{algo:adapt}, where we assign to each task $\ell = \Gamma/m$  
number of workers in one round and set classification threshold $\mathcal{X}_{1,1} = 0$ so as to classify all the tasks (choosing $T=1$ and $s_1=1$). 
Since this performs the non-adaptive inner-loop {\em once}, 
this is a non-adaptive algorithm, and  has been introduced for the standard DS model in \cite{KOS14OR}.

For numerical experiments, 
we make a slight modification to our proposed Algorithm \ref{algo:adapt}.
In the final round, when the classification threshold is set to zero,
we include all the responses collected thus far when running the message passing Algorithm \ref{algo:mp}, and not just the fresh samples collected in that round. This creates dependencies between rounds, which makes the analysis challenging. However, in practice we see improved performance and it allows us to use the given fixed budget efficiently.

We run synthetic experiments with $m =1800$ and fix $n = 1800$ for the non-adaptive version. The crowds are generated from the spammer-hammer model where a worker is a hammer ($p_j=1$) with probability  $0.3$ 
and a spammer ($p_j=1/2$) otherwise.  
In the left panel, we take difficulty level $\lambda_a$ to be uniformly distributed over $\{1,1/4,1/16\}$, that gives $\lambda = 1/7$. In the right panel, we take $\lambda_a = 1$ with probability $3/4$, otherwise we take it to be $1/4$ or $1/16$  with equal probability, 
that gives $\lambda = 4/13$. 
Our adaptive algorithm improves significantly over its non-adaptive version, 
and our main results in Theorems \ref{thm:ub_adapt} and \ref{thm:main_thm} predicts such gain of adaptivity. 
 In particular, for the left panel, the non-adaptive algorithm's error scaling depends on smallest $\lambda_{\rm min}$ that is $1/16$ while for the adaptive algorithm it scales with $\lambda = 1/7$. In the left  figure, it can be seen that the adaptive algorithm requires approximately a factor of $\lambda_{\rm min}/\lambda=7/16$ more queries to achieve the same error as achieved by the non-adaptive scheme. For example, 
 non-adaptive version of Algorithm \ref{algo:adapt} requires
 $\Gamma/m= 360$ to achieve error rate $0.002$, whereas 
 the adaptive approach only requires $180\simeq 360\times7/16=157.5$. 
Quantifying such a gap is one of our main results in Theorems 
\ref{thm:ub_adapt} and \ref{thm:main_thm}. 
 This gap widens in the right panel to approximately $\lambda_{\rm min}/\lambda=13/64$ as predicted. %, and the adaptive algorithm achieves zero error as the number of queries increase. 
For a fair comparison with the non-adaptive version, we fix the total budget to be $\Gamma$ and assign workers in each round until the budget is exhausted, 
such that we are strictly using budget at most $\Gamma$ deterministically. %$C_{\delta}$ is $1$ and $s_t=1$ for $t \in \{1,2,3\}$.   

\subsubsection{Performance Guarantee}
Algorithm \ref{algo:adapt} is designed in such a way that 
 we are not wasting any budget on any of the tasks; 
 we are not getting unnecessarily high accuracy  on easier tasks, 
 which is the root cause of inefficiency for non-adaptive schemes. 
%Further, with the appropriate choice of the constant $C_\delta$, we are guaranteed that this algorithm uses at most $\Gamma$ assignments in expectation. 
In order to achieve this goal, 
the internal parameter $\rho^2_{t,u}$ computed 
in line \ref{a4} of Algorithm \ref{algo:adapt} has to satisfy 
$\rho^2_{t,u}= (1/{|M|}) \sum_{i\in[M]}  \lambda_i $,  
which is the average difficulty 
of the remaining tasks. 
Such a choice is important in choosing the right threshold 
${\mathcal X}_{t,u}$. 

As the set $M$ of  remaining tasks is changing over the course of the algorithm, we need to estimate this value 
in each sub-routine. We provide an estimator of $\rho^2_{t,u}$ in Algorithm \ref{algo:est}  that only uses the sampled responses that are already collected. All numerical results are based on this estimator. However, analyzing the sensitivity of the performance with respect to the estimation error in $\rho^2_{t,u}$ is quite challenging, and for a theoretical analysis, we assume we have access to an oracle that provides the exact value of 
$\rho^2_{t,u}= (1/{|M|}) \sum_{i\in[M]}  \lambda_i $, replacing Algorithm \ref{algo:est}.

\begin{theorem}\label{thm:ub_adapt}
Suppose Algorithm \ref{algo:est} returns the exact value of $\rho^2_{t,u}= (1/{|M|}) \sum_{i\in M }  \lambda_i $. 
With the choice of  $C_{\delta} = ( 4 + \ceil{\log(2\delta_{\max}/\delta_{\min})})^{-1}$, %in Algorithm \ref{algo:adapt},} 
for any given quantized prior distribution of task difficulty $\{\lambda_a, \delta_a\}_{a \in [T]}$ such that $\delta_{\rm max}/\delta_{\rm min}=O(1)$ and 
$\lambda_{\max}/\lambda_{\min}=O(1)$, 
and the budget  $\Gamma = \Theta(m \log m  )$, 
%\textcolor{blue}{there exists a positive numerical constant $C$ such that} 
the expected number of queries made by Algorithm \ref{algo:adapt} is asymptotically bounded by 
%makes at most $m\ell$ queries in expectation,
\begin{eqnarray*}
	\lim_{m\to\infty} \sum_{t\in[{T}],u\in [s_t]} \textcolor{black}{\ell_t\;\E[m_{t,u}]} \;\; \leq\;\; \Gamma\;,
\end{eqnarray*}
 \textcolor{black}{where $m_{t,u}$ is the number of tasks remaining unclassified in  the $(t,u)$ sub-round, and $\ell_t$ is the pre-determined number of workers assigned to each of these tasks in that round.}  
Further, Algorithm \ref{algo:adapt}
returns estimates $\{\t_i\}_{i \in [m]}$ that asymptotically achieve,  
\textcolor{black}{
\begin{eqnarray}
	\lim_{m \rightarrow \infty} \frac{1}{m} \sum_{i=1}^{m} \P[t_i \neq \t_i] 
	\;\; \leq \;\; C_1e^{-(C_{\delta}/4) (\Gamma/m)\lambda \sigma^2}\,,
	\label{eq:ub_adapt}
\end{eqnarray}}
if $(\Gamma/m) \lambda \sigma^2=\Theta(1)$, 
where  $C_1 = \log_2(2\delta_{\max}/\delta_{\min})\log_2(2\lambda_{\max}/\lambda_{\min})$, 
and 
\begin{eqnarray}
	\lim_{m \rightarrow \infty} \frac{1}{m} \sum_{i=1}^{m} \P[t_i \neq \t_i] 
	\;\; = \;\; 0 \,,
\end{eqnarray} 
if $(\Gamma/m) \lambda \sigma^2=\omega(1)$. 
\end{theorem}

A proof of this theorem is provided in Section \ref{sec:adapt_ub}. 
%The constant $C_{\delta}$ can be improved by optimizing over the choice of  $\gamma_a$'s by minimizing the expected number of queries that the algorithm makes. 
In this theoretical analysis, we are considering a family of 
problem parameters $(m,\cQ,\cP,\Gamma)$ in an increasing 
number of tasks $m$. All the problem parameters 
$\cQ$, $\cP$, and $\Gamma$ can vary as functions of $m$.  
For example, 
consider a family of   
$\cQ(q_i)= (1/2)\mathbb{I}(q_i=0) + (1/2) \mathbb{I}(q_i=1)$ independent of 
$m$
and $\cP(p_j) = (1-1/\sqrt{m})\mathbb{I}(p_j=0.5)+ (1/\sqrt{m})\mathbb{I}(p_j=1)$. 
As $m$ grows, most of the workers are spammers giving completely random answers. 
In this setting, we can ask how should the budget grow with $m$, 
in order to achieve a target accuracy of, say, $e^{-5}$?
We have $\lambda=1$ and $\sigma^2=1/\sqrt{m}$, indicating that 
the collective difficulty is constant but collective quality of the workers are 
decreasing in $m$. 
It is a simple calculation to show that 
$C_1=1$ and $C_\delta=1/5$ in this case, and the above theorem 
proves that $\Gamma=100 m^{3/2}$ is sufficient to  
achieve the desired error rate. 
Further such dependence of the budget in $m$ is also necessary, 
as follows from our lower bound in Theorem \ref{thm:lb_adapt}. 

Consider now a scenario where we have tasks with increasing difficulties in $m$. For example, 
$\cQ(q_i) = 
(1/4) \mathbb{I}(q_i=1/2+1/\log m) + (1/4) \mathbb{I}(q_i=1/2-1/\log m) +
(1/4) \mathbb{I}(q_i=1/2+2/\log m) + (1/4) \mathbb{I}(q_i=1/2-2/\log m) 
$ 
and $\cP(p_j) = \mathbb{I}(p_j=3/4)$. 
We have $\lambda = 32/(5(\log m)^2)$ and $\sigma^2=1/4$.
It follows from simple calculations that 
$C_1=2$ and $C_\delta=1/5$. 
It follows that it is sufficient and necessary to have budget scaling in this case as 
$\Gamma = \Theta(m (\log m)^2)$.

For  families of problem parameters for increasing $m$, 
we give  asymptotic performance guarantees. 
Finite regime of $m$ is challenging as our analysis 
relies on a version of central limit theorem and 
the resulting asymptotic distribution of the score value $x_i$'s. 
However, the numerical simulations in Figure \ref{fig:fig1} suggests that 
the improvement of the proposed adaptive approach is significant for   moderate values of $m$ as well.

Our main result in Eq.~\eqref{eq:ub_adapt} gives the sufficient condition of our approach in \eqref{eq:budget_ad_ub}.
Compared to the fundamental lower bound in Theorem \ref{thm:lb_adapt}, 
this proves the near-optimality of our adaptive approach. 
Under the regime considered in Theorem \ref{thm:ub_adapt},  
it is necessary and sufficient to have budget scaling as 
$\Gamma=\Theta((m/(\lambda\sigma^2))\log(1/\varepsilon))$. 

% -----------------------------------------------------------------------------------------------------------------

\section{Analysis of the inner-loop  and the minimax error rate under the non-adaptive scenario}
\label{sec:nonad_ub}

In this section, we provide the 
analysis of the non-adaptive task assignment and inference algorithm in 
the sub-routine  
 in  line \ref{a1}-\ref{a6} of Algorithm \ref{algo:adapt}. 
To simplify the notations, we 
consider the very first instance of the sub-round
where we have a set $M=[m]$ of tasks to be labelled, 
and all the subsequent subroutines will follow similarly up to a change of notations. 
Perhaps surprisingly, we show that 
this inner-loop itself achieves near optimal performance for {\em non-adaptive schemes}.  
We show that $\Gamma = O((m/(\lambda_{\rm min}\sigma^2))\log(1/\varepsilon))$ is sufficient to achieve 
a target probability of error $\varepsilon>0$ in Theorem \ref{thm:main_thm}. 
We show this is close to optimal by comparing it to a necessary condition 
that scales in the same way in Theorem \ref{thm:thm_lb}. 
First, here is the detailed explanation of the inner-loop. 

\bigskip
\noindent
{\bf Task assignment (line \ref{a1} of Algorithm \ref{algo:adapt}).} 
Suppose we are given a budget of $\Gamma = m \ell$, so that each task 
can be assigned to $\ell$ workers on average. 
Further assume that each worker is assigned $r$ tasks. 
We are analyzing a slightly more general setting than Algorithm \ref{algo:adapt} where 
$r=\ell$ for all instances. 
We follow the recipe of \cite{KOS14OR} 
and use a random regular graph for a non-adaptive task assignment. 
Namely, we know that we need to recruit $n=m\ell/r$ workers in total. 
Before any responses are collected, 
we make all the task assignments for all $n$ workers in advance 
and store it in a bipartite graph $G([m],[n],E)$ 
where $[m]$ are the task nodes, $[n]$ are the worker nodes, 
and $E\subseteq [m]\times[n] $ is the collection of 
edges indicating that task $i$ is assigned to worker $j$ if $(i,j)\in E$. 
This graph $E$ is drawn from a random regular graph with task degree $\ell$ and worker degree $r$. 
Such random graphs can be drawn efficiently, for example, using the configuration model \cite{richardson2008modern}. 

Under the {\em original Dawid-Skene model}, 
\cite{KOS14OR} showed that 
this non-adaptive task assignment achieves the 
minimax optimal error rate when 
labels are estimated using Algorithm \ref{algo:mp}. 
This was surprising, as
adaptive task assignments were shown to have no gain over  
this non-adaptive scheme. 
Under the {\em generalized Dawid-Skene model}, 
we are significantly improving upon this simple non-adaptive scheme by applying this to multiple rounds with adaptive choices in each round on which 
tasks to carry over to the next round. Our adaptive scheme uses this non-adaptive task assignments in the inner-loop repeatedly, making adaptive choices on which tasks are carried over to the next rounds. 

\bigskip\noindent
{\bf Inference algorithm (line \ref{a3} of Algorithm \ref{algo:adapt}).} 
The message passing algorithm of Algorithm \ref{algo:mp}, is 
a state-of-the-art spectral method based on non-backtracking operators, first introduced for inference in \cite{KOSnips11}. 
A similar approach has been later applied to other inference problems, e.g. \cite{KMM13,BLM15}. 
This is a message passing algorithm that operates on two sets of messages: 
the task messages $\{x_{i\to j}\}_{(i,j)\in E}$ capturing how likely the task is to be a positive task and 
the worker messages $\{y_{j\to i}\}_{(i,j)\in E}$ capturing how reliable the worker is. 
Consider a data collected on $m$ tasks and $n$ workers 
such that $A\in\{0,+1,-1\}^{m\times n}$ 
under the non-adaptive scenario with task assigned according to 
a random regular graph $E$ of task degree $\ell$ and worker degree $r$.
In each round, all messages are updated as 
\begin{eqnarray}
	\label{eq:update1}
	x_{i\to j} &=& \sum_{j'\in W_i \setminus j} A_{ij'} y_{j'\to i}\;, \text{ and } \\
	 	y_{j\to i} &=& \sum_{i'\in T_j \setminus i} A_{i'j} x_{i'\to j}\;, 
		\label{eq:update2}
\end{eqnarray}
where $W_i \subseteq[n]$ is the set of workers assigned to task $i$, 
and $T_j \subseteq [m]$ is the set of workers assigned to worker $j$.  
The first is taking the weighted majority according to how reliable each worker is, and 
the second is updating the reliability according to how many times the worker agreed with what we believe. 
After a prefixed $k_{\rm max}$ iterations, we provide a confidence score by 
aggregating the messages at each task node $i\in[m]$: 
\begin{eqnarray}
	\label{eq:update3}
	x_{i} &=& \sum_{j'\in W_i } A_{ij'} y_{j'\to i}\;. 
\end{eqnarray}
The precise description is given in Algorithm \ref{algo:mp}. 
Perhaps surprisingly, this algorithm together with the random regular task assignment achieve the minimax optimal error rate among all non-adaptive schemes. This will be made precise in the upper bound in Theorem \ref{thm:main_thm} 
and a fundamental lower bound in Theorem \ref{thm:thm_lb}. 
An intuitive explanation of why this algorithm works is provided in Section 
\ref{sec:spectral} via spectral interpretation of this approach. 

\begin{algorithm}[ht]
\caption{Message-Passing Algorithm }
\label{algo:mp}
  \begin{algorithmic}[1]
    \REQUIRE $E \in \{0,1\}^{|M| \times n}$, $\{A_{ij} \in \{1,-1\}\}_{(i,j) \in E}$, $k_{\max}$
    %\ENSURE Estimate $\{\t^{(k)}_i = {\rm{sign}(x_i)}\}_{i \in [m]}$
	\ENSURE $\{x_i \in \reals\}_{i \in [|M|]}$    
    \FORALL {$(i,j) \in E$}
    \STATE Initialize $y_{j \rightarrow i }^{(0)}$ with a Gaussian random variable $Z_{j \rightarrow i } \sim \N(1,1)$
    \ENDFOR
	\FORALL {$k = 1, 2,\cdots,k_{\max}$}
	\FORALL {$(i,j) \in E$}
	\STATE $x_{i \rightarrow j}^{(k)} \leftarrow \sum_{j' \in  W_i \setminus j} A_{ij'} y_{j' \rightarrow i}^{(k-1)}$
	\ENDFOR
	\FORALL {$(i,j) \in E$}
	\STATE $y_{j \rightarrow i}^{(k)} \leftarrow \sum_{i' \in T_j \setminus i} A_{i'j} x_{i' \rightarrow j}^{(k)}$
	\ENDFOR
	\ENDFOR
	\FORALL {$i \in [m]$}
	\STATE $x_i^{(k_{\rm max})} \leftarrow \sum_{j \in W_i} A_{ij} y_{j \rightarrow i}^{(k_{\rm max}-1)}$
	\ENDFOR
    %\STATE Output: $\hT = \sum_{q\in[\r]}  \sq (\uq \otimes \uq \otimes \uq)$
  \end{algorithmic}
\end{algorithm}

\subsection{Performance guarantee}

%  The following quantities are fundamental in capturing the dependence of the minimax rate on the distribution of task difficulties  and worker reliabilities: 
%\begin{eqnarray} 
%% \lambda \equiv \bigg(\sum_{a \in [K]} \frac{\delta_a}{\lambda_a} \bigg)^{-1}\,, \qquad \sq \equiv \E[(2q_i-1)^2] = \E[\lambda_i] , \text{ and } \qquad	\sp \equiv \E[(2p_j-1)^2] \;. \label{eq:defbeta}
%	 \sq \equiv \E_\cQ[(2q_i-1)^2] \;. \label{eq:defbeta1}
%\end{eqnarray}
%Let $n$ denote the total number of workers used, and $T_j$ denote the set of all tasks assigned to worker $j \in [n]$ and $W_i$ denote the set of all workers assigned to task $i \in [m]$ until the adaptive task assignment scheme has terminated.
%

%In this section, we provide upper bound on the error rate that the non-adaptive version of Algorithm \ref{algo:adapt} achieves. 
%Consider a non-adaptive version of  our approach where we apply it for one round 
%using an $(\ell,r)$ random regular graph, where $\ell$ is the given budget. Naturally,  the classification threshold is set to $\mathcal{X}_{t,u} = 0$ so as to classify all the tasks. 

For this non-adaptive scenario, 
we provide a sharper upper bound on the achieved error, 
that holds for all (non-asymptotic) regimes of $m$. 
%We can analyze error in this setting for any finite $m$. 
Define $\sigma_k^2$ as 
\begin{eqnarray}
 \sigma_k^2 &\equiv& \frac{2\sp }{\mu^2 \big(\hat{\ell}\hat{r}(\sq\sp)^2\big)^{k-1}} + 3\bigg( 1 + \frac{1}{\hat{r}\sq\sp} \bigg) \frac{1 - 1/\big(\hat{\ell}\hat{r}(\sq\sp)^2\big)^{k-1}}{1 - 1/ \big(\hat{\ell}\hat{r}(\sq\sp)^2\big)} \;,
\end{eqnarray}
where $\hat{\ell}=\ell-1$, $\hat{r}=r-1$, 
$\mu=\E_\cP[2p_j-1]$, 
$\sigma^2 = \E_\cP[(2p_j-1)^2]$, and $\rho^2 = \E_\cQ[(2q_i-1)^2]$. 
%With this, we prove the following upper bound on the probability of error for each task conditioned on 
% its difficulty level $\lambda_{i}$.  
This captures  the effective variance in the sub-Gaussian tail of the messages $x_i$'s after $k$ iterations of Algorithm \ref{algo:mp}, as shown in the proof of the following theorem in Section \ref{sec:nonadapt_ub}.

\begin{figure}[h]
 \begin{center}
	\includegraphics[width=.35\textwidth]{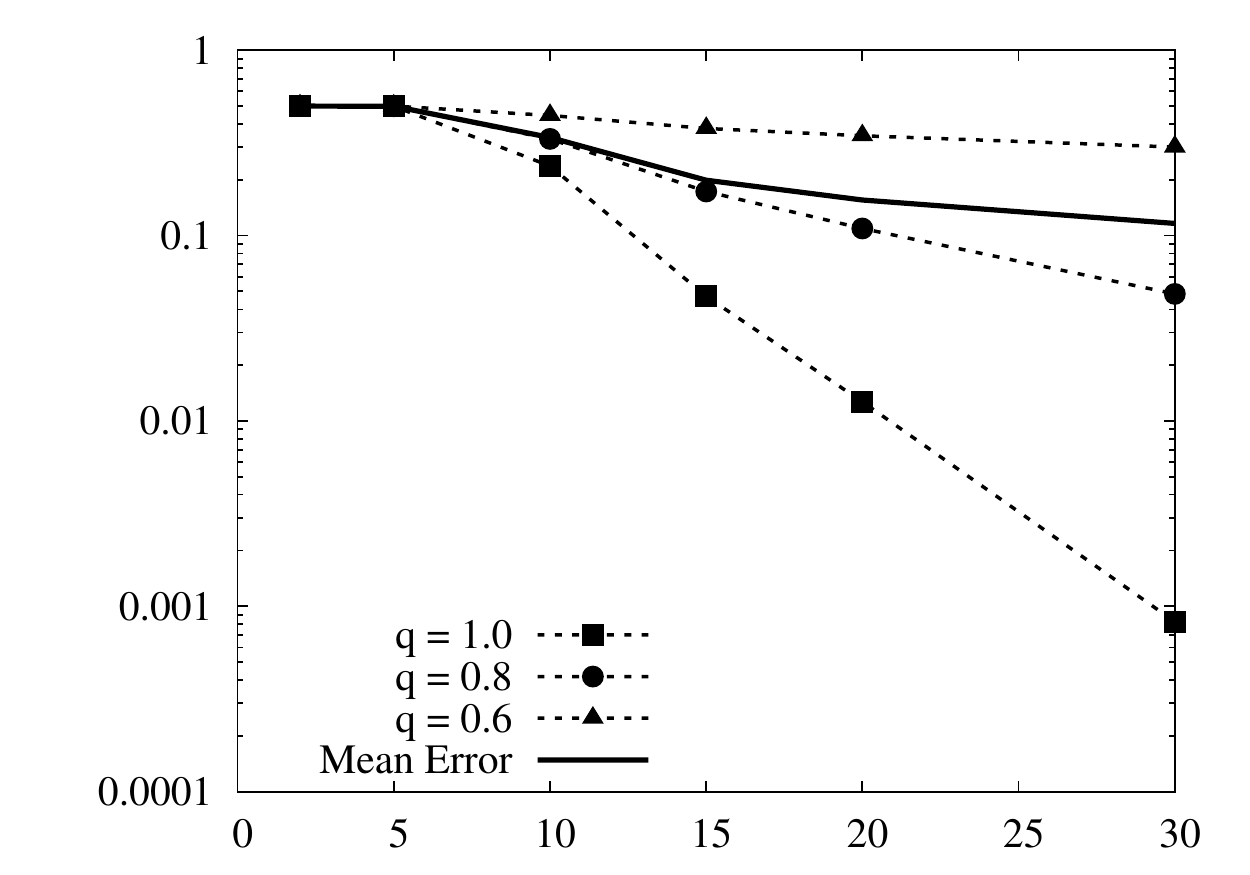}
	\put(-145,120){{\small{probability of error}}}	
	\put(-141,-7 ){\small{number of queries per task $\ell$}} 
	\includegraphics[width=.35\textwidth]{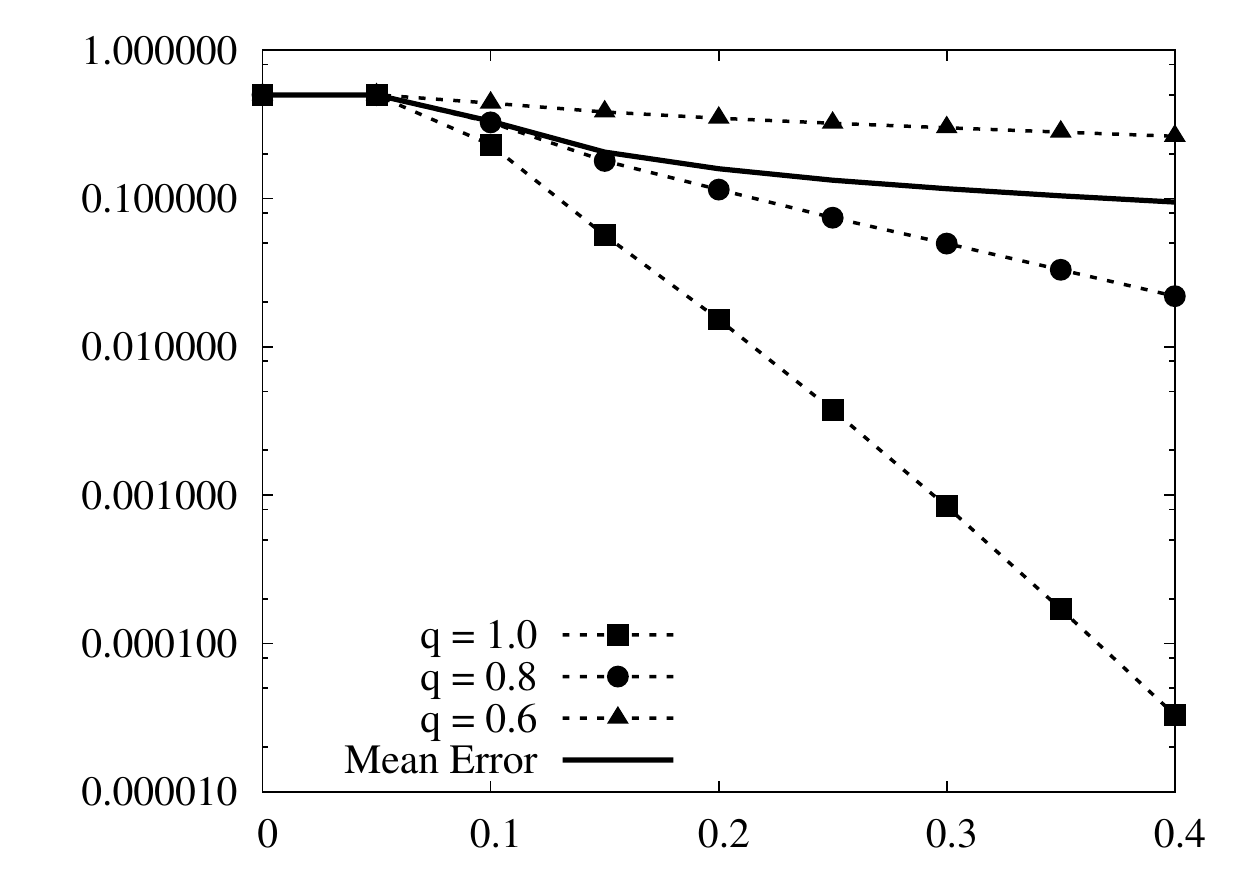}
	\put(-145,120){{\small{probability of error}}}	
	\put(-111,-7 ){\small{crowd quality $\sp$}} 
	\caption{Non-adaptive schemes suffer as average error is dominated by difficult tasks. Dotted lines are error achieved by those tasks with the same quality $q_i$'s, and the overall average error in solid line eventually has the same slope as the most difficult tasks with $q_i=0.6$. 
	}
	\label{fig:fig0}
\end{center}
\end{figure}

%when we run $k$ iterations of our inference algorithm with $(\ell,r)$-regular assignments on $m$ tasks with collective difficulty level being $\sq$ using a crowd with collective quality $\sp$. \\
\begin{theorem} \label{thm:main_thm}
	For any $\ell > 1$ and $r > 1$, suppose  $m$ tasks are assigned according to a random $(\ell,r)$-regular graph 	drawn from the configuration model. If $\mu > 0$, $\l\r \rho^4 \sigma^4 > 1$, and $\r\sq > 1$, 
	then for any $t \in \{\pm 1\}^m$, the estimate $\hat{t}_i^{(k)} = {\rm sign}(x_i^{(k)})$ 
	after $k$ iterations of Algorithm \ref{algo:mp} achieves
	\begin{eqnarray}
	\P\big[t_{i} \neq \hat{t}_{i}^{(k)} \big| \lambda_{i}\big]  \;\;&\leq&\;\;   e^{-\ell\sp\lambda_i/ (2\sigma_k^2)} + \frac{3\ell r }{m}(\hat{\ell}\hat{r})^{2k-2}.
	\end{eqnarray}
\end{theorem}
Therefore, the average error rate is bounded by 
\begin{eqnarray} 
 \frac{1}{m}\sum_{i = 1}^m \P[t_i \neq \hat{t}_i^{(k)}]  \;\;& \leq &\;\; \E_{\cQ}\bigg[e^{\frac{-\ell\sp\lambda_i}{2\sigma_k^2}}\bigg] 
+ \frac{3\ell r }{m}(\hat{\ell}\hat{r})^{2k-2}. \label{eq:average}
\end{eqnarray}
The second term, which is the probability that the resulting $(\ell,r)$-regular random graph is not locally tree-like, can be made small for large $m$ as long as $k=O(\sqrt{\log m})$ (which is the choice we make in Algorithm \ref{algo:adapt}). 
Hence, the dominant term in the error bound is the first term.
% that is the error decays mainly as $e^{- \ell\sp\lambda_i}$.
Further, when we run our algorithm for large enough numbers of iterations, $\sigma_k^2$ converges linearly to a finite limit $\sigma_{\infty}^2 \equiv \lim_{k \rightarrow \infty} \sigma_k^2$ such that
\begin{eqnarray}
\sigma_{\infty}^2 \;\; =\;\;  3\Big(1 + \frac{1}{\r\sq\sp } \Big) \frac{(\l\r\sq\sp)^2} {(\l\r\sq\sp)^2 - 1} \;, 
\end{eqnarray} 
which  is upper bounded by a constant for large enough $\hat{r} \rho^2 \sigma^2$ and $\hat{\ell} $,  for example $\hat{r} \rho^2 \sigma^2\geq 1 $ 
and $\hat{\ell} \geq 2$.
%Hence, for a wide range of parameters, the average error in \eqref{eq:average} is dominated by 
%$\E_{\cQ}\big[e^{{-\ell\sp\lambda_i}/{2\sigma_k^2}}\big] = \sum_a \delta_a e^{-C \ell \sigma^2 \lambda_a}$ for some constant $C$. 
%When all $\delta_a$'s are strictly positive, the error is dominated by the difficult tasks with $\lambda_{\rm min}=\min_a \lambda_a$, as illustrated in Figure \ref{fig:fig0}. 
Hence, for a wide range of parameters, the average error in \eqref{eq:average} is dominated by 
$\E_{\cQ}\big[e^{{-\ell\sp\lambda_i}/{2\sigma_k^2}}\big]$. 
When the fraction of tasks with worst-case difficulty $\lambda_{\min}$ is strictly positive, the error is dominated by them as illustrated in Figure \ref{fig:fig0}. 
Hence, it is sufficient to have budget 
\begin{eqnarray}
	\Gamma_\varepsilon  \;\; \geq \;\; \frac{C'' m}{\lambda_{\rm min} \sigma^2} \log(1/\varepsilon)\;, 
\end{eqnarray}
 to achieve an average error of $\varepsilon>0$. Such a scaling is also necessary as we show in the next section. 
% Recall that our adaptive algorithm precisely addresses this issue of different scaling in error in tasks with varying levels of difficulty. For each task $i$, we target to achieve the same error rate by assigning $\ell_i = (\ell\lambda)/\lambda_i$.

This is further illustrated in Figure \ref{fig:fig0}.
The error decays exponentially in $\ell$ and $\sp$ as predicted, but the rate of decay crucially hinges on the individual difficulty level of the task being estimated. 
We run synthetic experiments with $m=n=1000$ and the crowds are generated from the spammer-hammer model where 
$p_j=1$ with probability $\sigma^2$ and $p_j=1/2$ with probability $1-\sigma^2$, where the choice of this probability is chosen to match the 
collective difficulty $\sigma^2=\E[(2p_j-1)^2]$.
We fix $\sigma^2=0.3$ and vary $\ell$ in the left figure and 
fix $\ell=30$ and vary $\sigma^2$ in the right figure. 
We let $q_i$'s take values in $\{0.6,0.8,1\}$ with equal probability 
such that $\rho^2=1.4/3$.
The error rate of each task grouped by their difficulty is plotted in the dashed lines, matching  
predicted $e^{-\Omega(\ell \sp (2q_i-1)^2)}$. 
The average error rates in solid lines are dominated by those of the difficult tasks, which is a universal drawback for all non-adaptive schemes.

% ---------------------------------------------------------------------------------------------------------------------------------------

\subsection{Fundamental limit under the non-adaptive scenario}
\label{sec:nonadap}

Theorem \ref{thm:main_thm} implies that 
it suffices to assign $\ell \geq (c/(\sp\lambda_i))\log(1/\varepsilon)$ workers
to achieve an error smaller than $\varepsilon$ for a task $i$. 
We show in the following theorem that this scaling is also necessary 
when we consider all {\em non-adaptive} schemes. 
Even the best non-adaptive task assignment with the best inference algorithm 
still required budget scaling in the same way. 
Hence, applying {\em one round} of Algorithm %\ref{algo:mp} 
\ref{algo:adapt}  (which is a non-adaptive scheme) 
is near-optimal in the non-adaptive scenario compared to a minimax rate 
where  
% in terms of the average number of workers assigned to each task, in a minimax scenario where we consider the best task assignment scheme with the best inference algorithm, and 
the nature chooses the worst distribution of worker $p_j$'s among the set of distributions with the same $\sp$.  
We provide a proof of the theorem in Section \ref{sec:nonadapt_lb}.

\begin{theorem} \label{thm:thm_lb} \textcolor{black}{
There exists a positive constant $C'$ and a distribution $\F$ of workers  with average reliability $\E[(2p_j-1)^2]=\sp$ s.t. when $\lambda_i <1 $, if the number of workers assigned to task $i$ by any non-adaptive task assignment scheme is less than $(C'/(\sp\lambda_i))\log(1/\epsilon)$, then no algorithm can achieve conditional probability of error on task $i$ less than $\epsilon$ for any $m$ and $r$.
% under the worst-case worker distribution.
}
\end{theorem}

For formal comparisons with the upper bound, consider a case where 
the induced distribution on task difficulties $\lambda_i$'s, $\lt\cQ$, is same as its quantized version $\lq\cQ$
such  that $\lt\cQ(\lambda_i) = \sum_{a=1}^T \delta_a {\mathbb I}_{(\lambda_i=\lambda_a)}$.
Since in this non-adaptive scheme, task assignments are done a priori, 
there are on average $\ell$ workers assigned to any task, regardless of their difficulty. 
In particular, 
 if the total budget is less than 
\begin{eqnarray}
	\Gamma_\varepsilon  &\leq& C' \frac{m}{\lambda_{\rm min} \sigma^2} \log \frac{\delta_{\rm min}}{\varepsilon},
	\label{eq:budger_nonad_lb}
\end{eqnarray}
then there will a a proportion of 
at least $\delta_{\rm min}$ tasks with error larger than $\varepsilon/\delta_{\rm min}$, resulting in 
overall average error to be larger than $\varepsilon$ even if the rest of the tasks are error-free.
%then no algorithm can achieve average error less than  $\varepsilon$. 
Compared to the adaptive case in \eqref{eq:budget_lb} (nearly achieved up to a constant factor in \eqref{eq:budget_ad_ub}), the gain of adaptivity is a factor of $\lambda/\lambda_{\rm min}$.
 When $\delta_{\rm min} < \varepsilon$, the above necessary condition is 
 trivial as the RHS is negative. In such a case, the necessary condition
  can be tightened to 
$C' ({m}/{\lambda_a \sigma^2}) \log ({\sum_{b=1}^a \delta_b}/{\varepsilon})$ where $a$ is the smallest integer such that $\sum_{b=1}^a \delta_b > \varepsilon$.

\section{Spectral interpretation of Algorithm \ref{algo:mp} and parameter estimation}
\label{sec:spectral} 

In this section, we give a spectral analysis of Algorithm \ref{algo:mp}, 
which leads to a 
spectral algorithm for estimating $\rho^2$ (Algorithm \ref{algo:est}), to be used in the inner-loop of Algorithm \ref{algo:adapt}. 
This spectral interpretation  provides a natural explanation of how Algorithm \ref{algo:mp} is extracting information and estimating the labels. 
Precisely, we are computing the top eigenvector of a matrix known as 
a weighted non-backtracking operator, via standard power method. 
Note that the above mapping is a {\em linear mapping} from the messages to the messages. 
This mapping, if formed into a $2|E| \times  2|E| $ dimensional matrix $B$ is known as the non-backtracking operator. 
Precisely, for $(i,j),(i',j')\in E $, 
\begin{eqnarray*}
	B_{(i\to j),(j'\to i')} = \left\{ 
	\begin{array}{rl}
		A_{i'j'} & \text{if $j=j'$ and $i\neq i'$ }\;, \\
		A_{i'j'} & \text{if $j\neq j'$ and $i= i'$ }\;, \\
		0 & \text{ otherwise }, 
	\end{array}
	\right.
\end{eqnarray*}
and the message update of Equations \eqref{eq:update1} 
and \eqref{eq:update2} are simply 
\begin{eqnarray*}
	 \begin{bmatrix}
	 x\\
	 y
	 \end{bmatrix}
	\;\; = \;\;B\;
		 \begin{bmatrix}
	 x\\
	 y
	 \end{bmatrix}\;,
\end{eqnarray*}
where $x$ and $y$ denote vectorizations of $x_{i\to j}$'s and $y_{i\to j}$'s.
This is exactly the standard power method to compute the singular vector of 
the matrix $B\in\reals^{2|E|\times 2|E|}$. 

The spectrum, which is the set of eigenvalues of this square but non-symmetric matrix $B$ illustrates 
when and why spectral method might work. 
First consider decomposing the data matrix as 
\begin{eqnarray*}
	A  \;\; =  \;\; \underbrace{\E[A]}_{\text{true signal}} + \underbrace{(A-\E[A])}_{\text{random noise}} \;.
\end{eqnarray*} 
Simple analysis shows that $\E[A|q,p]$, 
where the expectation is taken with respect to the randomness in the graph and also in the responses, 
 is a rank one matrix with spectral norm $\|\E[A|q,p]\| = \sqrt{\ell r \hat{\rho}^2 \hat{\sigma}^2 }$, where 
\begin{eqnarray*}
	\hat{\rho}^2 \; \equiv \; \frac1m \sum_{i=1}^m (2q_i-1)^2 \;,\;\; \text{ and } \;
	\hat{\sigma}^2 \; \equiv \; \frac1n \sum_{j=1}^n (2p_j-1)^2 \;. 
\end{eqnarray*}
This is easy to see as 
$\E[A_{ij}|q,p] = (\ell/n) (2q_i-1)(2p_j-1)  $. 
It follows that 
the expected matrix is $\E[A|q,p] = \sqrt{\ell\,r/(mn)}\,   \sqrt{\hat{\rho}^2 \hat{\sigma}^2 mn} \, uv^T  $, where $u$ and $v$ are norm-one vectors 
with $u_i=(1/\sqrt{\sum_{i'\in[m]}(2q_{i'}-1)^2})(2q_i-1)$ and 
$v_j=(1/\sqrt{\sum_{j'\in[n]}(2p_{j'}-1)^2})(2p_j-1)$.

Also, typical random matrix analyses, such as those in \cite{keshavan2009matrix,KOS13SIGMETRICS}, 
show that the spectral norm (the largest singular value) of the noise matrix $(A-\E[A|q,p])$ is bounded by $C( \ell  r )^{1/4}$ with some constant $C$.
Hence, when the spectral norm 
of the signal is larger then 
that of the noise, i.e.~$\|\E[A|q,p]\| > \|(A-\E[A|q,p])\|$, the top eigenvector of this matrix $A$ corresponds to the true underlying signal, 
and we can hope to estimate the true labels from this top eigenvector. On the other hand, 
if $\|\E[A|q,p]\| < \|(A-\E[A|q,p])\|$, one cannot hope to recover any signal from the top eigenvector of $A$.
This phenomenon is known as the spectral barrier.

This phenomenon is more prominent in the non-backtracking operator matrix $B$. 
Note that $B$ is not symmetric and hence the eigen values 
are complex valued.  
Similar spectral analysis can be applied to  show that 
when we are above the spectral barrier, 
the top eigenvalue is real-valued and concentrated around the mean $\Lambda_1(B) \simeq \sqrt{(\ell-1) (r-1) \hat{\rho}^2 \hat{\sigma}^2}$ and the mode of the rest of the complex valued 
eigenvalues are bounded within a circle of radius: 
$|\Lambda_i(B) | \leq ( (\ell-1)(r-1))^{1/4}$. 
Hence, the spectral barrier is exactly when 
$ \Lambda_1(B)=|\Lambda_i(B) |$
which happens at $(\ell-1)(r-1)\hat{\rho}^4 \hat{\sigma}^4 = 1$, and this  plays a crucial role in the performance guarantee in Theorem \ref{thm:main_thm}.
Note that because of the bipartite nature of the graph we are considering, 
we always have a pair of dominant eigenvalue as 
$\Lambda_1(B) =  \sqrt{(\ell-1) (r-1) \hat{\rho}^2 \hat{\sigma}^2}$ and 
$\Lambda_2(B) = - \sqrt{(\ell-1) (r-1) \hat{\rho}^2 \hat{\sigma}^2}$. 

\begin{figure}[h]
	 \begin{center}
	 \hspace{-1.3cm}
	\includegraphics[width=0.45\textwidth]{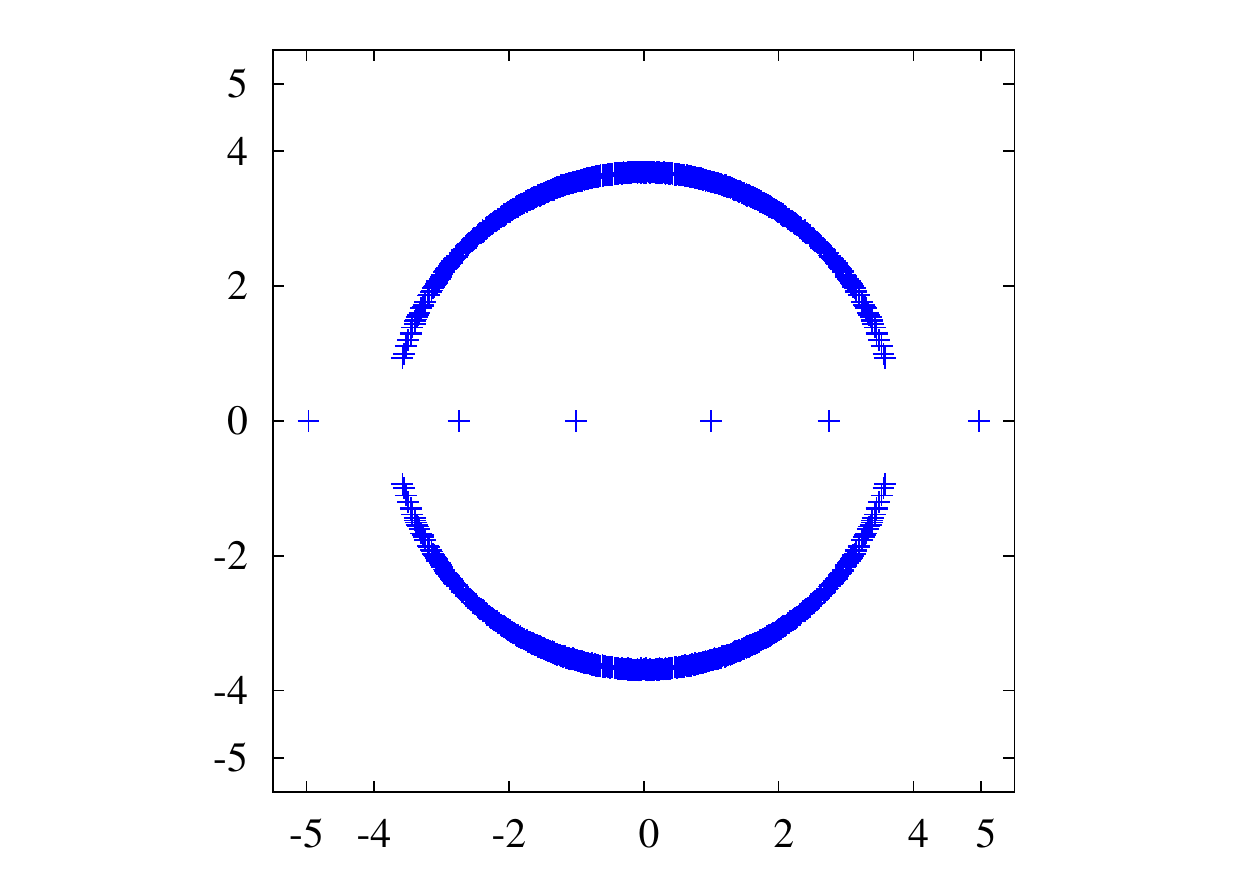} \hspace{-1.6cm}
	\put(-104,-9){real part of $\Lambda_i(B)$}
	\put(-260,75){imaginary part of $\Lambda_i(B)$}
	\includegraphics[width=0.45\textwidth]{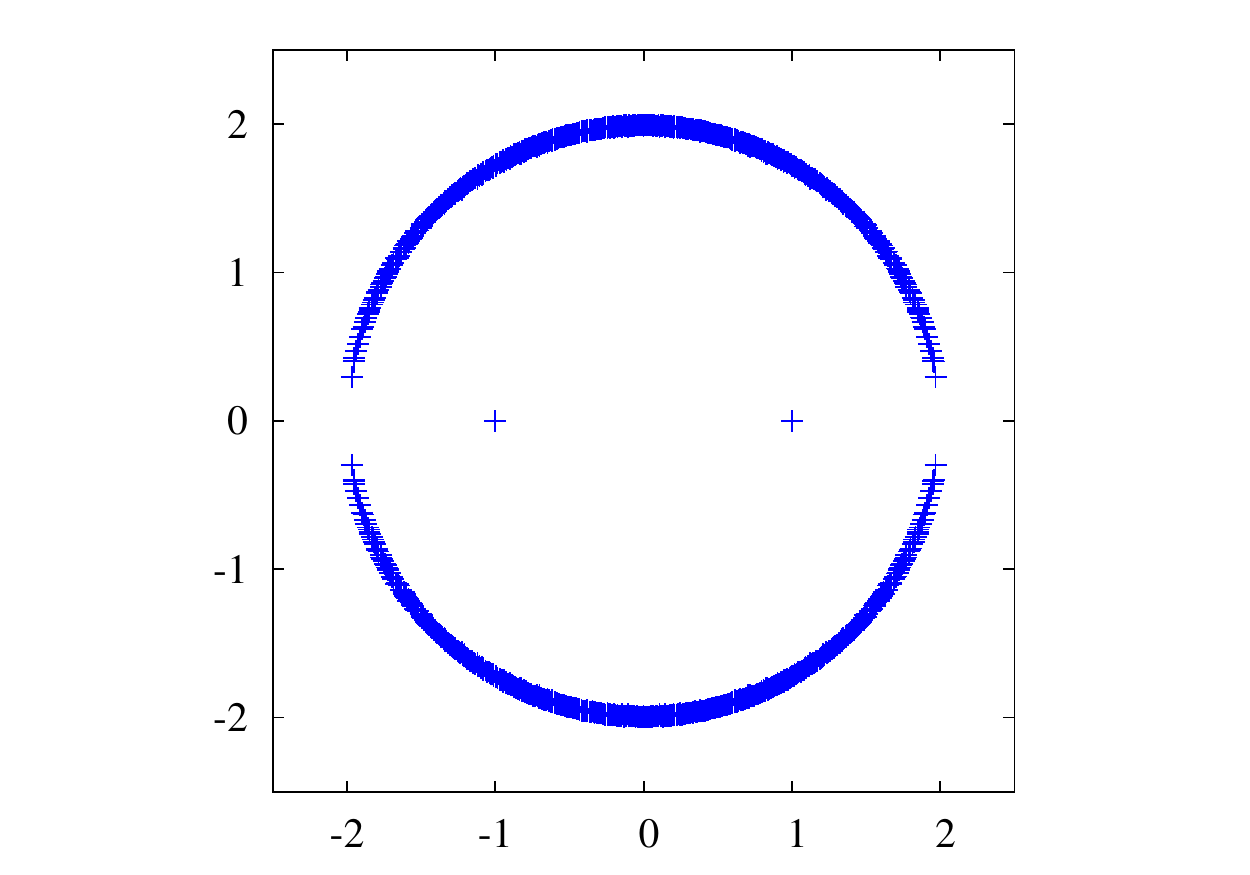} \hspace{-3cm}
	\put(-65,-9){real part of $\Lambda_i(B)$}
	\end{center}
	\caption{Scatter plot of the complex-valued eigenvalues of two realizations of non-backtracking matrix $B$ of the model with $m=n=300$, $\sp = 0.3, \sq = 1.4/3$. On left for $\ell = 15$ and right for $\ell = 5$ which are above and below spectral barrier,  respectively. We can clearly see the two top eigen values at (5,0) and (-5,0)}
\label{fig:spectrum}
\end{figure}

Figure \ref{fig:spectrum} illustrates two sides of the spectral barrier. 
The one on the left shows the scatter plot of the complex valued eigen values of $B$. 
Notice a pair of top eigen values at $\sqrt{0.3 \times (1.4/3) \times 14\times 14} \simeq 5.24$ and 
$- 5.24$ as predicted by the analysis.  
They always appear in pairs, due to the bipartite nature of the graph involved. 
The rest of the spurious eigenvalues are constrained within a circle of radius 
$(14\times 14)^{1/4} \simeq 3.74$ as predicted. 
The figure on the right is when we are below the spectral barrier, since 
the eigenvalue corresponding to the signal is $\sqrt{0.3 \times (1.4/3) \times 4\times 4} \simeq 1.5$
which is smaller than $(4\times 4)^{1/4} = 2$. 
The relevant eigenvalue is buried under other spurious eigenvalues and does not show.

\bigskip\noindent
{\bf Parameter estimation algorithm (line \ref{a4} of Algorithm \ref{algo:adapt}).}
Among other things, this spectral 
interpretation gives an estimator for the problem parameter $\rho^2$, to be used in the inner-loop of Algorithm \ref{algo:adapt}. 
Consider the data matrix $\widetilde{A}$ defined below. 
Again, a simple analysis shows that $\E[\widetilde{A}|q,p]$  is a rank one matrix with $\|\E[\widetilde{A}|q,p]\| = \sqrt{\ell r \hat{\rho}^2 \hat{\sigma}^2}$. 
Since the spectral norm of the noise matrix $\|\widetilde{A} - \E[\widetilde{A}|q,p]\|$ is upper bounded by $C(\ell r)^{1/4}$ for some constant $C$,  
%in the limit of $|M|$, since $\ell = r = \Theta(\log |M|)$, 
we have $\|\widetilde{A}\|/\sqrt{ \ell r \sigma^2} = \sqrt{\hat{\rho}^2 \hat{\sigma}^2/\sigma^2}  + O((\ell r \hat{\rho}^4 \hat{\sigma}^4)^{-1/4})$. 
We know $\ell$ and $r$, and assuming we know $\sigma$, this provides a natural estimator for $\hat{\rho}^2$.  
Note that $\hat{\sigma}^2 = \sigma^2 + O(\log (n) /\sqrt{n})$ with high probability. 
The performance of this estimator is empirically evaluated, as we use this in all our numerical simulations to implement our  
adaptive scheme in Algorithm \ref{algo:adapt}.
% Hence, $\rho^2$ is close to the squared normalized top singular value of $\widetilde{A}$.

\begin{algorithm}[ht]
\caption{ Parameter Estimation Algorithm }
\label{algo:est}
  \begin{algorithmic}[1]
    \REQUIRE assignment graph adjacency matrix $E \in \{0,1\}^{|M| \times n}$, binary responses from the crowd $\{A_{ij} \}_{(i,j) \in E}$, task degree $\ell$, worker degree $r$, worker collective quality parameter $\sigma^2$
    %\ENSURE Estimate $\{\t^{(k)}_i = {\rm{sign}(x_i)}\}_{i \in [m]}$
	\ENSURE estimate $\rho^2$ of $(1/|M|)\sum_{i\in M} \lambda_i$
	\STATE Construct matrix $\widetilde{A}\in\{0,\pm1\}^{|M|\times n }$ such that 
	$$\widetilde{A}_{i,j}=\left\{ \begin{array}{rl} 
	A_{i,j}&\text{, if ${(i,j) \in E}$} \\
	0 & \text{, otherwise}
	\end{array}\right.$$
	for all $i\in [|M|]$, $j\in [n]$.
	\STATE Set $\sigma_1(\widetilde{A})$ to be the top singular value of matrix $\widetilde{A}$ 
	\STATE $\rho^2 \leftarrow \big(\sigma_1(\widetilde{A})/\sqrt{\ell r\sigma^2}\big)^2$
    %\STATE Output: $\hT = \sum_{q\in[\r]}  \sq (\uq \otimes \uq \otimes \uq)$
  \end{algorithmic}
\end{algorithm}

%\begin{algorithm}[ht]
%\caption{ Parameter Estimation Algorithm }
%\label{algo:est}
%  \begin{algorithmic}[1]
%    \REQUIRE $E \in \{0,1\}^{|M| \times n}$, $\{A_{ij} \}_{(i,j) \in E}$, $\ell$, $r$, $\sigma^2$
%    %\ENSURE Estimate $\{\t^{(k)}_i = {\rm{sign}(x_i)}\}_{i \in [m]}$
%	\ENSURE $\rho^2$    
%	\STATE Construct the (weighted) non-backtracking matrix $B\in\{0,\pm1\}^{2|E|\times 2|E| }$ such that 
%	$$B_{(i,j),(i',j')}=\left\{ \begin{array}{rl} 
%	A_{i,a}&\text{, if $j=i'$} \\
%	0 & \text{, otherwise}
%	\end{array}\right.$$
%	for all directed edges $(i,j)\in E$ and $(i',j')\in E$.
%	\STATE Set $\sigma_1(B)$ to be the top singular value of matrix $B$ 
%	\STATE $\rho^2 \leftarrow \sigma_1(B)/\sigma^2$
%    %\STATE Output: $\hT = \sum_{q\in[\r]}  \sq (\uq \otimes \uq \otimes \uq)$
%  \end{algorithmic}
%\end{algorithm}

% ---------------------------------------------------------------------------------------------------------------------------------------

\section{Alternative inference algorithm for the generalized DS model}
\label{sec:algo}
%{\em{Majority Voting.}} The majority voting simply 
%follows what the majority of the workers agree on, i.e. 
%\begin{eqnarray}
%t_i^{{\rm{MV}}} = {\rm{sign}}\big(\sum_{j \in W_i} \A_{ij} \big)\,.
%\end{eqnarray}
%This is the optimal ML estimator when all workers have the same reliability, and is used widely in practice.  
%However, the performance degrades significantly 
%as the qualities of the workers become diverse, in particular, in the case of the spammer-hammer model. 

Our main contribution is a general framework for adaptive crowdsourcing:  
starting with a small-budget, classify  tasks with high-confidence, and then    
gradually increase the budget per round, classifying remaining  tasks. 
If we have other inference algorithms with which we can get reliable 
confidence levels in the  estimated task labels, 
we can replace 
Algorithm \ref{algo:mp}. 
In this section, we propose such a potential candidate  and 
 discuss the  computational challenges involved. 

\begin{figure}[h]
 \begin{center}
	\includegraphics[width=.35\textwidth]{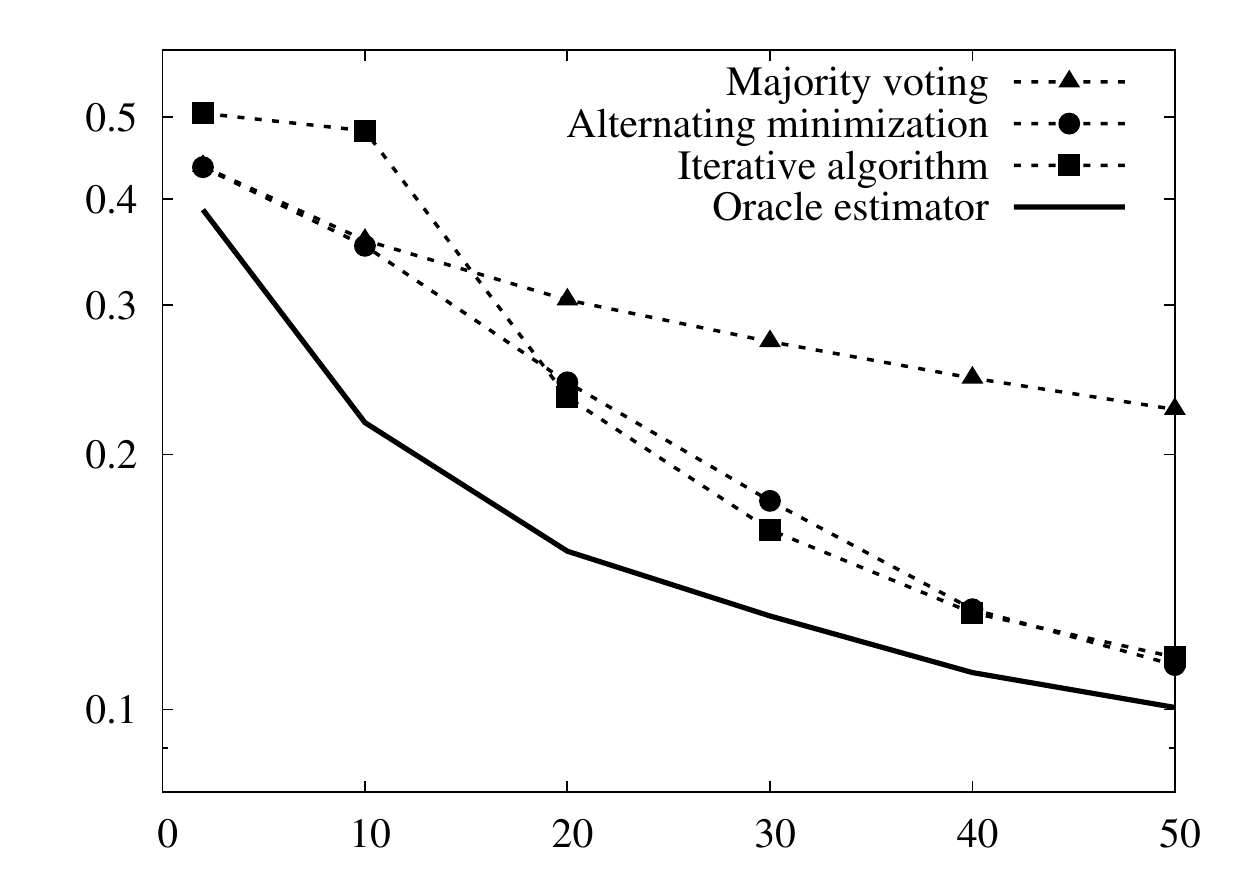}
	\put(-160,117){{\small{probability of error}}}	
	\put(-141,-7 ){\small{number of queries per task $\Gamma/m$}} 
	\hspace{1cm}
	\includegraphics[width=.35\textwidth]{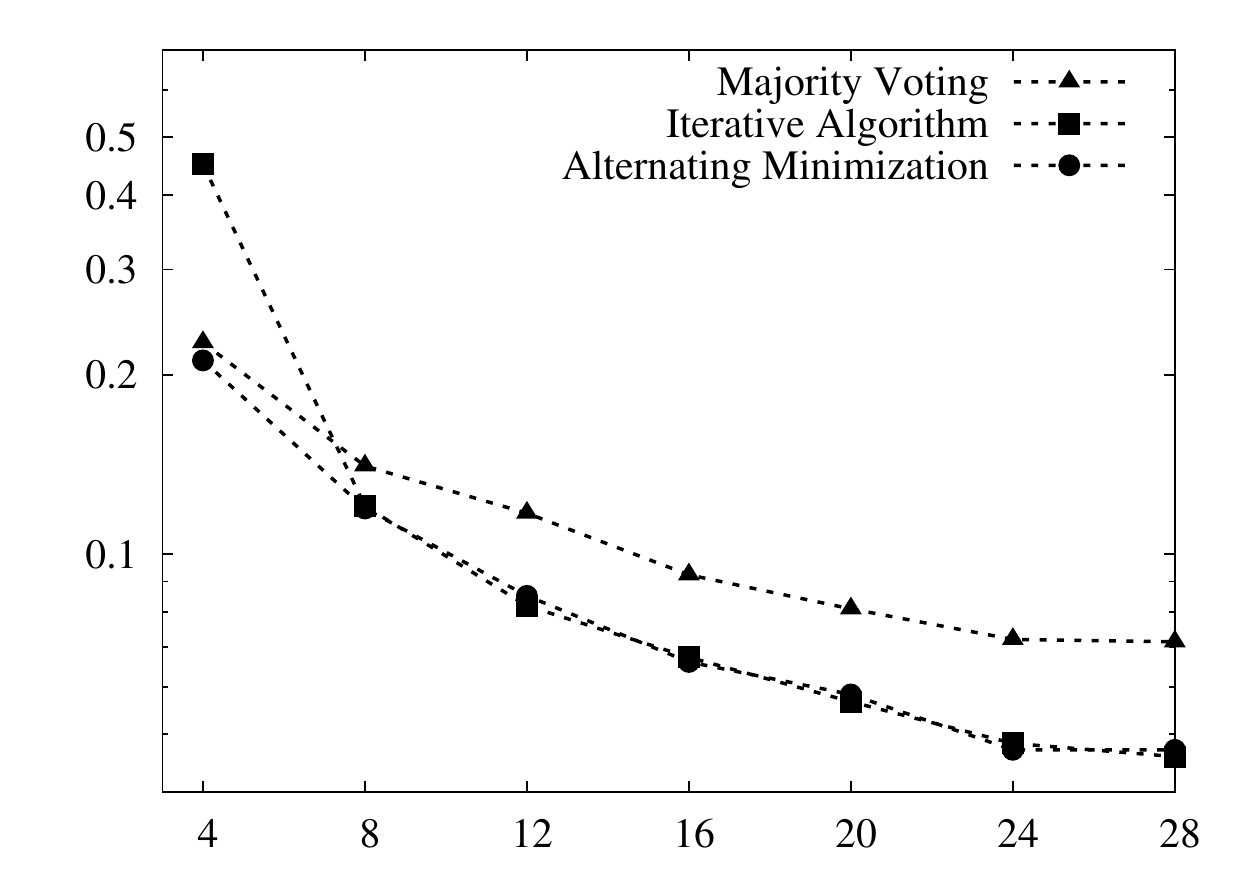}
	\put(-160,117){{\small{probability of error}}}
	\put(-141,-7 ){\small{number of queries per task $\Gamma/m$}} 
\end{center}
\caption{The iterative algorithm improves over majority voting and has similar performance as alternating minimization on both synthetic data (left) and real data from Amazon's Mechanical Turk (right).}
\label{fig:comp}
\end{figure}

Under the original DS model, various standard methods such as Expectation Maximization (EM) and 
Belief  Propagation (BP) provide efficient inference algorithms that also work well in practice \cite{Liu12}.  
However, under the generalized DS model, both approaches fail to give computationally tractable inference algorithms. 
\textcolor{black}{The reason is that both tasks and workers are parametrized by continuous variables, making EM and BP computationally infeasible.} 
In this section, we propose an 
alternative inference algorithm  based on alternating minimization. 
This approach enjoys the benefits of EM and BP, such as seamlessly extending to $k$-ary alphabet labels, while remaining 
computationally manageable.
%We numerically compare the performance with the proposed Algorithm \ref{algo:mp} in  
Figure \ref{fig:comp} illustrates how this alternating minimization performs at least as well as the iterative algorithm (Algorithm \ref{algo:mp}), 
and improves significantly when the budget is critically small, i.e.~only a few 
workers are assigned to each task.

We propose to maximize the  posterior distribution, 
\begin{align}\label{eq:eqgm1}
\P[\q, \p| \A ] \;\; \propto \;\; \prod_{i \in [m]} \P_\cQ[q_i] \; \prod_{j \in [n]} \bigg( \P_\cP[p_j ] 
\prod_{i' \in W_j} \P[A_{i'j}| p_{i'},q_j]\bigg)\,.
\end{align}
Although this function is not concave, 
maximizing over $\q$ (or $\p$) fixing $\p$ (or $\q$) is simple due to the bipartite nature of the graph. 
 Define a function $g:\{\pm1\}\times[0,1]\times[0,1] \to [-\infty,0]$ such that
\begin{align}
g(A_{ij},q_i,p_j) = \begin{cases}
 \log(q_ip_j + \bq_i\bp_j)\;\;\; \text{if}\; A_{ij}=1 \\
 \log(\bq_ip_j + q_j\bp_i) \;\;\; \text{if}\; A_{ij} = -1
 \end{cases}
\end{align}
The logarithm of the joint posterior distribution \eqref{eq:eqgm1} is   
\begin{align}
 \cL(\q,\p | \A  )
\;\; = \;\;\sum_{i \in [m]}\sum_{j \in W_i} g(A_{ij},q_i,p_j) 
 + \sum_{i \in [m]}\log(\P_\cQ[q_i ]) + \sum_{j \in [n]}\log(\P_\cP[p_j ])\,. \label{eq:eq_am1}
\end{align}
With properly chosen prior distributions $\cQ$ and $\cP$, 
in particular Beta priors, it is easy to see that the 
 log likelihood is a concave function of $\p$ for fixed $\q$. 
 The same is true when fixing $\p$ and considering a function over $\q$.  
Further, each coordinate $p_j$ (and $q_i$) can be maximized separately. 
We start with $q_i = |W_i^{+}|/(|W_i^{+}| + |W_i^{-}|)$ and perform alternating minimization on \eqref{eq:eq_am1} with respect to $\q$ and $\p$ iteratively until convergence, where $W_i^{+} = \{j \in W_i: A_{ij} = 1 \}$, $W_i^{-} = \{j \in W_i:A_{ij} = -1 \}$, and $W_i$ is the set of workers assigned to task $i$.   

In Figure \ref{fig:comp}, we compare our algorithm with alternating minimization and majority voting on simulated data and real data. The first plot is generated under the same settings as the first plot of Figure \ref{fig:fig0} 
except that here we use $n = m = 300$ and $\sp = 0.2$. 
It shows that Algorithm \ref{algo:mp} and alternating minimization performs almost the same after the 
spectral barrier, 
 while the proposed Algorithm \ref{algo:mp} 
 fails below the spectral barrier as expected from the spectral analysis of Section \ref{sec:spectral}. 
 For the figure on the left, we choose $\sp = 0.2, \sq = 1.4/3$. 
From the analyses in Section \ref{sec:spectral}, we predict the spectral barrier to be at $\ell = 11$. 
%In simulation results it is around $\ell = 14$.
 %alternating minimization performs better before the phase transition. 
%Note that, implementing alternating minimization is costly as it requires solving $mn$ optimization problems for each iteration. 
In the second plot, we compare all the three algorithms on real data collected from Amazon Mechanical Turk in 
\cite{KOS14OR}. This dataset considers binary classification tasks for comparing closeness in 
human perception of colors; 
three colors are shown in each task 
and the worker is asked to indicate 
``whether the first color is more similar to the second color or the third color." 
This is asked on $50$ of such color comparison tasks and 
$28$ workers are recruited to complete all the tasks. 
We take the ground truth according to which 
color is closer to the first color in pairwise distances in the {\em Lab color} space. 
The second plot shows probability of error of the three algorithms when 
number of queries per task $\ell=\Gamma/m$ is varied. We generated responses for 
different values of $\Gamma/m$ by uniformly sub-sampling. The alternating minimization and iterative algorithm perform similarly. 
However, for very small $\Gamma$, alternating minimization outperforms the iterative algorithm.  

% ---------------------------------------------------------------------------------------------------------------------------------------
\section{Proofs}
\label{sec:proof}

In this section, we provide the proofs of the main technical results. 

\subsection{Proof of Theorem \ref{thm:lb_adapt}} \label{sec:adapt_lb}
In this section, we first prove a slightly stronger result  in Lemma \ref{lem:lb_adapt_pre} 
and prove Theorem \ref{thm:lb_adapt} as a corollary. 
%We first prove a standard minimax lower bound on the average probability of error for a slightly general case. 
Lemma \ref{lem:lb_adapt_pre} is stronger as it is an {\em adaptive} lower bound that holds for {\em all} discrete prior distribution $\cQ$. 
The lower bound in Equation \eqref{eq:lb_lem_adapt} is adaptive in the sense that 
it automatically adjusts for any given $\cQ$ as shown in its explicit dependence in $\delta_{\rm min}$ and $\lambda$.
On the other hand, Theorem \ref{thm:lb_adapt} only has to hold for {\em one} worst-case prior distribution $\cQ$. 

%is stated for the worst-case prior distribution $\cQ \in \cQ_{\lambda}$, we prove it for any discrete distribution $\cQ$ that achieves the collective task difficulty $\lambda$ and has the minimum probability mass $\delta_{\min}$. 
Let $\cQ_{\lambda,\delta_{\rm min}}$ be the set of all discrete prior  
distributions on $q_i$ such that the collective 
task difficulty  is $\lambda$, and the minimum probability mass in it is $\delta_{\min}$, i.e. 
%\begin{eqnarray}\label{eq:Q_lambda_delta}
%\cQ_{\lambda,\delta_{\rm min}} \equiv \left\{{\rm discrete\;} \cQ \, \Big| \, \Big( \E_{\cQ}\left[\frac{1}{(2q_i-1)^2}\right]\Big)^{-1} = \lambda  \;,\; \min_{q_a\in {\rm supp}(\cQ)} \cQ(q_a) =\delta_{\rm min}\right\}\,,
%\end{eqnarray}
\begin{eqnarray}\label{eq:Q_lambda_delta}
\cQ_{\lambda,\delta_{\rm min}} \equiv \left\{{\rm discrete\;} \cQ \, \Big| \, \Big( \E_{\cQ}\left[\frac{1}{(2q_a-1)^2}\right]\Big)^{-1} = \lambda  \;,\; \min_{\lambda_a\in {\rm supp}(\lt\cQ)} \lt\cQ(\lambda_a) =\delta_{\rm min}\right\}\,,
\end{eqnarray}
where $\lt\cQ$ is the induced distribution on $\lambda_a$'s.
We  let $\Tau_{\Gamma}$ be the set of all task assignment schemes that make at most $\Gamma$ queries to the crowd in expectation. 
We prove a lower bound on the standard minimax error rate: the error that is achieved by the best inference algorithm $\t$ using the best adaptive task assignment scheme $\tau \in \Tau_\Gamma$ under a worst-case worker parameter distribution $\F \in \F_{\sigma^2}$ and any 
task parameter distribution $\cQ \in \cQ_{\lambda,\delta_{\rm min}}$. 
Note that instead of maximizing over $\cQ \in \cQ_{\lambda,\delta_{\rm min}}$, 
our result holds for all discrete $\cQ\in \cQ_{\lambda,\delta_{\rm min}}$.
%Note that we prove it for any $\cQ \in \cQ_{\lambda,\delta_{\rm min}}$ not the worst-case $\cQ$.
A proof of this lemma is provided in the following section. 

\begin{lemma} \label{lem:lb_adapt_pre} 
For $\sigma^2<1$,  for any discrete $\cQ \in \cQ_{\lambda,\delta_{\rm min}}$, there exists a positive constant $C'$ such that 
the average probability of error is lower bounded by  
\begin{eqnarray}
\min_{\tau \in \Tau_\Gamma, \t}\;\; \;\;\max_{\F \in \F_{\sp} } \;\;\;\;\frac{1}{m} \sum_{i=1}^m \P[t_i \neq \t_i] 
\;\; \;\;& \geq & \;\; \;\; \frac{1}{2} \,\delta_{\rm min} e^{-C' \frac{\Gamma \lambda \sp}{m} } \;,
\label{eq:lb_lem_adapt}
\end{eqnarray}
where $m$ is the number of tasks, 
$\Gamma$ is the expected budget allowed  in $\Tau_\Gamma$, 
$\lambda$ is the collective difficulty of the tasks from a prior distribution $\cQ \in \cQ_{\lambda,\delta_{\rm min}}$ 
defined in \eqref{eq:Q_lambda_delta}, 
$\sigma^2$ is the collective reliability of the crowd from 
a prior distribution $\cP$ defined in \eqref{eq:defmu}, 
and $\delta_{\rm min}$ is defined in \eqref{eq:Q_lambda_delta}. 
%\textcolor{magenta}{Need to define $\delta_{\rm min}$ rigorously if $\cQ$ is continuous.}
\end{lemma}

Theorem \ref{thm:lb_adapt} follows immediately from Lemma \ref{lem:lb_adapt_pre} as it considers the worst-case $\cQ \in \cQ_{\lambda}$ whereas the Lemma is proved for any discrete $\cQ \in \cQ_{\lambda,\delta_{\rm min}}$. For any given $\lambda$, there exists a discrete distribution $\cQ \in \cQ_{\lambda,\delta_{\rm min}}$, namely a distribution that is supported at two points $q = (1 \pm \sqrt{\lambda})/2$ with equal probability mass of $1/2$. Such a distribution has $\delta_{\min} = 1/2$ and therefore the Theorem \ref{thm:lb_adapt} follows. 

\subsection{Proof of Lemma \ref{lem:lb_adapt_pre}} 
\label{sec:adapt_lb_pre}
Let $W_i \subseteq [n]$ 
denote the (random) set of workers assigned to task $i$ in the end, 
when $n$ (random) number of workers have provided their responses. 
For a task assignment scheme $\tau$, we let 
\begin{eqnarray}
	\ell_{i,q_i}^{(\tau)}(\cQ,\cP) \; \equiv\;  \E[ \, |W_i| | q_i \, ] \;, 
\end{eqnarray}
denote the conditional expectation of number of workers assigned to a task $i$ conditioned on its quality $q_i$.
Let 
\begin{eqnarray}
	\Tau_{\ell_{i,q_i}} \;\equiv\;  \left\{ \tau: \ell_{i,q_i}^{(\tau)}=\ell_{i,q_i}\right\}\;, 
\end{eqnarray}
denote the set of all task assignment schemes that in expectation assign 
$\ell_{i,q_i}$ workers to the $i$-th task conditioned on its quality $q_i$. Further, let 
\begin{eqnarray}
	\Tau_{\{{\ell_{i,q_i}}\}_{i= 1}^m} \;\equiv\;  \left\{ \tau: \left({\{\ell_{i,q_i}\}}_{i=1}^m\right)^{(\tau)} ={\{\ell_{i,q_i}\}}_{i=1}^m\right\}\;,
\end{eqnarray}
denote the set of all task assignment schemes that in expectation assign 
$\ell_{i,q_i}$ workers to each task $i \in [m]$ conditioned on its quality $q_i$. 
The fundamental lower bound crucially relies on the following technical 
lemma, whose proof is provided in the following section. 
\begin{lemma} \label{lem:lb_adapt} 
For any $\sigma^2<1$,  there exists a positive constant $C'$ and a prior distribution $\cP^* \in \cP_{\sigma^2}$ such that for each task $i \in [m]$, 
for all $\cQ$, 
\begin{eqnarray*}
  \min_{\tau \in \Tau_{\ell_{i,q_i}} , \t}\;\;  \P[t_i \neq \t_i | q_i] & \geq & \frac{1}{2} e^{-C'\lambda_i \sp \, \ell_{i,q_i} }\,,
\end{eqnarray*}
where $\lambda_i=(2q_i-1)^2$. 
\end{lemma}
This proves a lower bound on {\em per task} probability of error that decays exponentially with exponent 
scaling as $\lambda_i \sigma^2\ell_{i,\q_i}$. 
The easier the task ($\lambda_i = (2q_i-1)^2$ large), the more reliable the workers are ($\sigma^2$ large), 
and the more workers assigned to that task ($\ell_{i,\q_i}$ large), the smaller the achievable error. 
To get a lower bound on the minimax {\em average} probability of error, 
where the error probability is over the randomness in the latent variables from 
$(\cQ,\cP)$ and the randomness in the task assignment scheme $\tau$ and the responses $A$, 
we have,
\begin{eqnarray}
&&\min_{\tau \in \Tau_\Gamma, \t}\;\; \max_{  \F \in \F_{\sp} } \frac{1}{m} \sum_{i=1}^m \P_{(\cP,\cQ,\tau)} [t_i \neq \t_i] \nonumber\ \\
&=&  \min_{\tau \in \Tau_\Gamma, \t}\;\; \max_{  \F \in \F_{\sp} } \frac{1}{m} \sum_{i=1}^m \E_{ q_i\sim \cQ } \big[ \, \P_{(\cP,\cQ,\tau)}[t_i \neq \t_i  | q_i]  \,\big] \label{eq:opt_2}\\
&= &	\min_{\ell_{i,\q_i}: \sum_{i \in[m]} \E_\cQ [\ell_{i,\q_i}] \leq \Gamma} \;\; \;\;
		\min_{\tau \in \Tau_{\{\ell_{i,\q_i}\}_{i=1}^m}, \t}\;\; \;\;\max_{\F \in \F_{\sp} } \;\; \frac{1}{m} \sum_{i=1}^m \E_{q_i \sim \cQ } \big[ \, \P_{(\cP,\cQ,\tau)}[t_i \neq \t_i  | q_i ]  \, \big] \label{eq:opt_3}\\
&\geq& 	\min_{\ell_{i,\q_i}: \sum_{i \in[m]} \E_\cQ [\ell_{i,\q_i}]= \Gamma} \;\; \;\; 
		\min_{\tau \in \Tau_{\{\ell_{i,\q_i}\}_{i=1}^m}, \t}\;\; \;\;
		\frac{1}{m} \sum_{i=1}^m \E_{q_i \sim \cQ } \big[ \, \P_{(\cP^*,\cQ,\tau)}[t_i \neq \t_i  | q_i ]  \, \big] \label{eq:opt_3_1}\\
&\geq& \min_{\ell_{i,\q_i}: \sum_{i \in[m]} \E_\cQ [\ell_{i,\q_i}] = \Gamma} \;\; \;\; 
		\frac{1}{m} \sum_{i=1}^m  \left\{ \; \E_{q_i \sim \cQ } \left[ \, \min_{\tau \in \Tau_{ \ell_{i,\q_i} }, \t}\;\; \;\;  \P_{(\cP^*,\cQ,\tau)}[t_i \neq \t_i  | q_i ]  \, \right]  \right\} \label{eq:opt_3_2}\\
&\geq&  \min_{\ell_{i,\q_i}: \sum_{i \in[m]} \E_\cQ [\ell_{i,\q_i}] = \Gamma} \;\; \;\;
		\frac{1}{m} \sum_{i=1}^m \E_{q_i \sim \cQ } \left[ \, \frac{1}{2}  e^{-C'\lambda_i \sp \ell_{i,q_i}} \,\right] \label{eq:opt_4}\\
&=&\min_{\ell_a: \sum_{a\in[T]} \delta_a \ell_a = \Gamma/m} \;\;\; \sum_{a = 1}^{T} \frac{1}{2} \delta_a e^{-C'  \lambda_a \sp  \ell_a }\, \label{eq:lb_ad1}\\
&= &\frac{1}{2} e^{-C' \frac{\Gamma \lambda \sp }{m}}   \bigg(\sum_{a = 1}^T  \delta_a e^{-\lambda \sum_{a' \neq a} (\delta_{a'}/\lambda_{a'}) \log(\lambda_{a}/\lambda_{a'})}   \bigg)  \label{eq:opt_soln}\\
&\geq& \frac{1}{2}  \delta_{\min} e^{-C' \frac{\Gamma \lambda \sp }{m}}  \,,
\end{eqnarray}
where $T$ is the support size of the discrete distribution $\lt\cQ$.
\eqref{eq:opt_3_1} follows from the fact that fixing a prior $\cP^*$ provides a lower bound. 
\eqref{eq:opt_3_2} follows from the fact that exchanging min and sum (and also expectation which is essentially a weighted sum) provides a lower bound. 
\eqref{eq:opt_4} uses Lemma \ref{lem:lb_adapt}. 
\eqref{eq:lb_ad1} follows by change of notations in \eqref{eq:opt_4}.  
\eqref{eq:opt_soln} follows by solving the optimization problem in \eqref{eq:lb_ad1} where the optimal choice of $\ell_a$ is, 
\begin{eqnarray}\label{eq:opt_ell}
\ell_a & \;\;=\;\; &  \frac{\lambda}{\lambda_a} \frac{\Gamma}{m} \;+\; \frac{\lambda}{\lambda_a C' \sp} \bigg (\sum_{a' \neq a} \frac{\delta_{a'}}{\lambda_{a'}} \log\Big(\frac{\lambda_{a}}{\lambda_{a'}}\Big)   \bigg)\,.
\end{eqnarray}
The summand in \eqref{eq:opt_soln} does not depend upon the budget $\Gamma/m$, and it is lower bounded by $\delta_{\min}>0$. This follows from the fact that in the summand the term corresponding to $a$ such that $\lambda_a = \lambda_{\min}$ is lower bounded one $\delta_{\min}$.  

\subsection{Proof of Lemma \ref{lem:lb_adapt}}
We will show that there exists a family of worker reliability distributions $\F^* \in \F_{\sp}$ such that for any adaptive task assignment scheme that assigns $\E[|W_i||q_i]$ workers in expectation to a task $i$ conditioned on its difficulty $q_i$, the conditional probability of error of task $i$ conditioned on $q_i$ is lower bounded by $\exp{(-C'\lambda_i\sigma^2 \E[|W_i||q_i])}$. 
We define the following family of distributions according to the spammer-hammer model with imperfect hammers. We assume that $\sigma^2 < a^2$ and 
\begin{eqnarray*}
	p_j = \left\{ 
	\begin{array}{rl}
	 1/2, & \text{w.p.} \;\;\; 1-\sigma^2/a^2 , \\
	  1/2(1+a), & \text{w.p.} \;\;\;  \sigma^2/a^2.
	 \end{array}
	\right.\;,
\end{eqnarray*}
such that $E[(2p_j-1)^2] = \sigma^2$.
Let $\E[W_i|q_i]$ denote the expected number of workers conditioned on the task difficulty $q_i$, that the adaptive task assignment scheme assigns to the task $i$. We consider a labeling algorithm that has access to an oracle that knows reliability of every worker (all the $p_j$'s). Focusing on a single task $i$, since we know who the spammers are and spammers give no information about the task, we only need the responses from the reliable workers in order to make an optimal estimate. Let $\mathscr{E}_i$ denote the  conditional error probability of the optimal estimate conditioned on the realizations of the answers $\{A_{ij}\}_{j \in W_i}$ and the worker reliability $\{p_j\}_{j \in W_i}$.  We have $\E[\mathscr{E}_i | q_i] \equiv \P[t_i \neq \t_i | q_i]$. The following lower bound on the error only depends on the number of reliable workers, which we denote by $\ell_i$. 

Without loss of generality, let $t_i = +1$. Then, if all the reliable workers agreed on ``--'' answers, the maximum likelihood estimation would be ``--'' for this task, and vice-versa. For a fixed number of $\ell_i$ responses, the probability of error is minimum when all the workers agreed. Therefore, since probability that a worker gives ``--'' answer is $p_j(1-q_i) + q_i(1-p_j) = (1-a(2q_i-1))/2$  from \eqref{eq:defA}, we have,
\begin{eqnarray}
\E[\mathscr{E}_i |q_i, \ell_i] \geq \E[\mathscr{E}_i | \text{all } \ell_i \text{ reliable workers agreed, } q_i, \ell_i ]  \geq  \frac{1}{2}\bigg(\frac{1 - a(2q_i-1))}{2}\bigg)^{\ell_i}\,,
\end{eqnarray}
for any realizations of $\{A_{ij}\}$ and $\{p_j\}$. By convexity and Jensen's inequality, it follows that
\begin{eqnarray}
\E[\ell_i |q_i] \geq \frac{\log(2\E[\mathscr{E}_i | q_i])}{\log((1 - a(2q_i-1))/2)}\,.\label{eq:adapt_lb0}
\end{eqnarray}
When we recruit $|W_i|$ workers, using Doob's Optional-Stopping Theorem \cite[10.10]{williams1991probability}, conditional expectation of reliable number of workers is 
\begin{eqnarray}
\E[\ell_i|q_i] = (\sigma^2/a^2)\E[|W_i| \,| q_i]\,. \label{eq:adapt_lb1}
\end{eqnarray}
Therefore, from \eqref{eq:adapt_lb0} and \eqref{eq:adapt_lb1}, we get
\begin{eqnarray}
\E[|W_i|\,| q_i] \geq \frac{1}{\sigma^2} \frac{a^2}{\log((1- a(2q_i-1))/2)} \log (2\E[\mathscr{E}_i | q_i])\,.
\end{eqnarray}
Maximizing over all choices of $a \in (0,1)$, we get,
\begin{eqnarray}\label{eq:adapt_lb2}
\E[|W_i|\,| q_i] \geq \frac{-0.27}{\sigma^2(2q_i-1)^2} \log (2\E[\mathscr{E}_i | q_i])\,,
\end{eqnarray}
for $a = 0.8/(2q_i-1)$ which as per our assumption of $\sigma^2 < a^2$ requires that $\sigma^2(2q_i-1)^2 < 0.64$. By changing the constant in the bound, we can ensure that the bound holds for any value of $\sigma^2$ and $q_i$. Theorem readily follows from Equation \eqref{eq:adapt_lb2}.  

To verify Equation \ref{eq:adapt_lb1}: Define $X_{i,k}$ for ${k \in [|W_i|]}$ to be a Bernoulli random variable, for a fixed $i \in [m]$ and fixed task difficulty $q_i$. Let $X_{i,k}$ take value one when the $k$-th recruited worker for task $i$ is reliable and zero otherwise. Observe that the number of reliable workers is $\ell_i = \sum_{k = 1}^{|W_i|} X_{i,k}$. From the spammer-hammer model that we have considered, $\E[X_{k,i} - \sigma^2/a^2] = 0$. Define $Z_{i,k} \equiv \sum_{k'= 1}^{k} (X_{i,k'} - \sigma^2/a^2)$ for ${k \in [|W_i|]}$. Since $\{(X_{i,k} - \sigma^2/a^2)\}_{k \in [|W_i|]}$ are mean zero i.i.d. random variables, $\{Z_{i,k}\}_{k \in [|W_i|]}$ is a martingale with respect to the filtration $\mathcal{F}_{i,k} = \sigma(X_{i,1},X_{i,2},\cdots, X_{i,k})$. Further, it is easy to check that the random variable $|W_i|$ for a fixed $q_i$ is a stopping time with respect to the same filtration $\mathcal{F}_{i,k}$ and is almost surely bounded assuming the budget is finite. Therefore using Doob’s Optional-Stopping Theorem \cite[10.10]{williams1991probability}, we have $\E[Z_{i,|W_i|}] = \E[Z_{i,1}] = 0$. That is we have, $\E[X_{i,1}+X_{i,2}+\cdots+ X_{i,|W_i|}] = (\sigma^2/a^2)\E[|W_i|]$. Since this is true for any fixed task difficulty $q_i$, we get Equation \eqref{eq:adapt_lb1}. 
     
% ---------------------------------------------------------------------------------------------------------------------------------------

\subsection{Proof of Theorem \ref{thm:ub_adapt}} \label{sec:adapt_ub}
First we show that the messages returned by Algorithm \ref{algo:mp} are normally distributed and identify their conditional means and conditional variances in the following lemma. Assume in a sub-round $(t,u)$, $t \in [T], u \in[s_t]$, the number of tasks  remaining unclassified are $m_{t,u}$ and the task assignment is performed according to an $(\ell_t, r_t)$-regular random graph.  
To simplify the notation, let $\hat{\ell_t} \equiv \ell_t-1$, $\hat{r_t} \equiv r_t - 1$, and recall $\mu = \E[2p_j-1]$, $\sigma^2 = \E_\cP[(2p_j-1)^2]$. Note that  $\mu, \sigma^2$ remain same in each round. Let $\rho^2_{t,u}= (1/{|M|}) \sum_{i\in[M]}  \lambda_i $ be the exact value of average task difficulty of the tasks present in the $(t,u)$ sub-round.
When $\ell_t$ and $r_t$ are increasing with the problem size, the messages converge to a Gaussian distribution due to the central limit theorem. We provide a proof of this lemma in Section \ref{sec:gaussian}.

\begin{lemma}
	\label{lem:gaussian}
	Suppose for \textcolor{black}{$\ell_t = \Theta(\log m_{t,u})$ and $r_t =\Theta(\log m_{t,u})$,} tasks are assigned according to $(\ell_t,r_t)$-regular random graphs. In the limit $m_{t,u} \rightarrow \infty$, if $\mu > 0$, then 
	after \textcolor{black}{$k = \Theta(\sqrt{\log m_{t,u}})$} number of iterations in Algorithm \ref{algo:mp}, the conditional mean $\mu_q^{(k)}$ and the conditional variance $\big( \rho^{(k)}_q \big)^2$ conditioned on the task difficulty $q$ of the message $x_i$ corresponding to the task $i$ returned by the Algorithm \ref{algo:mp}  are
	\begin{eqnarray} \label{eq:abcd}
		&&\mu^{(k)}_q =  (2q-1) \mu \ell_t (\l_t\r_t\rho_{t,u}^2\sp)^{(k-1)} \;, \nonumber\\
			&&\big( \rho^{(k)}_q \big)^2 \nonumber\\
			&&=  \mu^2 \ell_t ( \l_t\r_t\rho_{t,u}^2\sp)^{2(k-1)} 
\bigg( \rho_{t,u}^2-(2q-1)^2  + \frac{\rho_{t,u}^2\l_t(1-\rho_{t,u}^2\sp)(1+\r_t\rho_{t,u}^2\sp)\big(1-(\l_t\r_t\rho_{t,u}^4\sigma^4)^{-(k-1)}\big)}{\l_t\r_t\rho_{t,u}^4\sigma^4-1}\bigg)  \nonumber \\
&&+ \ell_t (2-\mu^2\rho_{t,u}^2)(\l_t\r_t)^{k-1}\,.		
	\end{eqnarray}
\end{lemma}

We will show in \eqref{eq:adapt6} that the probability of misclassification for any task in sub-round $(t,u)$ in Algorithm \ref{algo:adapt} is 
upper bounded by $e^{-(C_{\delta}/4)(\Gamma/m) \lambda \sigma^2}$. Since, there are at most \textcolor{black}{$C_1 = s_{\max}T \leq \log_2(2\delta_{\max}/\delta_{\min})\log_2(2\lambda_{\max}/\lambda_{\min})$} rounds, using union bound we get the desired probability of error. In \eqref{eq:adapt10}, we show that the expected total number of worker assignments across all rounds is at most $\Gamma$. 

Let's consider any task $i \in [m]$ having difficulty $\lambda_i$. Without loss of generality assume that $t_i = 1$ that is $q_i > 1/2$. Let us assume that the task $i$ gets classified in the $(t,u)$ sub-round, $t \in [T], u \in[s_t]$. That is the number of workers assigned to the task $i$ when it gets classified is $\ell_t = C_{\delta} (\Gamma/m)   (\widehat\lambda/\lambda_t)$ and the threshold $\mathcal{X}_{t,u}$ set in that round for classification is  $\mathcal{X}_{t,u} = \sqrt{\lambda_t} \mu \ell_t \big((\ell_t-1)(r_t-1)\rho^2_{t,u}\sigma^2 \big)^{k_t-1}$. From Lemma \ref{lem:gaussian} the message $x_i$ returned by Algorithm \ref{algo:mp} is Gaussian with conditional mean and conditional variance as given in \eqref{eq:abcd}.
Therefore in the limit of $m$, the probability of error in task $i$ is
\begin{eqnarray}
 \lim_{m \rightarrow \infty} \P\big[\t_i \neq t_i | q_i\big] &=& \lim_{m \rightarrow \infty} \P\big[x_i < -\mathcal{X}_{t,u} | q_i\big] \nonumber\\
 &=& \lim_{m \rightarrow \infty} Q\Big(\frac{\mu^{(k)}_{q_i} + \mathcal{X}_{t,u}}{\rho^{(k)}_{q_i}}\Big) \label{eq:adapt1}\\
 &\leq &  \lim_{m \rightarrow \infty} \exp\Big(\frac{-(\mu^{(k)}_{q_i} + \mathcal{X}_{t,u})^2}{2(\rho^{(k)}_{q_i})^2}\Big) \label{eq:adapt2}\\
&=&   \exp\Big(\frac{-((2q_i-1)+ \sqrt{\lambda_t})^2\ell_t\sp}{2(1 - (2q_i-1)^2\sp)}\Big) \label{eq:adapt3} \\
&\leq & \exp\Big(\frac{-\lambda_t\ell_t\sp}{2}\Big) \nonumber\\
&=& \exp\Big(\frac{-C_{\delta} (\Gamma/m)   \widehat\lambda \sp}{2}\Big) \label{eq:adapt5}\\
&\leq & \exp\Big(\frac{-C_{\delta} (\Gamma/m)  \lambda \sp}{4}\Big)\,, \label{eq:adapt6}
\end{eqnarray}   
where $Q(\cdot)$ in \eqref{eq:adapt1} is the tail probability of a standard Gaussian distribution, and \eqref{eq:adapt2} uses the Chernoff bound. \eqref{eq:adapt3} follows from substituting conditional mean and conditional variance from Equation \eqref{eq:abcd}, and using $\ell_t = \Theta(\log m_{t,u})$, $k = \Theta(\sqrt{\log m_{t,u}})$ where $m$ grows to infinity. \eqref{eq:adapt5} uses $\ell_t = C_{\delta} (\Gamma/m) (\widehat\lambda/\lambda_t)$, our choice of $\ell_t$ in Algorithm \ref{algo:adapt} line 4.
\eqref{eq:adapt6} uses the fact that for the quantized distribution  $\{\lambda_a, \delta_a\}_{a \in [T]}$, $\widehat \lambda = \big(\sum_{a \in [T]} (\delta_a/\lambda_a)\big)^{-1} \geq \lambda/2$. 
We have established that our approach guarantees the desired level of accuracy. We are left to show that we use at most $\Gamma$ assignments in expectation. 

We upper bound the expected total number of workers used for tasks of quantized difficulty level $\lambda_a$'s for each $ 1 \leq a \leq T$. Recall that our adaptive algorithm runs in $T$ rounds indexed by $t$, where each round $t$ further runs $s_t$ sub-rounds. The total expected number of workers assigned to $\delta_a$ fraction of tasks of quantized difficulty $\lambda_a$ in $t=1$ to $t=a-1$ rounds is upper bounded by $m\delta_a \sum_{t=1}^{a-1} s_t\ell_t$. The upper bound assumes the worst-case 
(in terms of the budget) that these tasks do not get classified in any of these rounds as the threshold $\mathcal{X}$ set in these rounds is more than absolute value of the conditional mean message $x$ of these tasks.

 Next, in $s_{t=a}$ sub-rounds the threshold $\mathcal{X}$ is set less than or equal to the absolute value of the conditional mean message $x$ of these tasks, 
 i.e. $\mathcal{X} \leq |\mu_{q_a}^{(k)}|$ for $(2q_a-1)^2=\lambda_a$. 
 Therefore, in each of these $s_a$ sub-rounds, probability of classification of these tasks is at least $1/2$. That is the expected total number of workers assigned to these tasks in $s_a$ sub-rounds is upper bounded by $2m\delta_a\ell_a$. Further, $s_a$ is chosen such that the fraction of these tasks remaining un-classified at the end of $s_a$ sub-rounds is at most same as the fraction of the tasks having difficulty $\lambda_{a+1}$. That is to get the upper bound, we can assume that the fraction of $\lambda_{a+1}$ difficulty tasks at the start of $s_{a+1}$ sub-rounds is $2\delta_{a+1}$, and the fraction of $\lambda_a$ difficulty tasks at the start of $s_{a+1}$ sub-rounds is zero. Further, recall that we have set $s_{T} = 1$ as in this round our threshold  $\mathcal{X}$ is equal to zero. Therefore, we have the following upper bound on the expected total number of worker assignments.
\begin{eqnarray}
\sum_{i = 1}^m \E[|W_i|] &\leq & 2m\delta_1\ell_1 + \sum_{a=2}^{T-1} 4 m\delta_a \ell_a + 2m\delta_{T}\ell_{T} + \sum_{a =2}^{T}\Big(m\delta_a \sum_{b=1}^{a-1} s_b\ell_b \Big) \nonumber \\
& \leq &   \sum_{a=1}^{T} 4 m\delta_a \ell_a + s_{\max} \sum_{a=1}^{T} m\delta_a \ell_a \label{eq:adapt7} \\
 & \leq & (4 + \ceil{\log(2\delta_{\max}/\delta_{\min})}) \sum_{a=1}^{T}m \delta_a \ell_a \label{eq:adapt8} \\
& \leq &  (4 + \ceil{\log(2\delta_{\max}/\delta_{\min})}) \Gamma C_{\delta}  \label{eq:adapt9} \\
&=& \Gamma\,,\label{eq:adapt10}
\end{eqnarray} 
Equation \eqref{eq:adapt7} uses the fact that  $\ell_t = ( C_{\delta} (\Gamma/m)(\widehat\lambda/\lambda_t)$ where $\lambda_t$'s are separated apart by at least a ratio of 2 (recall the quantized distribution), therefore $\sum_{t=1}^{a-1} \ell_t \leq \ell_a$. Equation \eqref{eq:adapt8} follows from the choice of $s_t$'s in the algorithm. Equation \eqref{eq:adapt9} follows from using $\ell_t = ( C_{\delta} (\Gamma/m)(\widehat\lambda/\lambda_t)$ and  $\lambda = (\sum_{a \in [T]} (\delta_a/\lambda_a))^{-1}$, and Equation \eqref{eq:adapt10} uses $C_{\delta} = ( 4 + \ceil{\log(2\delta_{\max}/\delta_{\min})})^{-1}$.

\subsection{Proof of Lemma \ref{lem:gaussian}} \label{sec:gaussian}
We omit subscripts $t$ and $(t,u)$ from all the quantities for simplicity of notations. Also, we use notation $\ell$, average budget per task, $\ell = \Gamma/m$.
We will prove it for a randomly chosen task $\i$, and all the analyses naturally holds for a specific $i$, 
when conditioned on $q_i$. 
Let $n$ be the number of workers, that is $n = (m r) /\ell$. In our algorithm, we perform task assignment on a random bipartite graph $\G([m]\cup[n],E)$ constructed according to the configuration model. Let $\G_{i,k}$ denote a subgraph of  $\G([m]\cup[n],E)$ that includes all the nodes that are within $k$ distance from the the ``root" $i$. If we run our inference algorithm for one run to estimate $\t_i$, we only use the responses provided by the workers who were assigned to task $i$. That is we are running inference algorithm only on the local neighborhood graph $\G_{i,1}$. Similarly, when we run our algorithm for $k$ iterations to estimate $\t_i$, we perform inference only on the local subgraph $\G_{i,2k-1}$. Since we update both task and worker messages at each iteration, the local subgraph grows by distance two at each iteration. We use a result from \cite{KOS14OR} to show that the local neighborhood of a randomly chosen task node $\i$ is a tree with high probability. Therefore, assuming that the graph is locally tree like with high probability, we can apply a technique known as {\em{density evolution}} to estimate the conditional mean and conditional variance. 
The next lemma shows that the local subgraph converges to a tree in probability, in the limit $m \rightarrow \infty$ for the specified choice of $\ell,r$ and $k$.
\begin{lemma}[Lemma 5 from \cite{KOS14OR}] \label{lem:lemma1}
For a random $(\ell,r)$-regular bipartite graph generated according to the configuration model,
\begin{eqnarray}\label{eq:eq2}
\P\big[\G_{\i,2k-1} \;{\text{\rm{is not a tree}}}\big]  \leq  \big((\ell-1)(r-1)\big)^{2k-2} \frac{3\ell r}{m}.
\end{eqnarray}
\end{lemma}

{\bf{Density Evolution.}} Let $\{x^{(k)}_{i \rightarrow j}\}_{(i,j) \in E}$ and $\{y^{(k)}_{j \rightarrow i}\}_{(i,j) \in E}$ denote the messages at the $k$-th iteration of the algorithm. For an edge $(i,j)$ chosen uniformly at random, let $\x^{(k)}_q$ denote the random variable corresponding to the message $x^{(k)}_{i \rightarrow j}$ conditioned on the $i$-th task's difficulty being $q$. Similarly, let $\y_p^{(k)}$ denote the random variable corresponding to the message $y^{(k)}_{j \rightarrow i}$ conditioned on the $j$-th worker's quality being $p$.  
%If we choose an edge $(i,j)$ uniformly at random, the values of $x$ and $y$ messages on that randomly chosen edge define random variables whose randomness comes from random choice of the edge, any randomness introduced by the inference algorithm, the graph $\G$, and the realizations of $\q_i$'s, $\p_j$'s and $\A_{ij}$'s. 
%As proved in Lemma \ref{lem:lemma1}, the $(\ell,r)$-regular random graph locally converges in distribution to a $(\ell,r)$-regular tree with high probability. 

At the first iteration, the task messages are updated according to $x_{i \rightarrow j}^{(1)} = \sum_{j' \in \partial i \setminus j } \A_{ij'}y_{j' \rightarrow i }^{(0)}$. Since we initialize the worker messages $\{y^{(0)}_{j \rightarrow i}\}_{(i,j) \in E}$ with independent Gaussian random variables with mean and variance both one, if we know the distribution of  $\A_{ij'}$'s, then we have the distribution of $x_{i \rightarrow j }^{(1)}$. Since, we are assuming that the local subgraph is tree-like, all $x_{i \rightarrow j }^{(1)}$ for $i \in \G_{\i,2k-1}$ for any randomly chosen node $\i$ are independent. Further, because of the symmetry in the construction of the random graph $\G$ all messages $x_{i \rightarrow j }^{(1)}$'s are identically distributed. Precisely, $x_{i \rightarrow j }^{(1)}$ are distributed according to $\x_q^{(1)}$ defined in Equation  \eqref{eq:eq7}. In the following, we recursively define $\x_q^{(k)}$ and $\y_p^{(k)}$ in Equations \eqref{eq:eq7} and \eqref{eq:eq8}.        

For brevity, here and after, we drop the superscript $k$-iteration number whenever it is clear from the context. Let $\x_{q,a}$'s and $\y_{p,b}$'s be independent random variables distributed according to $\x_q$ and $\y_p$ respectively. We use $a$ and $b$ as indices for independent random variables with the same distribution. Also, let $\z_{p,q,a}$'s and $\z_{p,q,b}$'s be independent random variables distributed according to $\z_{p,q}$, where 
\begin{eqnarray} \label{eq:eq6}
\z_{p,q} = \begin{cases}
				+1 \;\; \text{w.p.}\;\;\;\; pq + (1-p)(1-q)\,,\\
				-1 \;\; \text{w.p.}\;\;\;\; p(1-q) + (1-p)q\,.
			\end{cases}
\end{eqnarray} 
This represents the response given by a worker conditioned on the task having difficulty $q$ and the worker having ability $p$. Let $\F_1$ and $\F_2$ over $[0,1]$ be the distributions of the tasks' difficulty level and workers' quality respectively. Let $\q \sim \F_1$ and $\p \sim \F_2$. Then $\q_a$'s and $\p_b$'s are independent random variables distributed according to $\q$ and $\p$ respectively. Further, $\z_{p,\q_a,a}$'s and $\x_{\q_a,a}$'s are conditionally independent conditioned on $\q_a$; and $\z_{\p_b,q,b}$'s and $\y_{\p_b,b}$'s are conditionally independent conditioned on $\p_b$.

%We initialize $\y_p$ with a Gaussian distribution, whence it is independent of the latent variable $p : \y_p^{(0)} \sim \N(1,1)$. 
Let $\overset{d}{=}$ denote equality in distribution. Then for $k \in \{1,2,\cdots\}$, the task messages (conditioned on the latent task difficulty level $q$) are distributed as the sum of $\ell-1$ incoming messages that are i.i.d. according to $\y_{\p}^{(k-1)}$ and weighted by i.i.d. responses:
\begin{eqnarray}\label{eq:eq7}
\x_q^{(k)} \overset{d}{=} \sum_{b \in [\ell-1]} \z_{\p_b,q,b}\y_{\p_b,b}^{(k-1)}.
\end{eqnarray}
Similarly, the worker messages (conditioned on the latent worker quality $p$) are distributed as the sum of $r-1$ incoming messages that are i.i.d. according to $\x_{\q}^{(k)}$ and weighted by the i.i.d. responses:
\begin{eqnarray} \label{eq:eq8}
\y_p^{(k)} \overset{d}{=} \sum_{a \in [r-1]} \z_{p,\q_a,a}\x_{\q_a,a}^{(k)}.
\end{eqnarray}
For the decision variable $\x_{\i}^{(k)}$ on a task $\i$ chosen uniformly at random, we have
\begin{eqnarray}\label{eq:eq9}
\hat{\x}_q^{(k)} \overset{d}{=} \sum_{a \in [\ell]} \z_{\p_a,q,a}\y_{\p_a,a}^{(k-1)}.
\end{eqnarray}
     
{\bf{Mean and Variance Computation.}}
Define $m_{\q}^{(k)} \equiv \E[\x^{(k)}_{\q}|\q]$ and $\hat{m}_{\p}^{(k)} \equiv \E[\y^{(k)}_{\p}|\p]$, $\nu^{(k)}_{\q} \equiv {\rm{Var}}(\x_{\q}^{(k)} | \q)$ and $\hat{\nu}^{(k)}_{\p} \equiv {\rm{Var}}(\y_{\p}^{(k)}| \p)$. Recall the notations $\mu \equiv \E[2\p-1]$, $\sq \equiv \E[(2\q-1)^2]$, $\sp \equiv \E[(2\p-1)^2]$, $\hat{\ell} = \ell-1$, and $\hat{r} = r - 1$. Then from \eqref{eq:eq7} and \eqref{eq:eq8} and using $\E[\z_{p,q}] = (2p-1)(2q-1)$ we get the following:
\begin{align} 
&m_{\q}^{(k)} = \l(2\q-1) \E_{\p}\big[(2\p-1) \hat{m}_{\p}^{(k-1)} \big],  \label{eq:eq10a}\\
&\hat{m}_{\p}^{(k)} = \r(2\p-1) \E_{\q}\big[(2\q-1) m_{\q}^{(k)} \big], \label{eq:eq10}\\
&\nu^{(k)}_{\q} = \l\Big\{\E_{\p}\big[\hat{\nu}^{(k-1)}_{\p} + (\hat{m}_{\p}^{(k-1)})^2 \big] -  (m_{\q}^{(k)}/\l)^2\Big\},  \label{eq:eq11a}\\
&\hat{\nu}^{(k)}_{\p} = \r\Big\{\E_{\q}\big[{\nu}^{(k)}_{\q} + (m_{\q}^{(k)})^2 \big] -  (\hat{m}_{\p}^{(k)}/\r)^2\Big\}. \label{eq:eq11}
\end{align}
Define $m^{(k)} \equiv \E_{\q}[(2\q-1)m_{\q}^{(k)}]$ and $\nu^{(k)} \equiv \E_{\q}[\nu_{\q}^{(k)}]$. From \eqref{eq:eq10a} and \eqref{eq:eq10}, we have the following recursion on the first moment of the random variable $x_{q}^{(k)}$:
\begin{eqnarray} \label{eq:eq12}
m_{\q}^{(k)} = \l\r(2\q-1)\sp m^{(k-1)}, m^{(k)} = \l\r\sq\sp m^{(k-1)}\,.
\end{eqnarray} 
From \eqref{eq:eq11a} and \eqref{eq:eq11}, and using $\E_{\q}[(m_{\q}^{(k)})^2] = (m^{(k)})^2/\sq$ (from \eqref{eq:eq12}), and $\E_{\p}[(\hat{m}_{\p}^{(k)})^2] = \r^2\sp (m^{(k)})^2$ (from \eqref{eq:eq10}) , we get the following recursion on the second moment:
\begin{eqnarray}
\nu^{(k)}_{\q} &=& \l\r\nu^{(k-1)} + \l\r(m^{(k-1)})^2\big((1-\sq\sp)(1+\r\sq\sp) + \r\sq(\sp)^2(\sq - (2\q-1)^2 )\big)/\sq\,, \label{eq:eq13}\\
\nu^{(k)} &=& \l\r\nu^{(k-1)} + \l\r(m^{(k-1)})^2(1-\sq\sp)(1+\r\sq\sp)/\sq. \label{eq:eq14}
\end{eqnarray}
Since $\hat{m}_{\p}^{(0)} = 1$ as per our assumption, we have $m^{(1)}_{\q} = \l\mup(2\q-1)$ and $m^{(1)} = \l\mup\sq$. Therefore from \eqref{eq:eq12}, we have $m^{(k)} = \l\mup\sq(\l\r\sq\sp)^{k-1}$ and $m^{(k)}_{\q} = \l\mup(2\q-1)(\l\r\sq\sp)^{k-1}$. Further, since $\hat{\nu}^{(0)}_{\p} = 1$ as per our assumption, we have $\nu^{(1)}_{\q} = \l(2 - \mup^2(2\q-1)^2)$ and $\nu^{(1)} = \l(2 - \mup^2\sq)$. This implies that $\nu^{(k)} = a\nu^{(k-1)} + bc^{k-2}$, with $a = \l\r$, $b = \mup^2\sq\l^3\r(1-\sq\sp)(1+\r\sq\sp)$ and $c = (\l\r\sq\sp)^2$. After some algebra, we have that $\nu^{(k)} = \nu^{(1)}a^{k-1} + bc^{k-2}\sum_{\ell=0}^{k-2}(a/c)^{\ell}$. For $\l\r(\sq\sp)^2 >1$, we have $a/c < 1$ and
\begin{align}\label{eq:eq15}
&\nu_{\q}^{(k)} = \l(2-\mup^2\sq)(\l\r)^{k-1} + \mup^2\l(\l\r\sq\sp)^{2k-2}(\sq - (2\q-1)^2) \nonumber\\  
&\hspace{0.8cm}+\bigg(\frac{1-1/(\l\r(\sq\sp)^2)^{k-1}}{\l\r\rho^4\sigma^4-1} \bigg)(1-\sq\sp)(1+\r\sq\sp)\mup^2\sq\l^2(\l\r\sq\sp)^{2k-2}.
\end{align} 
By a similar analysis, mean and variance of the decision variable $\hat{\x}_{\q}^{(k)}$ in \eqref{eq:eq9} can also be computed. In particular, they are $\ell/\l$ times $m_{\q}^{(k)}$ and $\nu_{\q}^{(k)}$. Gaussianity of the messages follows due to Central limit theorem.

\subsection{Proof of Theorem \ref{thm:main_thm}} \label{sec:nonadapt_ub}
The proof uses the results derived in the proof of Lemma \ref{lem:gaussian}. 

Let $\htk_i$ denote the resulting estimate of task $i$ after running the iterative inference algorithm for $k$ iterations. We want to compute the conditional probability of error of a task $\i$ selected uniformly at random in $[m]$, conditioned on its difficulty level, i.e.,
$$ \P\big[t_{\i} \neq \htk_{\i} \big| q_{\i}\big]\,.$$ 
In the following, we assume $q_{\i} \geq (1/2)$, i.e. the true label is $t_i = 1$. Analysis for $q_{\i} \leq (1/2)$ would be similar and result in the same bounds. 
Using the arguments given in Lemma \ref{lem:gaussian}, we have,
\begin{eqnarray} \label{eq:eq1}
\P\big[ t_{\i} \neq \htk_{\i}\big | q_{\i}\big] & \leq &   \P\big[ t_{\i} \neq \htk_{\i}\big  | \G_{\i,2k-1} \;\text{is a tree},\; q_{\i}\big] + \P\big[\G_{\i,2k-1} \;\text{is not a tree}\big].
\end{eqnarray}   
To provide an upper bound on the first term in \eqref{eq:eq1}, let $x_i^{(k)}$ denote the decision variable for task $i$ after $k$ iterations of the algorithm such that $\htk_i ={\rm{sign}}(x_i^{(k)})$. Then as per our assumption that $t_i = 1$, we have,
\begin{eqnarray}\label{eq:eq3}
\P\big[ t_{\i} \neq \htk_{\i} | \G_{\i,2k-1} \text{is a tree}, q_{\i}\big] & \leq & \P\big[ x_{\i}^{(k)} \leq 0  | \G_{\i,2k-1} \text{is a tree}, q_{\i}\big].
\end{eqnarray}
Next, we apply ``density evolution" \cite{mezard2009information} and provide a sharp upper bound on the probability of the decision variable $x_{\i}^{(k)}$ being negative in a locally tree like graph given $q_{\i} \geq (1/2)$. The proof technique is similar to the one introduced in \cite{KOS14OR}. Precisely, we show,
\begin{eqnarray}\label{eq:eq4}
\P\big[ x_{\i}^{(k)} \leq 0  | \G_{\i,2k-1} \;\text{is a tree}\;, q_{\i}\big]  =   \P\big[ \hat{\x}_{q}^{(k)} \leq 0 \big]\,,
\end{eqnarray}
where $\hat{\x}_q^{(k)}$ is defined in Equations \eqref{eq:eq7}-\eqref{eq:eq9} using density evolution. We will prove in the following that when $\hat{\ell}\hat{r}(\sq\sp)^2 > 1$ and $\hat{r}\sq > 1$, 
\begin{eqnarray}\label{eq:eq5}
\P\big[ \hat{\x}_q^{(k)} \leq 0 \big] & \leq & e^{-\ell\sp(2q_{\i}-1)^2/ (2\sigma_k^2)}. 
\end{eqnarray}
Theorem \ref{thm:main_thm} follows by combining Equations \eqref{eq:eq1},\eqref{eq:eq2},\eqref{eq:eq3} and \eqref{eq:eq4}.

we show that $\hat{\x}^{(k)}$ is sub-Gaussian with some appropriate parameter and then apply the Chernoff bound. A random variable $\x$ with mean $\mu$ is said to be {\em{sub-Gaussian}} with parameter $\sigma$ if for all $\lambda \in \reals$ the following bound holds for its moment generating function:
\begin{eqnarray} \label{eq:eq18}
\E[e^{\lambda\x}] &\leq & e^{\mu\lambda + (1/2)\sigma^2\lambda^2}\,. 
\end{eqnarray} 
Define, 
\begin{align}
&\stk \equiv 3\l^3\r\mup^2\sq(\r\sq\sp + 1)(\l\r\sq\sp)^{2k-4}\big(\frac{1 - 1/(\l\r(\sq\sp)^2)^{k-1}}{1 - 1/(\l\r\sq\sp)}\big)
+2\l(\l\r)^{k-1}\,, \label{eq:effectivevariance}
\end{align}
$m_{k} \equiv \mup\l(\l\r\sq\sp)^{k-1}$, and $m_{k,\q} \equiv (2\q-1)m_{k}$ for $k \in \mathbb{Z}$, where $\q \sim \F_1$. We will show that, $\x_{\q}^{(k)}$ is sub-Gaussian with mean $m_{k,\q}$ and parameter $\stk$ for $|\lambda| \leq 1/(2m_{k-1}\r\sq)$, i.e.,
\begin{eqnarray}\label{eq:eq19}
\E[e^{\lambda \x_{\q}^{(k)}}| \q] & \leq & e^{m_{k,\q}\lambda + (1/2)\stk\lambda^2}\,.
\end{eqnarray}

{\bf{Analyzing the Density.}} 
Notice that the parameter $\stk$ does not depend upon the random variable $\q$. By definition of $\hat{\x}_{\q}^{(k)}$, \eqref{eq:eq9}, we have $\E[e^{\lambda\hat{\x}_{\q}^{(k)}}| \q] = \E[e^{\lambda\x_{\q}^{(k)}}|\q]^{(\ell/\l)}$. Therefore, it follows that $\E[e^{\lambda\hat{\x}_{\q}^{(k)}}|\q] \leq e^{(\ell/\l)m_{k,\q}\lambda + (\ell/2\l)\stk\lambda^2}$. Using the Chernoff bound with $\lambda = - m_{k,\q}/(\stk)$, we have
\begin{eqnarray}\label{eq:eq20}
\P[\hat{\x}_{\q}^{(k)} \leq 0 \;|\; \q] \;\leq\; \E[e^{\lambda\hat{\x}_{\q}^{(k)}} |\q] \;\leq\; e^{-\ell m_{k,\q}^2/(2\l\stk)}\,.
\end{eqnarray}
Note that, with the assumption that $\q \geq (1/2)$, $m_{k,\q}$ is non-negative. Since 
\begin{eqnarray*} \label{eq:eq21}
\frac{m_{k,\q} m_{k-1,\q}}{\stk}  \leq  \frac{(2\q-1)^2\mup^2\l^2(\l\r\sq\sp)^{2k-3}}{3\mup^2\sp(\sq)^2\l^3\r^2(\l\r\sq\sp)^{2k-4}} = \frac{(2\q-1)^2}{3\r\sq}\,, 
\end{eqnarray*}
it follows that $|\lambda| \leq 1/(2m_{k-1}\r\sq)$. The desired bound in \eqref{eq:eq5} follows. 

Now, we are left to prove Equation \eqref{eq:eq19}. From \eqref{eq:eq7} and \eqref{eq:eq8}, we have the following recursive formula for the evolution of the moment generating functions of $\x_{\q}$ and $\y_{\p}$:
\begin{eqnarray}
\E[e^{\lambda\x_{\q}^{(k)}} |\q] &=& \big(\E_{\p}\big[(\p\q+ \pb\qb)\E[e^{\lambda\y_{\p}^{(k-1)}}|\p] + (\p\qb+ \pb\q)\E[e^{-\lambda\y_{\p}^{(k-1)}}|\p]\big]  \big)^{\l}\,,\label{eq:eq22}\\
\E[e^{\lambda\y_{\p}^{(k)}} |\p] &=& \big(\E_{\q}\big[(\p\q+ \pb\qb)\E[e^{\lambda\x_{\q}^{(k)}}|\q] + (\p\qb+ \pb\q)\E[e^{-\lambda\x_{\q}^{(k)}} |\q]\big]  \big)^{\r}\,, \label{eq:eq23}
\end{eqnarray}
where $\pb = 1- \p$ and $\qb = 1-\q$. We apply induction to prove that the messages are sub-Gaussian. First, for $k = 1$, we show that $\x_{\q}^{(1)}$ is sub-Gaussian with mean $m_{1,\q} = (2\q-1)\mup\l$ and parameter $\tilde{\sigma}_{1}^2 = 2\l$. Since, $\y_{\p}$ is initialized as a Gaussian random variable with mean and variance both one, we have $\E[e^{\lambda\y_{\p}^{(0)}}] = e^{\lambda + (1/2)\lambda^2}$. Substituting this into Equation \eqref{eq:eq22}, we get for any $\lambda$,
\begin{eqnarray} \label{eq:eq24}
\E[e^{\lambda\x_{\q}^{(1)}}|\q] &=& \big(\big(\E[\p]\q+\E[\pb]\qb\big)e^{\lambda} + \big(\E[\p]\qb+\E[\pb]\q\big)e^{-\lambda} \big)^{\l} e^{(1/2)\lambda^2\l} \\
& \leq & e^{(2\q-1)\mup\l\lambda + (1/2)(2\l)\lambda^2}\,,
\end{eqnarray}
where the inequality follows from the fact that $ae^{z} + (1-a)e^{-z} \leq e^{(2a-1)z + (1/2)z^2}$ for any $z \in \reals$ and $a \in [0,1]$ (Lemma A.1.5 from \cite{alon2004probabilistic}). Next, assuming $\E[e^{\lambda\x_{\q}^{(k)}} |\q] \leq e^{m_{k,\q}\lambda + (1/2)\stk\lambda^2}$ for $|\lambda| \leq 1/(2m_{k-1}\r\sq)$, we show that $\E[e^{\lambda\x_{\q}^{(k+1)}} |\q] \leq e^{m_{k+1,\q}\lambda + (1/2)\tilde{\sigma}^2_{k+1}\lambda^2}$ for  $|\lambda| \leq 1/(2m_{k}\r\sq)$, and compute appropriate $m_{k+1,\q}$  and $\tilde{\sigma}^2_{k+1}$. 

Substituting the bound $\E[e^{\lambda\x_{\q}^{(k)}} |\q] \leq e^{m_{k,\q}\lambda + (1/2)\stk\lambda^2}$ in \eqref{eq:eq23}, we have
\begin{align}
&\E[e^{\lambda\y_{\p}^{(k)}}| \p]  \nonumber\\
&\leq \big(\E_{\q}\big[(\p\q + \pb\qb)e^{m_{k,\q}\lambda} + (\p\qb + \pb\q) e^{-m_{k,\q}\lambda}  \big] \big)^{\r} e^{(1/2)\stk\lambda^2\r} \nonumber\\
&\leq \big(\E_{\q}\big[e^{(2\q-1)(2\p-1)m_{k,\q}\lambda + (1/2)(m_{k,\q}\lambda)^2} \big] \big)^{\r}e^{(1/2)\stk\lambda^2\r} \label{eq:eq25}\\
&=\big(\E_{\q}\big[e^{(2\p-1)(2\q-1)^2m_{k}\lambda + (1/2)(2\q-1)^2(m_{k}\lambda)^2} \big] \big)^{\r}e^{0.5\stk\lambda^2\r} \label{eq:eq251}
\end{align}
where \eqref{eq:eq25} uses the inequality $ae^{z} + (1-a)e^{-z} \leq e^{(2a-1)z + (1/2)z^2}$ and \eqref{eq:eq251} follows from the definition of $m_{k,\q} \equiv (2\q-1)m_{k}$. To bound the term in \eqref{eq:eq251}, we use the following lemma. 
\begin{lemma}\label{lem:lemma2}
For any random variable $s \in [0,1]$, $|z| \leq 1/2$ and $|t| < 1$, we have
\begin{eqnarray} \label{eq:lem2}
\E\big[e^{stz + (1/2)sz^2}\big] \leq \exp\big({\E[s]tz + (3/2)\E[s]z^2}\big)\,.
\end{eqnarray}
\end{lemma}
For $|\lambda| \leq 1/(2m_{k}\r\sq)$, using the assumption that $\r\sq >1 $, we have $m_{k} \lambda \leq (1/2)$. Applying Lemma \ref{lem:lemma2} on the term in \eqref{eq:eq251}, with $s = (2\q-1)^2$, $z= m_k\lambda$ and $t = (2p-1)$, 
we get
\begin{eqnarray}
\E[e^{\lambda\y_{\p}^{(k)}}| \p] 
\leq e^{\sq(2\p-1)\r m_{k}\lambda + (1/2)\big(3\sq m_{k}^2 + \stk\big)\lambda^2\r}\,. \label{eq:eq26}
\end{eqnarray}
 
Substituting the bound in \eqref{eq:eq26} in Equation \eqref{eq:eq22}, we get
\begin{eqnarray}
&&\E[e^{\lambda\x_{\q}^{(k+1)}}| \q] \nonumber\\
& \leq & \big(\E_{\p}\big[(\p\q+ \pb\qb)e^{\sq(2\p-1)m_{k}\lambda\r} + (\p\qb+ \pb\q)e^{-\sq(2\p-1)m_{k}\lambda\r} \big] \big)^{\l}e^{ (1/2)(3\sq m_{k}^2 + \stk)\lambda^2\l\r} \nonumber\\
& \leq & \big(\E_{\p}\big[e^{(2\q-1)(2\p-1)^2\sq m_{k}\lambda \r + (1/2)(2\p-1)^2(\sq m_{k}\lambda\r)^2}\big] \big)^{\l}e^{ (1/2)(3\sq m_{k}^2 + \stk)\lambda^2\l\r} \label{eq:eq27}\\
& \leq & e^{\l\r\sq\sp m_{k,\q}\lambda + (1/2)\l\r\big(\stk + 3\sq m_{k}^2(1+ \r\sq\sp ) \big)\lambda^2}\label{eq:eq28}\,,
\end{eqnarray}
where \eqref{eq:eq27} uses the inequality $ae^{z} + (1-a)e^{-z} \leq e^{(2a-1)z + (1/2)z^2}$. Equation \eqref{eq:eq28} follows from the application of Lemma \ref{lem:lemma2}, with $s = (2\p-1)^2$, $z = \sq m_k \lambda \r$ and $t = (2\q-1)$. For $|\lambda| \leq 1/(2m_{k}\r\sq)$, $|z| < (1/2)$.

In the regime where $\l\r(\sq\sp)^2 > 1$, as per our assumption, $m_{k}$ is non-decreasing in $k$. At iteration $k$, the above recursion holds for $|\lambda| \leq 1/(2\r\sq) \min \{1/m_{1},\cdots,1/m_{k-1} \} = 1/(2m_{k-1}\r\sq)$. Hence, we get the following recursion for $m_{k,\q}$ and $\stk$ such that \eqref{eq:eq19} holds for $|\lambda| \leq 1/(2m_{k-1}\r\sq)$:
\begin{eqnarray}
m_{k,\q} &=& \l\r\sq\sp m_{k-1,\q}, \nonumber\\
\tilde{\sigma}^2_{k} &=& \l\r\tilde{\sigma}_{k-1}^2 + 3\l\r(1 + \r\sq\sp )\sq m_{k-1}^2\,. \label{eq:tail}
\end{eqnarray}
With the initialization $m_{1,\q} = (2\q-1)\mup\l$ and $\tilde{\sigma}^2_{1} = 2\l$, we have $m_{k,\q} = \mup(2\q-1)\l(\sq\sp\l\r)^{k-1}$ for $k \in \{1,2,\cdots\}$ and $\stk = a\tilde{\sigma}_{k-1}^2 + bc^{k-2}$ for $k \in \{2,3\cdots\}$, with $a = \l\r$, $b = 3\l^3\r\mup^2\sq(1+\sq\sp\r)$, and $c = (\sq\sp\l\r)^2$. After some algebra, we have $\stk = \tilde{\sigma}_{1}^2 a^{k-1} + bc^{k-2}\sum_{\ell = 0}^{k-2}(a/c)^\ell$. For $\l\r(\sq\sp)^2 \neq 1$, we have $a/c \neq 1$, whence $\stk = \tilde{\sigma}_{1}^2 a^{k-1} + bc^{k-2}(1-(a/c)^{k-1})/(1-a/c)$. This finishes the proof of \eqref{eq:eq19}.\\

\subsection{Proof of Lemma    \ref{lem:lemma2}}
Using the fact that $e^a \leq 1 + a+  0.63a^2$ for $|a| \leq 5/8$,
\begin{eqnarray*}
&&\E\big[e^{stz + (1/2)sz^2}\big] \\
& \leq & \E\big[1 + stz + (1/2)sz^2 + 0.63\big(stz + (1/2)sz^2\big)^2 \big]\\
& \leq & \E\big[1 + stz + (1/2)sz^2 + 0.63\big((5/4)z\sqrt{s}\big)^2 \big]\\
& \leq & 1+ \E[s]tz + (3/2)\E[s]z^2 \\
&\leq & \exp\big(\E[s]tz + (3/2)\E[s]z^2\big)\,. 
\end{eqnarray*}

%-------------------------------------------------------------------------------------

\subsection{Proof of Theorem \ref{thm:thm_lb}} \label{sec:nonadapt_lb}
Let $\F$ denote a distribution on the worker quality $\p_j$ such that $\p_j \sim \F$. Let $\F_{\sp}$ be a collection of all distributions $\F$ such that:
\begin{eqnarray*}
\F_{\sp} = \big\{ \F \;| \;\E_{\F}[(2\p_j-1)^2] = \sp \big\}\,.
\end{eqnarray*}
Define the minimax rate on the probability of error of a task $i$, conditioned on its difficulty level $q_i$, 
as
\begin{eqnarray} \label{eq:eqlb1}
\min_{\tau \in \mathscr{T}_{\ell_i}, \t} \;\; \max_{t_i \in \{\pm\}, \F \in \F_{\sp}} \P[ t_i \neq \t_i \; | \; q_i]\,,
\end{eqnarray}
where $\mathscr{T}_{\ell_i}$ is the set of all nonadaptive task assignment schemes that assign $\ell_i$ workers to task $i$, and $\t$ ranges over the set of all estimators of $t_i$. Since the minimax rate is the maximum over all the distributions $\F \in \F_{\sp}$, we consider a particular worker quality distribution to get a lower bound on it. In particular, we assume the $\p_j$'s are drawn from a spammer-hammer model with perfect hammers:
\begin{eqnarray*}
\p_j \;\; = \;\; \begin{cases} 1/2 \;\;\; \text{with probability} \;1-\sp, \\ 
					   1 \;\;\; \text{otherwise.}
		\end{cases}
\end{eqnarray*}
Observe that the chosen spammer-hammer models belongs to $\F_{\sp}$, i.e. $\E[(2\p_j-1)^2] = \sp$. 
To get the optimal estimator, we consider an oracle estimator that knows all the $\p_j$'s and hence makes an optimal estimation. It estimates $\t_i$ using majority voting on hammers and ignores the answers of hammers. If there are no hammers then it flips a fair coin and estimates $\t_i$ correctly with half probability. It does the same in case of tie among the hammers. Concretely,
\begin{eqnarray*}
\t_i \;\; = \;\; {\rm{sign}}\bigg(\sum_{j \in W_i} \I\{j \in \H \}) A_{ij}\bigg)\,,
\end{eqnarray*} 
where $W_i$ denotes the neighborhood of node $i$ in the graph and $\H$ is the set of hammers. Note that this is the optimal estimation for the spammer-hammer model. We want to compute a lower bound on $\P[ t_i \neq \t_i |q_i]$. Let $\tl_i$ be the number of hammers answering task $i$, i.e.,$\tl_i = |W_i \cap \H|$. Since $p_j$'s are drawn from spammer-hammer model, $\tl_i$ is a binomial random variable Binom($\ell_i,\sp$). We first compute probability of error conditioned on $\tl_i$, i.e. $\P[ t_i \neq \t_i | \tl_i, q_i]$. For this, we use the following lemma from \cite{KOS14OR}.
\begin{lemma}[Lemma 2 from \cite{KOS14OR}]
For any $C < 1$, there exists a positive constant $C'$ such that when $(2q_i-1)  \leq C$, the error achieved by majority voting is at least
\begin{eqnarray}
\min_{\tau \in \mathscr{T}_{\tl}} \; \max_{t_i \in \{\pm\}} \P[ t_i \neq \t_i  |  \tl_i, q_i] \;\; \geq \;\; e^{-C'(\tl_i(2q_i-1)^2 + 1)}.
\end{eqnarray}
\end{lemma}   
Taking expectation with respect to random variable $\tl_i$ and  applying Jensen's inequality on the term in right side, we get a lower bound on the minimax probability of error in \eqref{eq:eqlb1} 
\begin{eqnarray}
\min_{\tau \in \mathscr{T}_{\tl},\t} \; \max_{\substack{\F \in \F_{\sp}\\t_i \in \{\pm\}}} \P[ t_i \neq \t_i | q_i]  \;\; \geq \;\; e^{-C'(\ell_i\sp(2q_i-1)^2 + 1)}\,.
\end{eqnarray}

%-------------------------------------------------------------------------------------

\section{Discussion}
\label{sec:discussion}

Recent theoretical advances in crowdsourcing systems have not been able to explain the 
gain in  {\em adaptive} task assignments, widely used in practice. 
This is mainly due to the fact that existing models of 
the worker responses failed to capture the heterogeneity of the tasks, while the gain in adaptivity is signified when tasks are widely heterogeneous. 
To bridge this gap, we propose studying the gain of adaptivity under a more general model recently introduced by  \cite{ZLPCS15}, which we call the generalized Dawid-Skene model. 

We identify that the minimax error rate decays as $e^{-C \lambda \sigma^2 \Gamma/m }$, 
where the dependence on the heterogeneity in the task difficulties is captured by 
the error exponent $\lambda$ defined as \eqref{eq:deflambda}. 
This is proved by showing a fundamental limit in 
Theorem \ref{thm:lb_adapt} analyzing the best possible adaptive task assignment scheme, together with the best possible 
inference algorithm, where the nature chooses the worst-case task difficulty parameters $q=(q_1,\ldots,q_m)$ and 
the worst-case worker reliability parameters $p=(p_1,\ldots,p_n)$. 
We propose an efficient adaptive task assignment scheme together with an efficient inference algorithm 
that matches the minimax error rate as shown in Theorem \ref{thm:ub_adapt}. 
To characterize the gain in adaptivity, we also identify the minimax error rate of non-adaptive schemes 
decaying as $e^{-C'\lambda_{\rm \min} \sigma^2 \ell}$, where $\lambda_{\rm min}$ 
is strictly smaller than $\lambda$. 
We show this fundamental limit in Theorem \ref{thm:thm_lb} and a matching efficient scheme in 
Theorem \ref{thm:main_thm}. 
Hence, the gain of adaptivity is captured in the 
budget required to achieve a target accuracy, 
which differ by a factor of $\lambda/\lambda_{\rm min}$.

Adaptive task assignment schemes for crowdsourced classifications have been first addressed in \cite{HJV13}, 
where a similar setting was assumed. Tasks are binary classification tasks, with heterogeneous difficulties, 
and workers arrive in an online fashion. 
One difference is that, \cite{HJV13} studies a slightly more general model where tasks are partitioned into a finite number of types and the worker error probability only depends on the type (and the identity of the worker), i.e. ${\mathbb P} (A_{ij} = t_i)= f(T(i),j) $ where $T(i)$ is the type of the task $i$. 
This includes the generalized Dawid-Skene model, if we restrict 
the difficulty $q_i$'s  from a finite set. 
\cite{HJV13} provides an adaptive scheme based on a linear program relaxation, and show that 
the sufficient condition  to achieve average error $\varepsilon$
is for the average total budget to be larger than, 
 \begin{eqnarray*}
 	\Gamma_\varepsilon &\geq&  C \frac{m}{\lambda_{\rm min} \lambda \sigma^2 } \big(\log (1/\varepsilon) \big)^{3/2} \;. 
 \end{eqnarray*}
Compared to the sufficient condition in \eqref{eq:budget_ad_ub}, this is larger by a factor of $(1/\lambda_{\rm min}) \sqrt{\log(1/\varepsilon)}$. In fact, this is larger than what can be achieved with a non-adaptive scheme in \eqref{eq:budger_nonad_lb}. 

On the other hand, there are other types of expert systems, where a finite set of experts are maintained and 
a stream of incoming tasks are assigned. This clearly departs from typical crowdsourcing scenario, as 
the experts are identifiable and can be repeatedly assigned tasks. 
One can view this as a multi-armed bandit problem with noisy feedback 
\cite{DCS09,ZSD10,EHR11,MX16}, and propose task assignment schemes with guarantees on the regret. 

We have provided a precise characterization of the minimax rate under the generalized Dawid-Skene model. 
Such a complete characterization is only known only for a few simple cases: 
binary classification tasks with symmetric Dawid-Skene model in \cite{KOS14OR} and 
binary classification tasks  with symmetric generalized Dawid-Skene model in this paper. 
Even for binary classification tasks, 
there are other models where such fundamental trade-offs are still unknown: 
e.g. permutation-based model in \cite{SBW16}. 
The analysis techniques developed in this paper does not directly generalize to such models, 
and it remains an interesting challenge.

Technically, our analysis could be improved in two directions: finite $\Gamma/m$ regime and parameter estimation.  
First, our analysis is asymptotic in the size of the problem, and 
also in the average degree of the task $\ell \equiv \Gamma/m$ 
which increases as $\log m$. 
This is necessary for applying the central limit theorem. 
However, in practice, we observe the same error rate when $\ell$ does not necessarily increases with $m$. 
In order to generalize our analysis to finite $\ell$ regime, 
we need sharp bounds on the tail of a sub-Gaussian tail of the distribution of the messages. 
This is partially plausible, and we provide an upper bound on this tail in \eqref{eq:eq28}.
However, the main challenge is that we also need a lower bound on this tail, which is generally difficult. 

Secondly, we empirically observe that our parameter estimation algorithm in Algorithm \ref{algo:est} works well in practice. It is possible to precisely analyze the sample complexity of this estimator using spectral analysis. 
However, such an error in the value of $\rho_{t,u}^2$  used in the inner-loop can result in accumulated errors over iterations, 
and it is not clear how to analyze it. 
Currently, we do not have the tools to analyze such 
error propagation, which is a challenging research direction. 
Also, the parameter estimation algorithm can be significantly improved, 
by applying some recent advances in 
estimating such smaller dimensional spectral properties 
of such random matrices, for example \cite{ZWJ15,KV16,LW17,KO17}, 
 which is an active topic for research. 
%and we could bring some of those techniques to improve the parameter estimation algorithm of  Algorithm \ref{algo:est} with provable guarantees. 
%However, it is not clear how to analyze how this error in $\rho_{t,u}$ will propagate in the overall adaptive approach.

\section*{Acknowledgements} 

This work is supported by NSF SaTC award CNS-1527754, NSF CISE award CCF-1553452, NSF CISE award CCF-1705007 and GOOGLE Faculty Research Award.

\bibliographystyle{plain}

\bibliography{Crowdsourcing}

\end{document}